%% file: main.tex
\documentclass[10pt,journal,compsoc]{IEEEtran}

\ifCLASSOPTIONcompsoc
  \usepackage[nocompress]{cite}
\else
  \usepackage{cite}
\fi

\ifCLASSINFOpdf
  \usepackage[pdftex]{graphicx}
\else
\fi

\usepackage[utf8]{inputenc} % allow utf-8 input
\usepackage{hyperref}       % hyperlinks
\usepackage{url}            % simple URL typesetting
\usepackage{booktabs}       % professional-quality tables
\usepackage{amsfonts}       % blackboard math symbols
\usepackage{nicefrac}       % compact symbols for 1/2, etc.
\usepackage{microtype}      % microtypography
\usepackage{xcolor}   
\usepackage{amsmath}
\usepackage{amssymb}
\usepackage{graphicx}
\usepackage{array}
\usepackage{comment}
\usepackage{url}
\usepackage{subfig}
\usepackage{times}
\usepackage{algorithm,algorithmic,xspace}
\usepackage{verbatim}
\usepackage{multirow}
\usepackage{colortbl}
\usepackage{bbding}
\usepackage{pifont}
\usepackage{wasysym}
\usepackage{amssymb}
\usepackage{breqn}
\usepackage{bbm}
\usepackage{tabularx}
\usepackage{subfig}
\usepackage{times}
\usepackage{enumerate}

\newcommand{\Tref}[1]{Table~\ref{#1}}

\newcommand{\Fref}[1]{Figure~\ref{#1}}
\newcommand{\Sref}[1]{Section~\ref{#1}}

\def\etal{\emph{ et al.}}
\def\ie{\emph{i.e.}}
\def\eg{\emph{e.g.}}

\def\eg{{\emph{e.g.}}}
\def\ie{{\emph{i.e.}}}
\def\etc{{\emph{etc.}}}
\def\etal{{\emph{et al.}}}

\def\loss{{\mathcal{L}}}

\def\Pi{{\mathbf{M}_{i}}}

\definecolor{myGreen}{rgb}{0, .8, .3}
\definecolor{myRed}{rgb}{0.8, .2, .2}

\def\edge{hand-edge} %todo

\makeatletter
\usepackage{array}
\usepackage{float}
\usepackage{makecell}
\begin{document}
\title{EvHandPose: Event-based 3D Hand Pose\\ Estimation with Sparse Supervision}

\author{Jianping Jiang${}^{\dag}$, Jiahe Li${}^{\dag}$, Baowen Zhang, \\
Xiaoming Deng${}^{\ddag}$,~\IEEEmembership{Member,~IEEE,}
Boxin~Shi${}^{\ddag}$,~\IEEEmembership{Senior~Member,~IEEE}
% <-this % stops a space
\IEEEcompsocitemizethanks{
\IEEEcompsocthanksitem J. Jiang, B. Shi are with National Key Laboratory for Multimedia Information Processing, National Engineering Research Center of Visual Technology, and AI Innovation Center, School of Computer Science, Peking University, Beijing 100871, China.
\IEEEcompsocthanksitem J. Li, B. Zhang, X. Deng are with Beijing Key Laboratory of Human Computer Interactions, Institute of Software, Chinese Academy of Sciences, Beijing 100190, China, and University of Chinese Academy of Sciences, Beijing, China.
\IEEEcompsocthanksitem ${}^{\dag}$ Equal contribution. ${}^{\ddag}$ Corresponding authors: shiboxin@pku.edu.cn, xiaoming@iscas.ac.cn}% <-this % stops an unwanted space
}

\IEEEtitleabstractindextext{%
\begin{abstract}
Event camera shows great potential in 3D hand pose estimation, especially addressing the challenges of fast motion and high dynamic range in a low-power way. 
However, due to the asynchronous differential imaging mechanism, it is challenging to design event representation to encode hand motion information especially when the hands are not moving (causing motion ambiguity), and it is infeasible to fully annotate the temporally dense event stream.
In this paper, we propose \textit{EvHandPose} with novel hand flow representations in Event-to-Pose module for accurate hand pose estimation and alleviating the motion ambiguity issue.
To solve the problem under sparse annotation, we design contrast maximization and \edge{} constraints in Pose-to-IWE (Image with Warped Events) module and formulate EvHandPose in a weakly-supervision framework.
We further build \textit{EvRealHands}, the first large-scale real-world event-based hand pose dataset on several challenging scenes to bridge the real-synthetic domain gap.
Experiments on EvRealHands demonstrate that EvHandPose outperforms previous event-based methods under all evaluation scenes, achieves accurate and stable hand pose estimation with high temporal resolution in fast motion and strong light scenes compared with RGB-based methods, generalizes well to outdoor scenes and another type of event camera, and shows the potential for the hand gesture recognition task.
\end{abstract}
}

\maketitle
\IEEEdisplaynontitleabstractindextext
\IEEEpeerreviewmaketitle

\IEEEraisesectionheading{\section{Introduction}\label{sec:introduction}}

\IEEEPARstart{H}{and} pose estimation plays a key role in computer vision, human-computer interaction, virtual reality, and robotics \cite{lepetit2020recent,deng2022recurrent}.
Most of the existing vision-based hand pose estimation approaches adopt conventional imaging sensors such as RGB or RGB-D cameras. Although these sensors are widely available and effective for various hand pose relevant applications, due to the limitations from their imaging mechanisms, it is still challenging to achieve robust and accurate hand pose results in many real scenarios, especially for those with fast motions and abrupt changes of lighting conditions.
\begin{figure}[t]
  \centering
  \includegraphics[width=\linewidth]{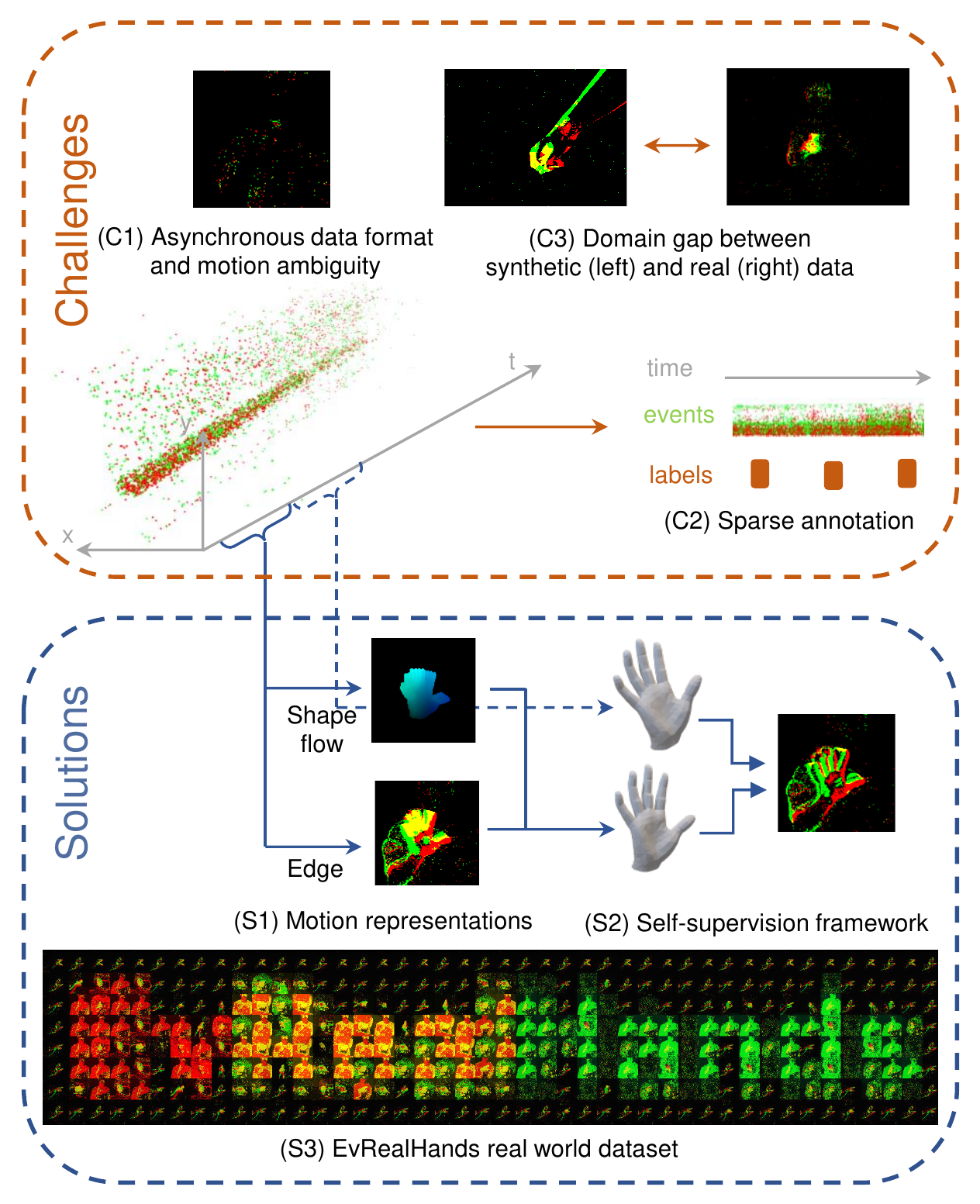}
  \caption{
  Hand pose estimation from event data faces several challenges: (C1) asynchronous data format and motion ambiguity issue caused by inherent data properties of events, (C2) sparse annotation with current annotation techniques, (C3) domain gap between synthetic and real data. In order to tackle these issues, we propose \textit{EvHandPose} by extracting novel motion representations including edge and shape flow from event streams with temporal information to predict hand poses (S1). We design a weakly-supervision framework to tackle the sparse annotation issue (S2). Furthermore, we present the first large real-world event-based hand pose dataset \textit{EvRealHands} (S3) (zoom-in for details).}
  \label{teaser}
\end{figure}
Event cameras \cite{Lichtsteiner2008} that generate neuromorphic events by measuring per-pixel intensity changes asynchronously have shown great potential in many vision tasks such as detection \cite{mitrokhin2018event}, tracking \cite{gehrig2018asynchronous}, 3D reconstruction \cite{rebecq2018emvs},
flow estimation \cite{hagenaars2021self}, simultaneous localization and mapping \cite{rebecq2016evo, zhou2021event}, due to the advantages of high dynamic range (HDR, 120 dB), low power consumption, and high temporal resolution (up to 1 $\mu$s). 
Recently, event cameras have also demonstrated their potential in robust 3D hand pose estimation for scenes with fast motions and broad dynamic range, because of their rich temporal information
and HDR property \cite{rudnev2021eventhands}.

% Working as an asynchronous differentiator for visual signal capture, event camera generates asynchronous event streams with time resolution up to 1 us at the expense of lacking in appearance and texture information.
Due to the lack of appearance and texture information in event streams, estimating hand poses using events faces three key challenges as shown in \Fref{teaser} (C1)-(C3).
% We consider the challenge of events from three perspectives.
First,  from the data inherent properties perspective, asynchronous event streams are spatially sparse and temporally dense, so existing image-based hand pose estimation solutions can not be directly applied to events.
A straightforward solution is to reconstruct intensity frames from events via an ``integration" process  as \cite{rebecq2019high} and \cite{mostafavi2021learning}.
However, such strategies would sacrifice the advantage of temporally fast and dense events, and it is still an open problem to design effective event representation and feature extractor for event-to-image conversion without losing these advantages.
Moreover, event cameras suffer from the motion ambiguity issue, since the static hand parts can not generate events under the assumption of brightness consistency, which may result in that different hand poses generate the same event stream (only relative motion is recorded).
Second, it is generally infeasible to annotate the asynchronous event stream  using existing hand pose annotation techniques.
For annotators, they could only provide sparsely annotated hand poses by sampling the key frames along the temporal dimension of event streams, so the event-based hand pose model is expected to be learned with sparse hand supervision in the temporal dimension. Recently, a pilot study \cite{rudnev2021eventhands} was conducted to learn 3D hand pose from event streams, which uses event simulators to generate synthetic dataset and evaluates the performance on real data. Although promising results are achieved, the third challenge is that the existing method \cite{rudnev2021eventhands} suffers from domain gap between synthetic and real data, since a small real dataset that can not support strong model generalization ability desired for different challenging scenarios.

To address the above challenges, we propose \textit{EvHandPose} -- a weakly-supervision framework to learn event-based 3D hand pose estimation under sparse supervision, as shown in \Fref{teaser} (S1)-(S3). First, in order to deal with the asynchronous data format and motion ambiguity issue, we design specific hand flow representation and apply temporal information to regress the accurate hand pose. 
And we propose to approximate the hand optical flow in a sequence of event stream by embedding the interpolated hand models.
Second, to address the sparse annotation challenge, we apply contrast maximization and \edge{} constraints in a weakly-supervision framework. We first train our Event-to-Pose module on labeled data to get prior knowledge
and then fine-tune the model by proposing a weakly-supervision framework, which is supervised by a Pose-to-IWE (Image with Warped Events) module during training.
Finally, to address the domain gap issue limited by synthetic datasets, we construct \emph{EvRealHands}, the first large-scale real event-based hand pose dataset under various challenging scenarios.
Experiments demonstrate that our event camera based solution outperforms conventional RGB camera based solutions in challenging scenarios such as fast motion and strong light with high temporal resolution, and generalize well to outdoor scenes and other event cameras. Our dataset, code and models will be made public after acceptance.

Our main contributions are threefold:
\begin{itemize}[\IEEEsetlabelwidth{Z}]
\item Novel hand flow representations: Asynchronous event data are effectively processed in Event-to-Pose module for accurate hand pose estimation and alleviating the motion ambiguity issue.
\item Weakly-supervision framework: Pose-to-IWE module is complementarily designed to tackle the sparse annotation issue with contrast maximization and \edge{} constraints specially designed for the hand.
\item Dataset: The first large-scale real dataset \emph{EvRealHands}, consisting of 79 minutes event sequences in indoor and outdoor scenes, 425 K RGB images, with accurate 3D pose and shape annotations is built to tackle the domain gap issue and facilitate future research.
\end{itemize}

\section{Related Work}\label{sec:related work}
\subsection{3D Hand Pose Estimation}
Prior arts of 3D hand pose estimation methods consist of appearance-based methods \cite{zimmermann2017learning, iqbal2018hand, ge20193d, zhang2019end, chen2021mobrecon} and model-based approaches \cite{oikonomidis2014evolutionary,tzionas2016capturing,oikonomidis2012tracking,ballan2012motion,kyriazis2014scalable}. 
Appearance-based methods directly learn the mapping from image space to hand pose space by training on large-scale datasets, which often require large-scale annotated dataset. Model-based methods are designed to fit a prior hand parametric model such as MANO \cite{romero2017embodied} to the image observation. However, these methods require a good initialization for optimization. 
In order to reduce the expensive annotation cost of large scale dataset, several efforts are adopted to learn hand pose from unlabeled data in weakly-supervised \cite{cai2018weakly, spurr2020weakly, deng2020weakly, kulon2020weakly} or self-supervised \cite{chen2021model, wan2019self, boukhayma20193d} manners.
These methods use additional constraints such as segmentation mask, depth, appearance, and biomechanics consistency to ensure the quality of the predicted hand pose.
When dealing with hand pose estimation from video sequences, learning from sparse-labeled data is effective to reduce the burden of full annotations for video sequences, and the existing methods often leverage motion or photometric consistency as constraints between labeled and unlabeled frames~\cite{neverova2019slim, hasson2020leveraging} to train models.
However, there is no weakly-supervised or self-supervised framework specially designed for event-based hand pose estimation.

\subsection{Event-based Pose Estimation}

The first challenge for event-based 3D hand pose estimation is the real-synthetic domain gap. Previous synthetic event-based dataset~\cite{rudnev2021eventhands} faces two issues: 1) lack of hand pose diversity and 2) domain gap of synthetic data due to the event simulator~\cite{gehrig20v2e}.
Although domain adaptation methods, such as NGA~\cite{hu20nga} and EvTransfer~\cite{messikommer22evtransfer}, tend to utilize the shared information between events and images to bridge the gap, they can hardly achieve superior results compared with supervised methods.
Since there is no large-scale real event-based hand dataset with 3D pose annotation, 
it motivates us to build a new real event-based hand dataset. In this work, we select multiple ubiquitous color cameras to get 3D hand pose annotations of event camera.

Since the events are mainly generated from the edge of the moving objects, they can preserve the information of hand movement. 
Existing methods use different representations of event streams, such as event count image (ECI)~\cite{maqueda2018count}, time surface~\cite{lagorce2016hots}, voxel~\cite{zhu2019unsupervised}, time-ordered recent event volumes (TORE)~\cite{Baldwin23TORE}, event spike tensor (EST)~\cite{gehrig2019spiketensor}, Matrix-LSTM~\cite{cannici2020matrixlstm} \etc{} for learning models, while they are not specifically designed to represent hand motions for 3D hand pose estimation.
Therefore, a key problem for event-based hand pose estimation is to leverage edge and movement features for pose estimation.

Optical flow is an effective feature in event-based vision tasks, and it can serve as a useful approximation to hand motion fields when events are used for hand pose estimation. 
Although optical flow from event streams has been extensively studied \cite{hagenaars2021self,zhu2018ev, zhu2019unsupervised}, to the best of our knowledge, no existing work is specifically designed for hand motion.
EV-FlowNet \cite{zhu2018ev} learns self-supervised optical flow by photometric consistency of gray-scale images from the DAVIS camera \cite{Lichtsteiner2008}, but it fails when capturing gray-scale images from fast motion or HDR scenes. 
To tackle this, Zhu \etal \cite{zhu2019unsupervised} learn an effective optical flow model in an unsupervised manner based on the contrast maximization method.
However, it is still a open problem to estimate the hand specific optical flow (\emph{hand flow}) by leveraging the hand shape model, and how to use hand flow to boost the hand pose estimation performance.  

% To utilize the edge information naturally encoded in events, EventCap \cite{xu2020eventcap} applies an edge fitting approach for human pose refinement which consists of two steps: 
% 1) finding correspondences of events and mesh vertices, 2) minimizing the distance between events and 2D projections from corresponding vertices.
% Finding correspondences is the main challenge in the edge fitting procedure.
% EventCap \cite{xu2020eventcap} obtains the closest event for each boundary vertex in its closest event search, but this leads to two issues when tackling hand pose estimation task: 1) events are not aligned at the same time with the hand mesh,  2) fingers will gather together in optimization, and the optimization can be stuck in local minimum because an event may have several corresponding vertices.
% Thus, it is important to design effective edge constraints specially designed for hand.
% Incorporating the hand model into the estimated optical flow can effectively restore the hand motion process.

In the realm of pose estimation, event camera is firstly applied to human pose estimation. EventCap \cite{xu2020eventcap} utilizes initial pose estimation via gray-scale images to reconstruct high-frequency human pose motion by event trajectories. However, EventCap \cite{xu2020eventcap} requires additional gray-scale images (usually with generated qualities for scenes that events demonstrate advantages) as input, and could lead to unstable pose results under challenging illumination conditions or fast motions. EventHPE \cite{zou2021eventhpe} estimates human pose from only event data, but predicting relative poses of neighboring event frames will lead to accumulated errors as time goes on.

When events are used for hand pose estimation, it is more challenging due to high degree-of-freedom hand motion, heavy self-occlusion and significant similarity between different hand parts. 
EventHands \cite{rudnev2021eventhands} is proposed to estimate 3D hand pose estimation from event streams. It estimates MANO parameters from Locally-Normalised Event Surfaces (LNES), which assigns higher weights to events close to the target timestamp and forms an event frame. However, it fails to resolve the motion ambiguity in pose estimation, \ie, static hand parts will not trigger events, which leads to spatial uncertainty of these static parts.
Jalees \etal \cite{nehvi2021differentiable} also propose a model-based approach to estimate 3D hand pose from event streams, which fits a hand model via nonlinear optimization and enforces event stream simulator to generate similar event frames to the real event camera. 
However, the hand tracking by optimization requires good initialization, and it is prone to be stuck at local minimum due to the non-convex optimization of the object function.

Besides, due to the asynchronous imaging mechanism between color cameras and event cameras, hand pose annotations from color cameras are very sparse compared to the temporally dense event streams.
To deal with the sparse annotation issue, we need to design effective constraints to enforce the predicted hand poses of the unlabeled frames to be plausible and well aligned with events.

\begin{figure*}[ht]
  \centering
  \includegraphics[width=\textwidth]{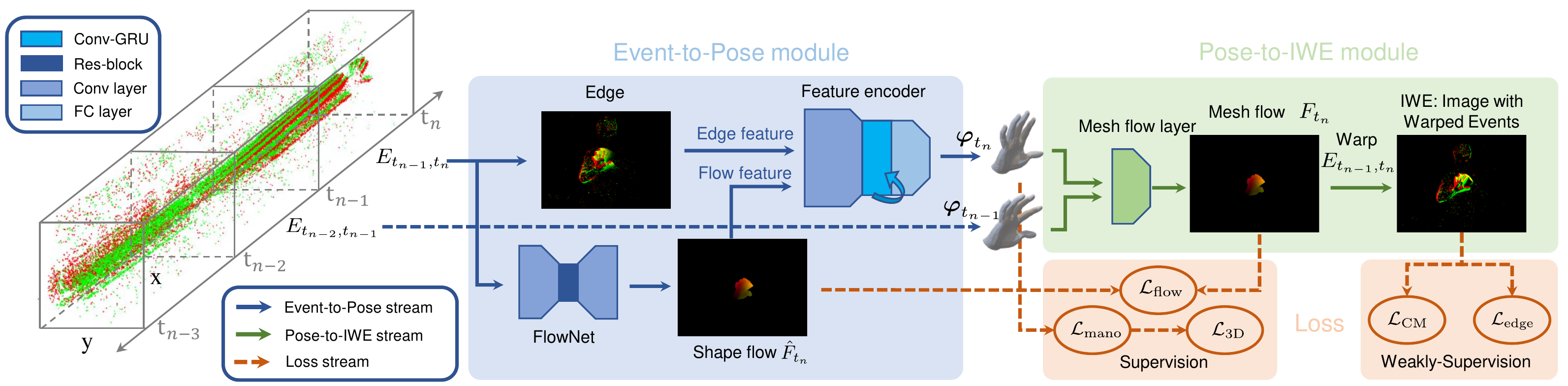}
  \caption{Overview of EvHandPose pipeline. Event streams are firstly split into several sub-segments $\{E_{t_{k-1}, t_{k}}\}$, and they are represented as event frames and volumes.
  Event-to-Pose module extracts edge of each frame and shape flow $\hat{F}_{t_n}$ of each volume by FlowNet, and predicts the hand parameters $\boldsymbol{\varphi}_{t_{n}}$ of the current event timestamp. 
  Given sparsely annotated and unannotated event frames, we learn the hand pose estimation model by designing several weakly-supervision loss functions.  
  Specially, we design a differential mesh flow layer (details in \Fref{Pose-to-IWE}) in Pose-to-IWE module to compute mesh flow ${F}_{t_n}$ with the neighboring hand parameters $\boldsymbol{\varphi}_{t_{n}}$ and $\boldsymbol{\varphi}_{t_{n-1}}$. 
  In order to address the issue of sparse annotation, we warp the events in $E_{t_{n-1}, t_{n}}$ with estimated hand flow to get IWE, and adopt contrast maximization loss and \edge{} loss upon IWE for weakly-supervision.
  In the 2D visualization of events in this paper, red pixels represent positive events, green pixels represent negative events, and yellow pixels represent both types of events.
}
  \label{Pipeline}
%   \vspace{-1.2em}
\end{figure*}

\section{Preliminaries}\label{sec: Preliminaries}

\subsection{Hand Model Representation}
We adopt a differentiable hand parametric MANO model \cite{romero2017embodied}, which can be formulated as:
\begin{equation}
    M(\boldsymbol{\beta}, \boldsymbol{\theta})=W\left(T_{P}(\boldsymbol{\beta}, \boldsymbol{\theta}), J(\boldsymbol{\beta}), \boldsymbol{\theta}, \mathcal{W}\right),
\end{equation}
where a skinning function $W$ (linear blend skinning \cite{LBS_2000_ToG}) is applied to an articulated rigged body mesh with shape $T_{P}$, joint locations $J$ defining a kinematic tree, pose parameter $\boldsymbol{\theta}$, shape parameter $\boldsymbol{\beta}$, and blend weights $\mathcal{W}$. 
Given pose and shape parameters, MANO model outputs a hand mesh with 778 vertices $\mathbf{v} = M(\boldsymbol{\beta}, \boldsymbol{\theta}) \in \mathbb{R}^{778 \times 3}$, and 3D joints $\mathbf{J}_{\text{3D}} \in \mathbb{R}^{21 \times 3}$ can be recovered by $\mathbf{J}_{\text{3D}} = J_{\text{reg}}({\boldsymbol{\beta}}, {\boldsymbol{\theta}})$, where $J_{\text{reg}}$ is the regression function from the hand mesh and pose parameters to 3D joints. Hereafter, we denote $\boldsymbol{\varphi}=(\boldsymbol{\beta}, \boldsymbol{\theta})$ to be \textit{hand parameters} for easy reference.

\subsection{Image Formation and Motion Ambiguity}
Event camera encodes the changes of per-pixel intensity $I(x, y, t)$ into event streams. An event $e_i = (x_i, y_i, t_i, p_i)$ is triggered at pixel $(x_i, y_i)$ at time $t_i$ when the logarithmic radiance change reaches the threshold:
\begin{equation}
    \log I(x_i, y_i, t_i) - \log I(x_i, y_i, t_p) = p_i C,
\end{equation}
where $t_p$ is the timestamp of the last event fired at pixel $(x_i, y_i)$, $p_i \in \{ -1, 1\}$ is the polarity of intensity change, $C$ is the contrast threshold.

Due to the differential imaging mechanism of the event camera, the static hand part will not generate events under constant lighting.
Events $ E_{t_{n-1}, t_n}$ within time interval $[t_{n-1},t_{n}]$ may correspond to multiple hand poses (we call it ``motion ambiguity" in event-based hand pose problem), because the static hand parts can have multiple pose states and their relative pose changes reflected by events could look the same.
Several efforts such as EventHPE \cite{zou2021eventhpe} are conducted to predict relative pose $\Delta\boldsymbol{\theta}_{t_n}$ between the current frame and the previous frame $\boldsymbol{\varphi}_{t_{n-1}}$. However, these approaches can not resolve the motion ambiguity issue, because the relative poses are still affected by the absolute hand poses even if they generate the same event stream.

One way to alleviate the challenge is to encode the previous event data to assist in estimating the current hand pose, because event camera is more likely to record complete hand movements over a long interval. In this paper, we use a Conv-GRU \cite{shi2015convolutional} recurrent model to embed the previous sequence features.
We divide the asynchronous event sequence into several sub-segments for hand pose prediction.
Given an event stream segment ranging from $t_{0}$ to $t_{N}$, we split the segment into $N$ sub-segments $\{[t_{i-1}, t_{i}]\}_{i=1}^{N}$ with the same duration. Each sub-segment of the event stream will be represented as event frame and event volume for hand pose estimation.

\section{Methods}\label{sec: Methods}

As shown in \Fref{Pipeline}, EvHandPose pipeline consists of two main parts: Event-to-Pose module and Pose-to-IWE (Image with Warped Events) module.
The Event-to-Pose module is designed to predict hand pose from the asynchronous event stream in Section~\ref{Motion Representations}.
It represents the edge feature as LNES \cite{rudnev2021eventhands} and predicts the shape flow (\ie{}, a new optical flow specially designed for hands) with the proposed FlowNet to explore motion information from the asynchronous event stream.
After fusing the edge feature and shape flow by a feature encoder (ResNet34~\cite{he2016deep} backbone), a Conv-GRU model is applied to alleviate the motion ambiguity issue by involving the temporal correlation between the neighboring event frames.
The Pose-to-IWE module is then designed to supervise Event-to-Pose module during training.
The key idea is to use the sparse annotations of hand shapes and hand joints as supervision when they are available, use a contrast maximization loss and \edge{} loss on the unannotated event frames to enhance hand pose estimation performance, and use hand flow loss for all event frames. In Section~\ref{sec:framework}, we elaborate on the details of these loss functions for annotated and unannotated event frames, respectively.

\subsection{Motion Representations}\label{Motion Representations}
Events in time interval $[t_{n-1},t_{n}]$ contains two types of hand motion aware information (\ie, optical flow and edge) under photometric consistency assumption. By recording the brightness difference between the hand and background, events in $[t_{n-1},t_{n}]$ describe the hand movement and can be used to resolve the motion ambiguity issue of hand pose estimation by including observations from previous status, and events at a specific timestamp $t_{n}$ indicates the accurate hand edges. Therefore, we propose new motion representations that include both hand flow and edge information for event-based hand pose estimation. 

As edges are naturally encoded in events, we utilize the edge information in Event-to-Pose module for accurate pose estimation.
Specifically, we follow EventHands \cite{rudnev2021eventhands} to use the effective 2D representation LNES.
The LNES image $I$ is updated by iterating each event $e_i$ in $ E_{t_{n-1}, t_n}$:
\begin{equation}
    I(x_i, y_i, p_i) = \frac{t_i - t_{n-1} }{t_n - t_{n-1}}.
\end{equation}
The LNES can reserve the latest event information because it will be overwritten when an event with the same polarity occurs at that location. 

\subsubsection{Hand Flow from Events}
\label{sec:hand motion}
\begin{figure}[t]
  \centering
  \includegraphics[width=\linewidth]{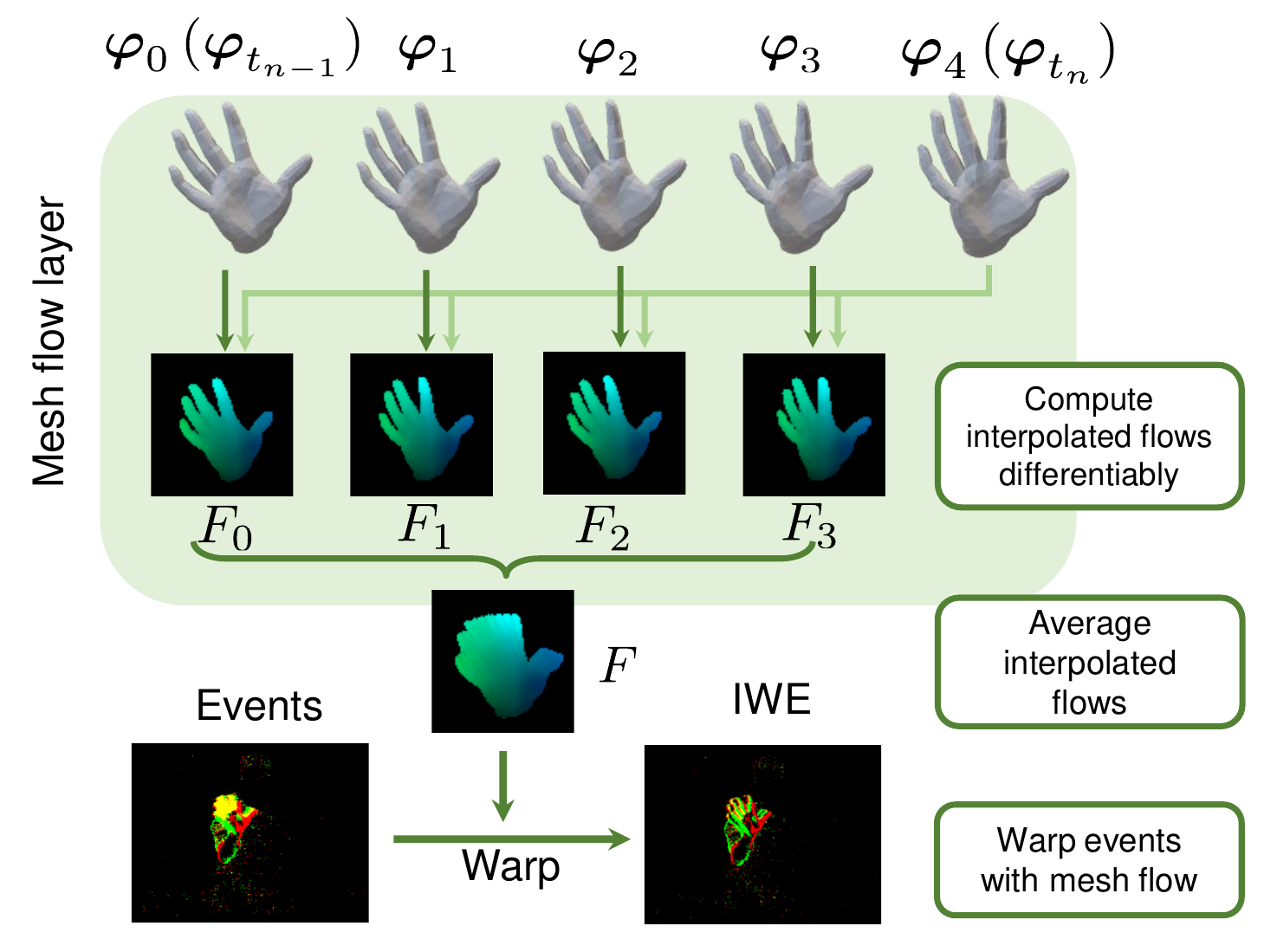}
  \caption{Illustration of Pose-to-IWE module. Given MANO parameters $\boldsymbol{\varphi}_{t_{n-1}}$ and $\boldsymbol{\varphi}_{t_{n}}$, we first conduct interpolation to get
  interpolated MANO parameters $\{\boldsymbol{\varphi}_k | k=0,1,\dots,K\}$, $K=4$ here. Then we compute the interpolated flows $\{F_k\}$ by projecting the movements between each hand mesh $M_k (k = 0,1,...,K-1)$ and the last hand mesh $M_K$ to the image plane. Lastly, we average all the interpolated flows to get the final hand mesh flow $F$. }
  \label{Pose-to-IWE}
\end{figure}

We use two types of representation of hand flow from events. The first type is the model-based hand flow $F$ (denoted as \textit{mesh flow}) by designing differential hand flow layer with the neighboring hand parameters $\boldsymbol{\varphi}_{t_{n}}$ and $\boldsymbol{\varphi}_{t_{n-1}}$, shown in \Fref{Pose-to-IWE}.
The second type of hand flow $\hat{F}$ (denoted as \textit{shape flow}) is obtained by the optical flow predicted by proposing a FlowNet (bottom right of Event-to-Pose module in \Fref{Pipeline}), which is learned using mesh flow as supervision.
\vspace{1mm}

\noindent\textbf{Mesh Flow.}
Desired motion representations for event-based hand modeling should precisely describe the hand movement and ignore disturbance from the background.
To achieve this, we calculate the dense hand flow by projecting two sequential MANO meshes on the image, and compute the displacements of corresponding vertices like \cite{zou2021eventhpe} and \cite{hasson2020leveraging}.

\Fref{Pose-to-IWE} illustrates hand flow computation process. Given two MANO models $M(\boldsymbol{\beta}_{t_{n-1}}, \boldsymbol{\theta}_{t_{n-1}})$ and $M(\boldsymbol{\beta}_{t_{n}}, \boldsymbol{\theta}_{t_{n}})$ at two event moments, we get $K+1$ hand shapes $\{M_k\}_{k=0}^K$ with hand parameters $\{\boldsymbol{\varphi}_{k}\}_{k=0}^K$ by linear interpolation of the hand parameters of the two MANO models as follows:
\begin{align}
    t_{k} &= (1- \frac{k}{K}) \cdot {t_{n-1}} + \frac{k}{K} \cdot {t_{n}}, &\quad k=0,1,..., K. \\
    \boldsymbol{\beta}_{k} &= (1- \frac{k}{K}) \cdot \boldsymbol{\beta}_{t_{n-1}} + \frac{k}{K} \cdot \boldsymbol{\beta}_{t_{n}}, &\quad k=0,1,..., K.\\
    \boldsymbol{\theta}_{k} &=  slerp(\boldsymbol{\theta}_{t_{n-1}}, \boldsymbol{\theta}_{t_{n}},\frac{k}{K}), &\quad k=0,1,..., K.
\end{align}
where spherical linear interpolation (\ie{}, $slerp(\cdot)$)~\cite{Shoemake85slerp} is used for pose parameters $\boldsymbol{\theta}$, and $\boldsymbol{\varphi}_{k} = (\boldsymbol{\theta}_{k}, \boldsymbol{\beta}_{k}, t_k)$ is the interpolated MANO parameters at time $t_{k}$.

For each interpolated hand mesh $M_k$, we can compute the speed $\mathbf{S}_k \in \mathbb{R}^{778 \times 2 }$ of all vertices of the hand mesh on the image as follows:
\begin{equation}
    \mathbf{S}_{k, i} = \frac{K\cdot\pi(\mathbf{v}_{K, i} - \mathbf{v}_{k, i})}{(K-k)\cdot (t_{n}-t_{n-1})} , \qquad k=0, 1, 2, ..., K-1,
\end{equation}
where $\mathbf{v}_{k, i}$ is the $i$-th vertex on the hand mesh $M_k$, and $\pi$ is the camera projection matrix.

Since the hand mesh represented by MANO model is a triangular mesh with 778 vertices, the displacements of the corresponding vertices of neighboring interpolated hand mesh $M_k$ can not get dense optical flow.
In order to deal with this issue, we get the corresponding triangle of each pixel $(x, y)$ by the first intersection of the ray from the camera optical center to the pixel $(x, y)$ and the hand mesh, and calculate the bary centric coordinates $b_{\mathbf{v}_j}(x, y)$ of the pixel $(x, y)$.
% \fi 
Then we can compute the dense optical flow by the weighted sum of the vertex speed $S_k$
with bary centric weights $b_{\mathbf{v}_j}$ as follows:
\begin{equation}
    F_{k}(x,y) = \sum_{\mathbf{v}_j \in V_{\text{bary}}(x, y)}b_{\mathbf{v}_j}(x, y)\mathbf{S}_{k, j},
\end{equation}
where ${V_{\text{bary}}(x, y)}$ is the vertices of the corresponding triangle of pixel $(x, y)$.
Then our mesh flow representation $F(x, y)$ , which is differentiable with respect to hand parameters, is calculated as: 
\begin{equation}
    F(x, y) = \frac{\sum_{k=0}^{K-1}F_{k}(x, y)}{\sum_{k=0}^{K-1}\mathbbm{1}(F_{k}(x, y))} ,
\end{equation}
where $\mathbbm{1}$ is the function to indicate whether there is flow at pixel $(x, y)$. Events are mainly triggered from the edge of the hand under constant lighting condition. If we warp a period of events to a certain timestamp, events from the same position on the hand will gather together to form sharp hand edges and accumulated an image with warped events. 
\Fref{Pose-to-IWE} illustrates that mesh flow layer and IWEs using our mesh flow have sharp edges, which will be used as the contrast maximization and \edge{} constraints in our weakly-supervision framework.

\vspace{1mm}
\noindent\textbf{Shape Flow.}
While mesh flow is obtained from the mesh flow layer and serves for weakly-supervision losses, Event-to-Pose module needs to directly extract motion information from event streams for accurate hand pose estimation.
Thus Event-to-Pose module uses FlowNet which is a U-Net like CNN backbone to predict hand shape flow from event volumes \cite{zhu2019unsupervised}.
Being different from \cite{zhu2019unsupervised}, our FlowNet is supervised by mesh flow and only learns hand movement information, not background movement information, thus reserving high quality hand movement information. We apply the endpoint error (EPE) between shape flow $\hat{F}_{t_{n}}$ and mesh flow $F_{t_n}$ as loss function, which is the normal measurement for optical flow estimation:
\begin{equation}
    \loss_{\text{flow}} = \sum_{n}\text{EPE}(\hat{F}_{t_n}, F_{t_n}).
\end{equation}

\subsection{Weakly-supervision Learning Framework}
\label{sec:framework}
The annotations are only available for sparse subset of event frames, because the annotations are provided with sparse frames of multiple color cameras (details in Section~\ref{sec: Dataset}).
In this section, we formulate the weakly-supervision framework that can learn the hand pose model using both the sparse annotated and unannotated event frames.

\subsubsection{Learning under Sparse Annotations}
\label{sec:supervision}
Event-to-Pose module of EvHandPose extracts edge feature and hand flow from event streams, and predicts the hand MANO parameters.
When annotations of 3D joints and hand MANO parameters are available, we use 3D joint loss $\loss_{\text{3D}}$ and hand MANO parameter loss $\loss_{\text{mano}}$ for model learning:
\begin{equation}
    \loss_{\text{3D}} = \sum_{n}{\lVert J_{\text{reg}}(\boldsymbol{\hat{\beta}}_{n}, \boldsymbol{\hat{\theta}}_{n}) - \mathbf{J}_{\text{3D}_{n}}\rVert^{2}},
\end{equation}
\begin{equation}
     \loss_{\text{mano}} = \sum_{n}{(\lambda_{\beta}\lVert\boldsymbol{\hat{\beta}}_{n} - \boldsymbol{\beta}_{n} \rVert^{2} + \lambda_{\theta}\lVert\boldsymbol{\hat{\theta}}_{n} - \boldsymbol{\theta}_{n} \rVert^{2})},
\end{equation}
where $\mathbf{J}_{\text{3D}_{n}}$ is the ground-truth 3D joints vector, 
MANO parameters with hat $(\boldsymbol{\hat{\beta}}, \boldsymbol{\hat{\theta}})$ are prediction results and those without hat $(\boldsymbol{\beta}, \boldsymbol{\theta})$ are ground-truth, 
$\lambda_{\boldsymbol{\beta}}$ and $\lambda_{\boldsymbol{\theta}}$ are loss weights and they are set to 0.25 and 5.0, respectively.

Therefore, the loss function $\loss_{\text{super}}$ to learn under supervision can be formulated as follows:
\begin{equation}
    \loss_{\text{super}} = \lambda_{\text{3D}}\loss_{\text{3D}} + \lambda_{\text{mano}}\loss_{\text{mano}},
\end{equation}
where $\lambda_{\text{3D}}$ and $\lambda_{\text{mano}}$ are loss weights,
and they are set to 1000 and 0.1 in our experiments, respectively.

\begin{figure}[t]
  \centering
  \includegraphics[width=\linewidth]{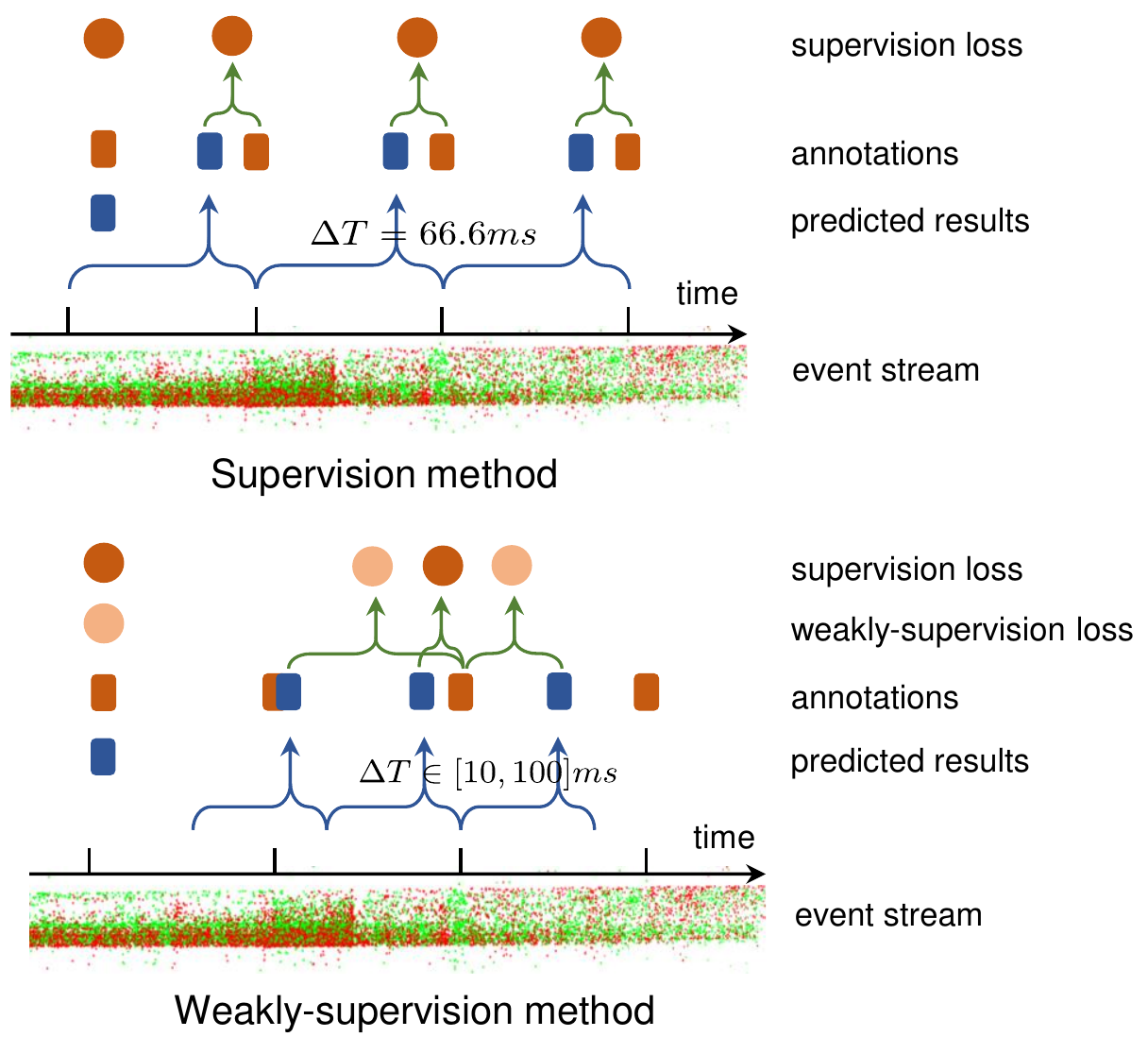}
  \caption{Comparison of supervision and weakly-supervision methods. In supervision method, the model can only learn hand poses at annotated timestamps and the interval of each
  event sub-segment has to be strictly overlapped with the annotated period. In weakly-supervision method, we only need partial labels and the outputs of unlabeled sub-segments are constrained by weakly-supervision loss. Therefore, EvHandPose can learn hand poses at any timestamp near the annotations with dynamic time window for event sub-segment representation.}
  \label{weakly-supervision}
\end{figure}

\subsubsection{Learning with Weakly-supervision}
\label{sec:weaklysupervision}

Since we get the annotated hand pose dataset EvRealHands using multiple RGB cameras, hand pose can be annotated at a pre-defined fps (\eg, 15 fps in EvRealHands). Due to the asynchronous image mechanism of event camera, the event streams can only have sparse hand pose annotations.  
In order to supervise the unannotated event sub-segment and leverage the dense event streams, we propose a weakly-supervision framework to effectively utilize the event with sparse annotations as shown in \Fref{weakly-supervision}. Specially, we design a contrast maximization loss to constrain the movement of the hand over time, and an \edge{} loss to learn hand pose at predicted time, and adopt smooth loss to encourage the smoothness of the sequential predicted hand poses. 

\vspace{1mm}
\noindent\textbf{Contrast Maximization Loss.}
Inspired by the contrast maximization method that was originally proposed to handle motion compensation problem \cite{gallego2018unifying}, we design a new contrast maximization loss (denoted as CM loss) for weakly-supervision. The key idea of CM loss is to enforce the IWE is sharp and has good contrast, an effective global statistical priors of IWE.

Event-to-Pose module can predict sequential hand MANO parameters $\{ \boldsymbol{\varphi}_{t_n}| n=1, ..., N\}$. 
For each neighboring pair of parameters $\{\boldsymbol{\varphi}_{t_{n-1}}$, $\boldsymbol{\varphi}_{t_{n}}\}$, we can compute the mesh flow $F_n$, and warp the events in $[t_{n-1},t_{n}]$ to get the image with warped events $\text{IWE}_{n, \hat{t}}$ at timestamp $\hat{t}$ as follows:
\begin{equation}
    \text{IWE}_{n, \hat{t}} = \text{Warp}(E_{t_{n-1}, t_n} ;F_n; \hat{t}),
\end{equation}
where $\text{Warp}(\cdot,\cdot,\cdot)$ is composed of two steps: 
We adopt commonly-used flow-constancy hypothesis~\cite{Lucas81flow}, transfer each event $e_i$ in $E_{t_{n-1}, t_n}$ to target time $\hat{t_i}$, and get the warped event $\hat{e_i}$ in new position:
\begin{equation}
    \hat{\boldsymbol{x}_i} = \boldsymbol{x}_i + F_{n}(\boldsymbol{x}_i) \cdot (\hat{t_i} - t_i).
\end{equation}
We warp events form the IWE by counting events at pixel $\boldsymbol{x}$:
\begin{equation}
    \text{IWE}_{n, \hat{t}}(\boldsymbol{x}, p) = \sum_{}\delta(\boldsymbol{x}-\hat{\boldsymbol{x}_i})\delta(p-p_i),
\end{equation}
where each polarity $p$ corresponds to one channel of the IWE.

Then we use the focus loss functions using variance measure in \cite{gallego2019focus} to evaluate how well events are aligned along the hand flow, \ie, the contrast of the image with warped event, as follow:
\begin{equation}
    \text{Var}(I) = \frac{1}{N_I}\sum_{i, j}{(h_{ij}-\mu_I)}^2,
\end{equation}
where $N_I$ is the number of pixels of image $I$, and $\mu_I = \frac{1}{N_I}\sum_{i, j}h_{ij}$ is the mean value of image $I$.

The CM loss is finally defined by computing the contrasts in both forward and backward direction as follows:
\begin{equation}
    \loss_{\text{CM}} = - \sum_{n}(\text{Var}(\text{IWE}_{n, t_{n}}) + \text{Var}(\text{IWE}_{n, t_{n-1}})),
\end{equation}
where $\text{IWE}_{n, t_{n}}$ and $\text{IWE}_{n, t_{n-1}}$ are the forward and backward $\text{IWEs}$ at timestamp $t_{n}$ and $t_{n-1}$, respectively.

\vspace{1mm}
\noindent\textbf{Hand-edge Loss.}
As shown in \Fref{Pose-to-IWE}, IWE should demonstrate sharp hand edges, and this inspires us to design an edge loss to enforce the alignment of the projection of predicted hand shape and the edge of IWE. Intuitively, such a constraint can bring a benefit that the local details of the reconstructed hand shape are well consistent to the event observation.
However, it is challenging to find reliable correspondences of events and hand mesh vertices due to two issues: 1) Events are not aligned at the same time with the hand shape and 2) fingers will gather together in optimization because an event may have several corresponding vertices. 
To solve the above misalignment issues, we first warp the events in $[t_{n-1},t_{n}]$ by mesh flow and get $\text{IWE}$ at timestamp $t_n$, then find the correspondence between pixels of IWE and hand mesh vertices by searching the proper hand vertex for each IWE pixel.
In order to reduce computation cost and make our method robust to noisy events, we only select $M$ IWE key pixels with the most accumulated events to calculate the edge loss. 
For each key pixel $\text{IWE}(x_i, y_i)$, we get a candidate set of hand mesh vertices $P_{i}$ that can be projected within a local neighborhood of the pixel $(x_i, y_i)$ (\eg, $12\times12$). For these vertices, we select a candidate vertex using both orientation weight $w_{j}^{o}$ and motion weight $w_j^m$.   

Since hand mesh vertices with the normal vectors perpendicular to the camera ray will generate more events, we design the orientation weight to each vertex $\mathbf{v}_j$ as follows:
\begin{equation}
    w_{j}^{o} = b_{\text{orient}} - \|\cos(\mathbf{n}_j, \mathbf{d}_j)\|,
\end{equation}
where $\mathbf{n}_j$ is the normal vector at $\mathbf{v}_j$, $\mathbf{d}_j$ is the camera ray pointing from the camera center to $\mathbf{v}_j$, constant and $b_{\text{orient}}$ can be any value slightly over 1 (set as 1.2 in our experiments).

We further obverse that vertices with large motion displacements often produce more events, so we design the motion weight of vertex $\mathbf{v}_j$ as follows:
\begin{equation}
    w_j^m = \sqrt{\left\|\pi(\mathbf{v}_{j, t_{n-1}})-\pi(\mathbf{v}_{j, t_n})\right\|^{2}+b_{\text{motion}}},
\end{equation}
where $\pi(\cdot)$ is the projection function, $\mathbf{v}_{j, t_{n-1}}$ is the vertex at timestamp $t_{n-1}$, and $b_{\text{motion}}$ can be any constant slightly over 0 (set as 4 in our experiments).

Next, we design a metric $\text{metric}_{i, j}$ with the orientation weight $w_{j}^{o}$ and the motion weight $w_j^m$, and select the corresponding vertex $\mathbf{v}_{P_i}$ of $\text{IWE}(x_i, y_i)$ with the highest metric value. Such a metric can be defined as:
\begin{equation}
    \text{metric}_{i, j} = \frac{ w_j^o \cdot w_j^m}{\left\|\pi(\mathbf{v}_{j, t_{n}})-\text{IWE}(x_i, y_i)\right\|^{2} + b_d},
\end{equation}
where $b_d=4.0$ is the constant shift of distance.

The \edge{} loss is finally defined as :
\begin{equation}
    \loss_{\text{\edge{}}} = \sum_{n}\frac{\sum_{i}^{M}w_{P_i}^o \cdot w_{P_i}^m \cdot {\left\|\pi(\mathbf{v}_{P_i, t_{n}})-\text{IWE}(x_i, y_i)\right\|}^2}{\sum_{i}^{M}w_{P_i}^o \cdot w_{P_i}^m}.
\end{equation}

\noindent\textbf{Smooth  Loss.} To ensure the smoothness of the predicted hand parameters of event streams, we use a smooth loss $\loss_{\text{smooth}}$:
\begin{align}
    \loss_{\text{smooth}} = \sum_{n}\text{sigmoid}(\text{max}(0, &\lambda_{\beta}{\lVert\hat{\boldsymbol{\beta}}_n - \hat{\boldsymbol{\beta}}_{n-1} \rVert}^{2} + \nonumber \\ &\lambda_{\theta}{\lVert\hat{\boldsymbol{\theta}}_n - \hat{\boldsymbol{\theta}}_{n-1}\rVert}^{2} - b_{\text{smooth}})),
\end{align}
where the sigmoid function is used to avoid the gradients of $\loss_{\text{smooth}}$ to be large, $b_{\text{smooth}}=0.5$ is the empirical margin of the smooth loss, and the hyper-parameters $\lambda_{\beta}$ and $\lambda_{\theta}$ are set to 0.2 and 1.0.

In summary, the predicted results from the Event-to-Pose module are fed into the Pose-to-IWE module. Those results of the unlabeled segments will be constrained by the weakly-supervision loss $\loss_{\text{weakly}}$:
\begin{equation}
    \loss_{\text{weakly}} = \lambda_{\text{CM}}\loss_{\text{CM}} +\lambda_{\text{\edge{}}}\loss_{\text{\edge{}}} + \lambda_{\text{smooth}}\loss_{\text{smooth}}, 
\end{equation}
where $\lambda_{\text{CM}}$, $\lambda_{\text{smooth}}$, and $\lambda_{\text{\edge{}}}$ are loss weights, and they are set to 3, 0.1, 0.2, respectively.

\subsubsection{Training Strategy}
Our training strategy consists of three steps. First, we train FlowNet under supervision with $\loss_{\text{flow}}$ and fix its parameters in the following steps. Then, we train EvHandPose under supervision with $\loss_{\text{super}}$ to get reasonable good parameter initialization. Finally, we train EvHandPose with both labeled data and unlabeled data with $\loss_{\text{all}}$, which contains supervision loss and weakly-supervision loss and can be formulated as follows:
\begin{equation}
    \loss_{\text{all}} = \lambda_{\text{super}}\loss_{\text{super}} + \lambda_{\text{weakly}}\loss_{\text{weakly}},%+ \lambda_{flow}L_{flow}
\end{equation}
where $\lambda_{\text{super}}$ and $ \lambda_{\text{weakly}}$ are weighting factors, and set to 2.0, 1.0.

\begin{table*}[t]
    \caption{Comparison of existing event-based hand pose estimation dataset and EvRealHands. EvRealHands involves multiple challenging real-world scenes. We measure the amount of event data by their sequences' lengths.}
    \begin{center}
    \resizebox{\linewidth}{!}{
    \begin{tabular}{c|ccccccccc}
    \toprule
    \multirow{2}{*}{Dataset}& \multirow{2}{*}{Source} & \multirow{2}{*}{Type} & \multirow{2}{*}{Resolution} &  \multirow{2}{*}{Subjects} & \multirow{2}{*}{Annotation} & \multirow{2}{*}{Frames} & \multicolumn{3}{c}{Scenes} \\
    & & & & &  &  & Strong light & Fast motion & Flash\\
    \cmidrule(r){1-10}
    NYU~\cite{Tompson14NYU} & Depth & Real & $640\times480$ & 2 & Tracking  & 243 K & None & None & None\\
    STB~\cite{Zhang16STB} & RGB + Depth & Real & $640\times480$ & 1 & Human  & 36 K & None & None & None\\
    BigHand2.2M~\cite{Yuan17bighand} & Depth & Real & $640\times480$ & 10 & Marker & 2.2 M & None & None & None\\
    FreiHAND~\cite{zimmermann2019freihand} & RGB & Real &  $224\times224$ & 32  & Semi-automatic & 134 K & None & None & None\\
    InterHand2.6M~\cite{moon2020interhand2} & RGB & Real & $512\times334$ & 27 & Human+machine & 2.6 M & None & None & None\\
    ContactPose~\cite{Brahmbhatt2020ContactPose} & RGB + Depth & Real & $1920\times1080$ & 50 & Human+machine & 2.9 M & None & None & None\\
    \cmidrule(r){1-10}
    EventHands~\cite{rudnev2021eventhands} & Event & Synthetic & $240\times180$ & 1 & Synthetic & 1.24 M & None & 12.6 s & None\\
    \textbf{EvRealHands} & \textbf{Event + RGB*} & \textbf{Real} & $346\times260$ & 10 & Human+machine & 425 K & \textbf{558.5 s} & \textbf{122.4 s} & \textbf{316.7 s}\\
    \bottomrule
    \end{tabular}}
    \end{center}
    \label{dataset comparison}
\end{table*}

\begin{figure}[t]
  \centering
  \subfloat[Multi-camera system]{\includegraphics[width=\linewidth]{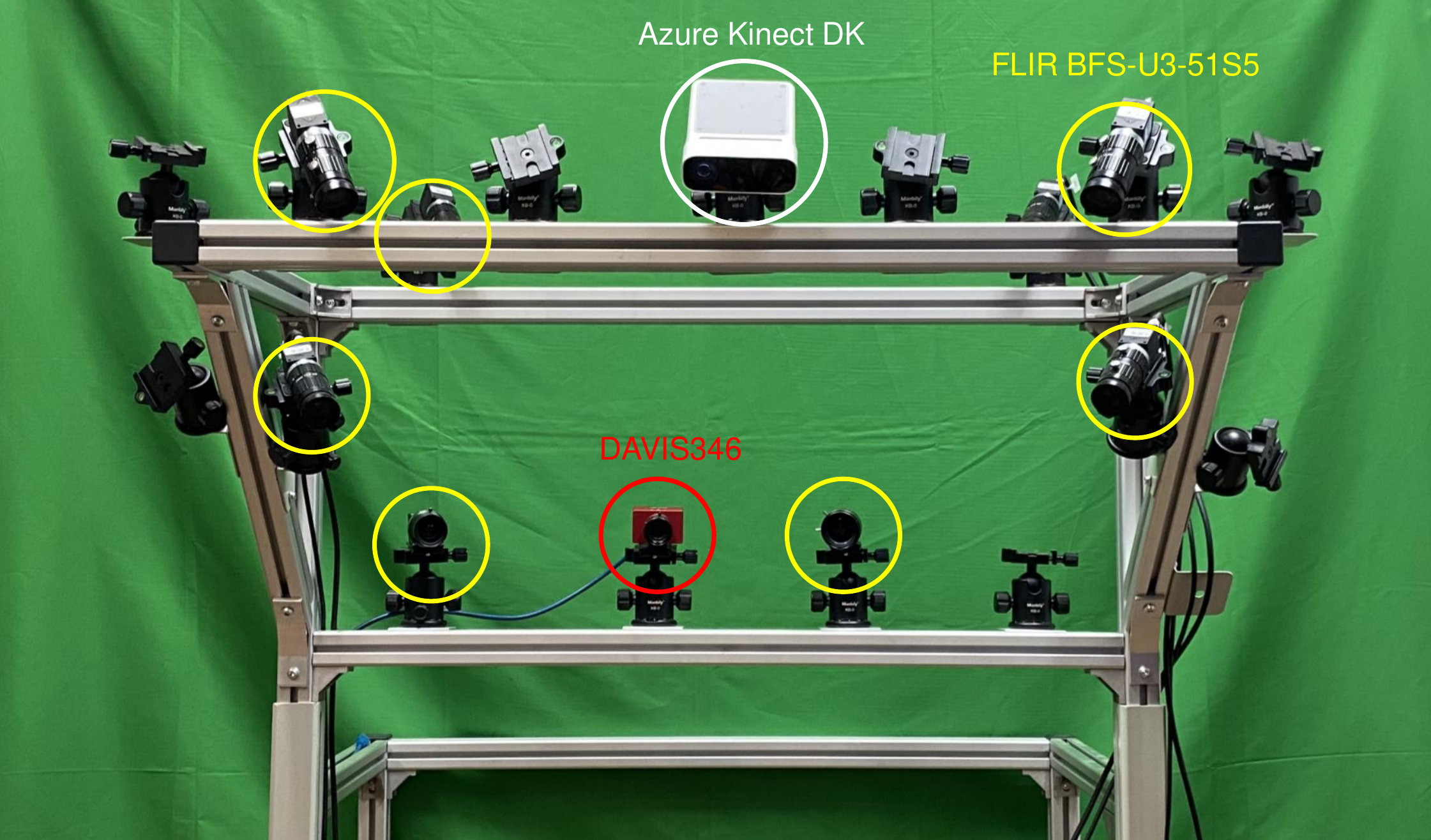}}\\
  \subfloat[DAVIS346 \& FLIR]{\rotatebox{90}{\includegraphics[width=3.2cm]{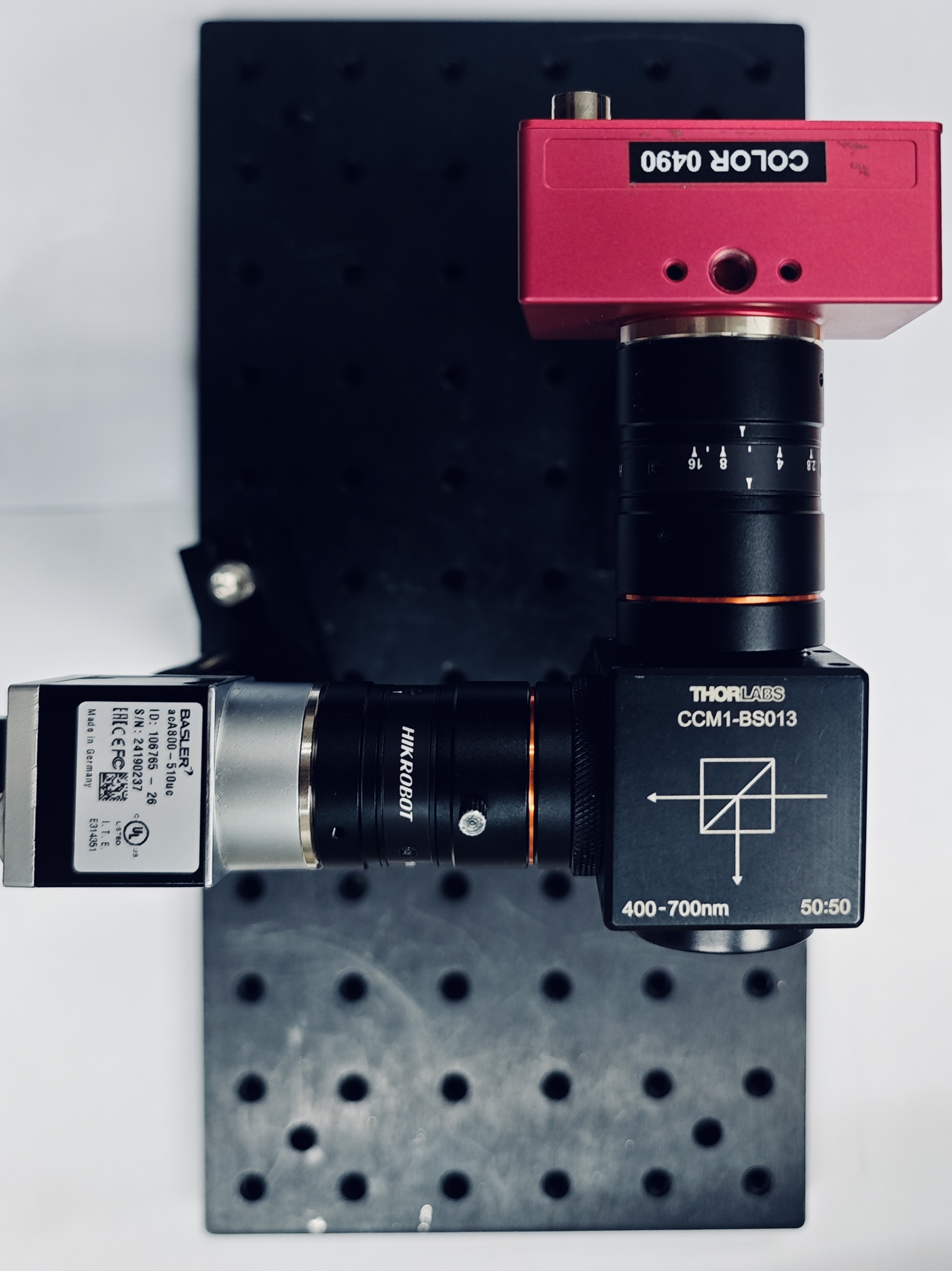}}}
    \subfloat[PROPHESEE \& FLIR]{\rotatebox{90}{\includegraphics[width=3.2cm]{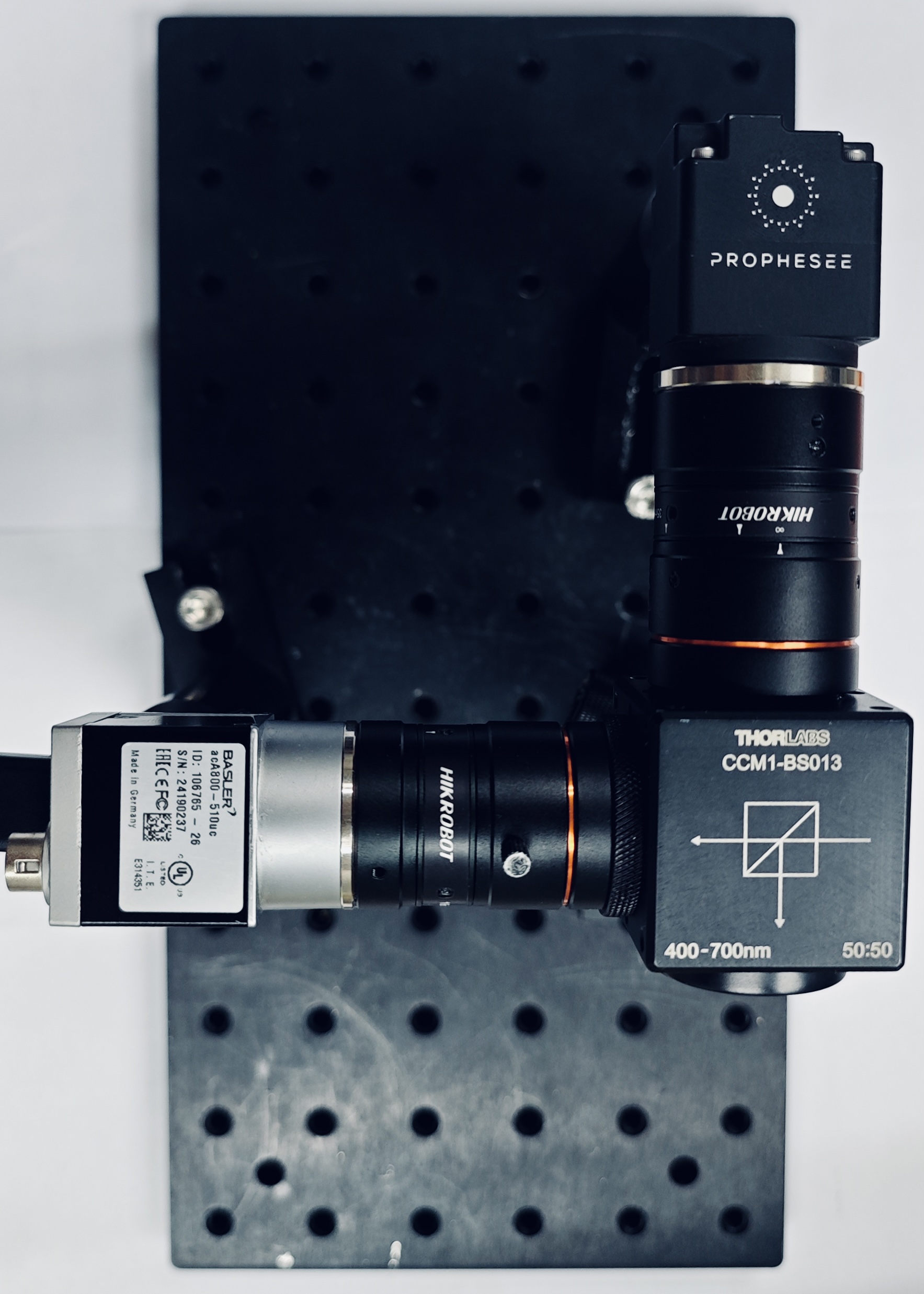}}}
  \caption{Capture system for creating EvRealHands dataset. In the multi-camera system (a), seven FLIR BFS-U3-51S5 cameras (yellow) are synchronized to capture multi-view RGB images and the event camera DAVIS346 (red) captures the event streams in indoor scenes. We also attach an Azure Kinect DK (white) as an auxiliary camera to use the depth image only for accurate  geometric calibration of the camera system.
    In the hybrid camera system for capturing outdoor sequences, we use the beam-splitter settings composed of the RGB camera FLIR BFS-U3-51S5 and the event camera, DAVIS346 Mono in (b) or PROPHESEE GEN 4.0 in (c).
}
  \label{multi-camera systems}
\end{figure}

\section{Dataset}\label{sec: Dataset}
We present the first real-world event-based hand pose dataset EvRealHands.
Our dataset contains event streams and RGB image with annotations of 3D hand pose and hand shape parameter, and it contains hand poses of commonly used hand gestures. EvRealHands is collected under common scenarios (\eg, normal lighting and hand movements) and challenging scenarios such as strong light, flash and fast motion scenes, indoors and outdoors, which are overlooked in previous hand pose datasets as shown in \Tref{dataset comparison}. Please refer to Appendix B in the supplemental material for information about detailed attributes, adequacy of data, scanned hand meshes, and hand gesture visualization and annotation.

\subsection{Data Acquisition}

\vspace{1mm}
\noindent\textbf{Capture System Setup.}
For indoor scenes, we employ a multi-camera fashion similar to \cite{simon2017hand} and \cite{han2020megatrack}.
As shown in \Fref{multi-camera systems} (a), the multi-camera system consists of 7 RGB cameras (FLIR, 2660$\times$2300 pixels) and an event camera (DAVIS346, 346$\times$260 pixels).
We synchronize all the cameras with an external TTL signal of 15 Hz. 
For calibration, we use a moving calibration chessboard and calculate the intrinsic and extrinsic parameters following \cite{heikkila1997four} with RGB frames from FLIR, APS frames from DAVIS346 and depth frames from an auxiliary RGB-D camera (Azure Kinect DK). The average pixel root mean square error on all the images in calibration reaches 0.184.
Since building a multi-camera system outdoor faces several challenges, such as power supply, we build a hybrid camera system in a compact form, as shown in \Fref{multi-camera systems} (b) and (c), to collect outdoor sequences for evaluation.
  The system is composed of an event camera (DAVIS346 Mono or PROPHESEE GEN 4.0) and an RGB camera (FLIR BFS-U3-51S5) via a beam-splitter (Thorlabs CCM1-BS013) with 50\% optical splitting~\cite{duan2021eventzoom}.
  The beam-splitter settings can assure the event streams and RGB images are of approximately the same view towards hand motions.

\begin{figure}[t]
    \centering
    \includegraphics[width=\linewidth]{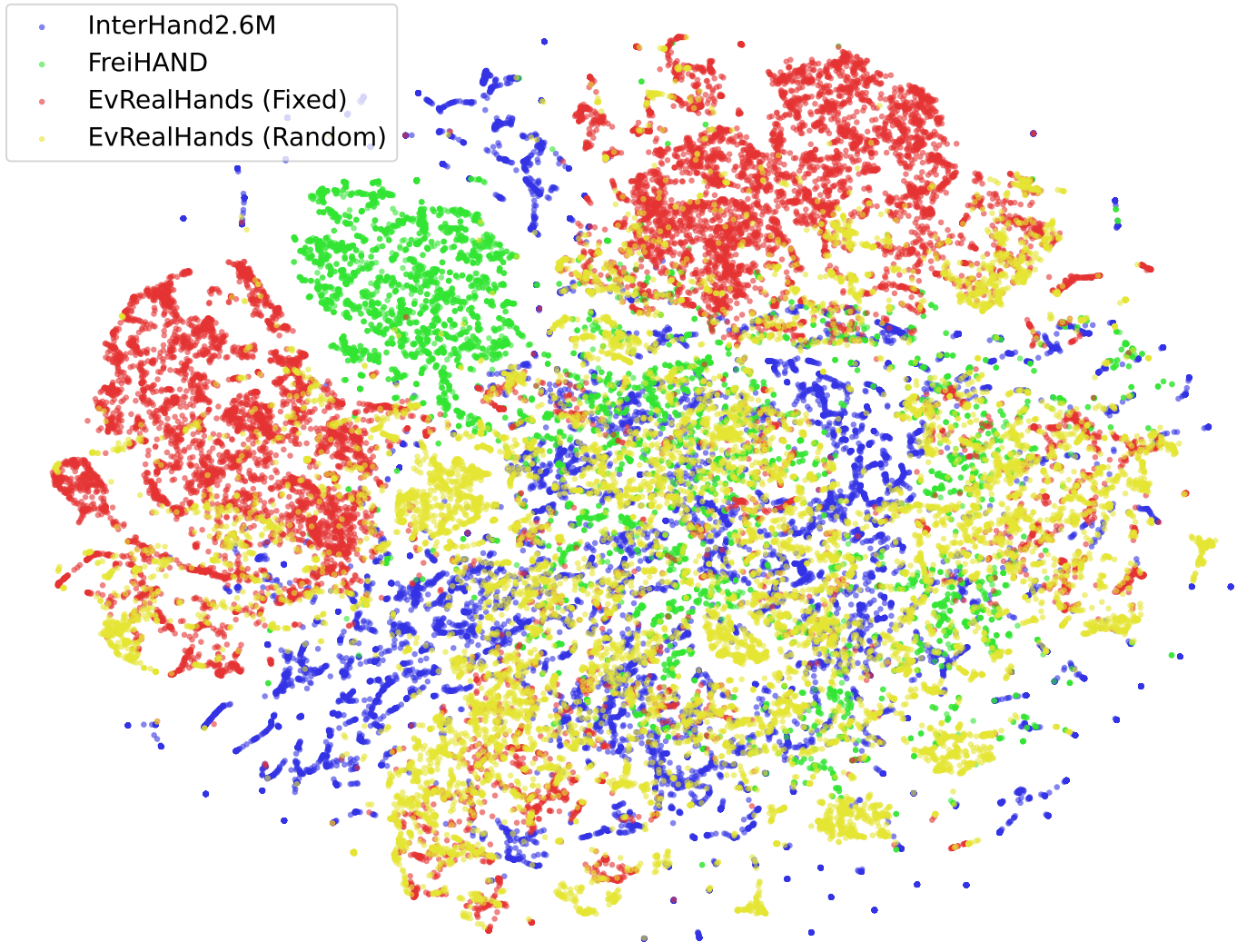}
    \caption{Visualization of hand pose distribution among EvRealHands, FreiHAND~\cite{zimmermann2019freihand}, and InterHand2.6M~\cite{moon2020interhand2} using t-SNE.
    In our visualization setup, each vector fed into t-SNE is the Euler angles of MANO hand pose parameters with 20 dimension as in \cite{moon2020interhand2}.
    Compared with FreiHAND~\cite{zimmermann2019freihand} (green) and InterHand2.6M~\cite{moon2020interhand2} (blue), EvRealHands (red for fixed poses, yellow for random poses) has more diversity in hand poses.
    }
    \label{t-SNE}
\end{figure}

\vspace{1mm}
\noindent \textbf{Hand Gestures.}
We collect 102 sequences from 10 subjects in total with duration ranging from 20 s to 60 s.
For each subject, we collect 3 sequences for each hand under normal lighting, which consists of 15 pre-defined hand poses similar to \cite{de20173d} about 40 s (fixed poses), random hand poses about 20 s (random short), random hand poses about 60 s (random long). As shown in \Fref{t-SNE}, the hand pose variance in random pose sequences is larger than that in fixed pose sequences.

% \begin{table}[t]
%     \caption{Detailed attributes of EvRealHands. The duration of event streams, the quantity of RGB images, and their annotations in each scene are shown below. Annotations [mc] means the 3D annotations are checked or annotated  manually, 
%     while Annotations [all] refer to all the machine and manual annotations. For fast motion sequences, the annotations are 2D keypoints.}
%     \begin{center}
%     \resizebox{\linewidth}{!}{
%     \begin{tabular}{c|ccccccc }
%     \toprule
%     \multirow{2}{*}{Data} & \multicolumn{2}{c}{Normal} & \multicolumn{2}{c}{Strong light} & \multicolumn{2}{c}{Flash} & Fast motion \\
%     & Fixed & Random & Fixed & Random & Fixed & Random & \\
%     \cmidrule(r){1-8}
%     Event streams (s) & 850.7 & 2874.5 & 210.3 & 348.2 & 158.8 & 157.9 & \rev{242.4} \\
%     RGB images  & 79.7 K & 267.6 K & 14.8 K & 28.3 K & 14.9 K & 13.8 K & \rev{7.8 K}\\
%     Annotations [mc] & 11.4 K & 9.7 K & 2.0 K & 1.9 K & 2.1 K & 1.9 K & 0.8 K  \\
%     Annotations [all] & 11.4 K & 37.2 K & 2.0 K & 3.9 K & 2.1 K & 1.9 K & 0.8 K  \\
%     \bottomrule
%     \end{tabular}
%     }
%     \end{center}
%     \label{dataset composition}
% \end{table}

\noindent \textbf{Scenes with Challenging Illuminations and Motions.}
We capture 12 sequences for strong light scenes and 8 sequences for flash scenes from 2 subjects in indoor scenes.
We utilize two glare flashlights with 2000 lumen.
We keep the flashlights on in strong light scenes and make the flashlights flash at a rate of 1 Hz in flash scenes to capture the event streams.
Simultaneously, we keep the annotation RGB cameras from overexposure by reducing their exposure time.
We also collect 10 fast motion sequences from 4 subjects. For outdoor scenes, we collect 6 DAVIS346 sequences and 6 PROPHESEE sequences for evaluation, including 6 scenes with fast motions (3 by DAVIS346 and 3 by PROPHESEE).

\subsection{Annotation}
\subsubsection{3D Joint Annotation}
Our dataset is of single hands and we apply 21 keypoints scheme \cite{zimmermann2017learning} to annotate each hand.
Inspired by Interhand2.6M \cite{moon2020interhand2} and FreiHand \cite{zimmermann2019freihand}, we use multi-view RGB images for 3D annotation and apply a 
two-stage process consisting of machine annotation and human annotation.
First, we use Mediapipe \cite{zhang2020mediapipe} to detect 2D hand keypoints on all the RGB images and triangulate 2D keypoints to obtain 3D keypoints with RANSAC method.
Then we manually verify all the keypoints re-projected by 3D keypoints and select the unqualified views for human annotation. Since our annotations have been checked 
manually, we do not perform a bootstrapping procedure as \cite{simon2017hand}.
Due to the cost of human annotation, manual annotation is only applied to fixed pose and random short sequences and machine annotation is then applied to random long sequences.
For fast motion sequences, we can not get accurate 3D joint annotations due to the severe motion blur in the captured images. To quantitatively evaluate our method, 
we manually annotate the 2D joints on the ECI~\cite{maqueda2018count} frames from the event sequences.
For outdoor sequences, we only annotate 2D bounding boxes for qualitative evaluation.

\subsubsection{Shape Annotation}
We use MANO shape parameter $\boldsymbol{\beta}$ to represent the hand shape.
We first scan the hands of each subject with an Artec Eva 3D Scanner, and get meshes of static hands.
Then we fit a hand MANO model to each mesh to get the prior shape parameter.
For each frame of multiple RGB cameras, we fit hand MANO parameters with prior shape as initialization under the constraints from 2D keypoints, 3D joints, and regularization terms similar to FreiHand \cite{zimmermann2019freihand}.

\section{Experiments}\label{sec: Experiments}

In this section, we will first introduce the evaluation metrics in \Sref{Dataset and Evaluation Metrics}.
Then we will compare our method with baselines on different scenes in \Sref{Comparison with State-of-the-art Methods}.
Finally, we will perform ablation study to demonstrate the claimed effect of each component we propose in \Sref{sec: exper ablation}.
For implementation details, 120 fps inference results, different methods of CM loss and hand-edge loss, and downstream hand gesture recognition task, please refer to Appendix A in the supplemental material and the video.

\subsection{Evaluation Metrics}\label{Dataset and Evaluation Metrics}
% \subsubsection{Training and Evaluation Data}
% We collect sequences from 7 out of 10 subjects as training data, 1 as validation data, and 2 as evaluation data.
% The training data included about 4 minutes of strong light sequences, 2 minutes of flash sequences, and no fast motion data.
% We evaluate the method in indoor (8 sequences under normal scenes, 4 sequences of strong light, 4 sequences of flash, and 2 sequences of fast motion), DAVIS346 outdoor scenes (3 sequences under challenging outdoor illumination, 3 fast motion sequences), and PROPHESEE outdoor scenes.
% % The evaluation data included 8 sequences under normal scenes, 4 sequences of strong light, 4 sequences of flash, and 2 sequences of fast motion.
% In the sequences mentioned above, the left and right hands are equally divided.
% In our training process, we flip the left hand data into the right hand for increasing the amount of data.

% \subsubsection{Evaluation Metrics}
Similar to previous evaluation metrics \cite{rudnev2021eventhands} and \cite{chen2021mobrecon}, we apply root-aligned mean per joint position error (MPJPE, mm), Procrustes Analysis (PA)~\cite{pa75} based MPJPE (PA-MPJPE, mm), and the area under the curve (AUC) of PCK (percentage of correct keypoints) with thresholds ranging from 0 to 10 cm for 3D annotated sequences.
For fast motion sequences, we do not have 3D annotations.
We compute root-aligned 2D-MPJPE (pixels) on keypoints which have been normalized by the palm length (distance from wrist to MCP joint) as evaluation metric.
Since our 2D joint annotations are on the imaging plane of the event camera, the 3D joints predicted by the RGB-based approach are projected onto the imaging plane of the event camera for evaluation.

\subsection{Comparison with State-of-the-art Methods}\label{Comparison with State-of-the-art Methods}

\subsubsection{Baselines}\label{Baselines}
In order to compare the hand pose estimation performance of event camera and RGB camera under challenging scenes, we select MobRecon \cite{chen2021mobrecon} as the baseline, because MobRecon \cite{chen2021mobrecon} can efficiently achieve the state-of-the-art results on popular RGB datasets such as FreiHAND \cite{zimmermann2019freihand}, \etc{} 
Since pixel resolution of the RGB camera is much higher than event camera DAVIS346, the scale of cropped images for MobRecon~\cite{chen2021mobrecon} are much larger than the event representation in EvHandPose. Thus we resize the RGB images to make the hand occupy similar pixels as events for fair comparison.
For MobRrecon \cite{chen2021mobrecon}, we select the data of one viewpoint from our camera acquisition system and divide the dataset according to the above convention for the sake of fairness. 
In strong light scenes, a direct approach is to preprocess the overexposed images using HDR techniques and then perform hand pose estimation using RGB-based methods. For comparing this approach, we choose a representative image HDR method, SingleHDR~\cite{liu2020single}, to preprocess overexposed images, followed by hand pose estimation using MobRecon~\cite{chen2021mobrecon}. This approach is designed to compare the performance between the HDR-preprocessing method and the event-based methods.
To compare with the most relevant solutions, \ie, event-based hand pose estimation methods, we select EventHands~\cite{rudnev2021eventhands} and the method proposed by Jalees \etal \cite{nehvi2021differentiable}.
EventHands~\cite{rudnev2021eventhands} is the first neural event-based 3D hand pose estimation approach, which demonstrates better performance than RGB-based methods on their proposed synthetic dataset.
Jalees \etal's method \cite{nehvi2021differentiable} is the model-based approach for hand pose tracking.
We use their officially released codes for effective evaluation.

\begin{figure*}[t]
  \centering
  \includegraphics[width=\textwidth]{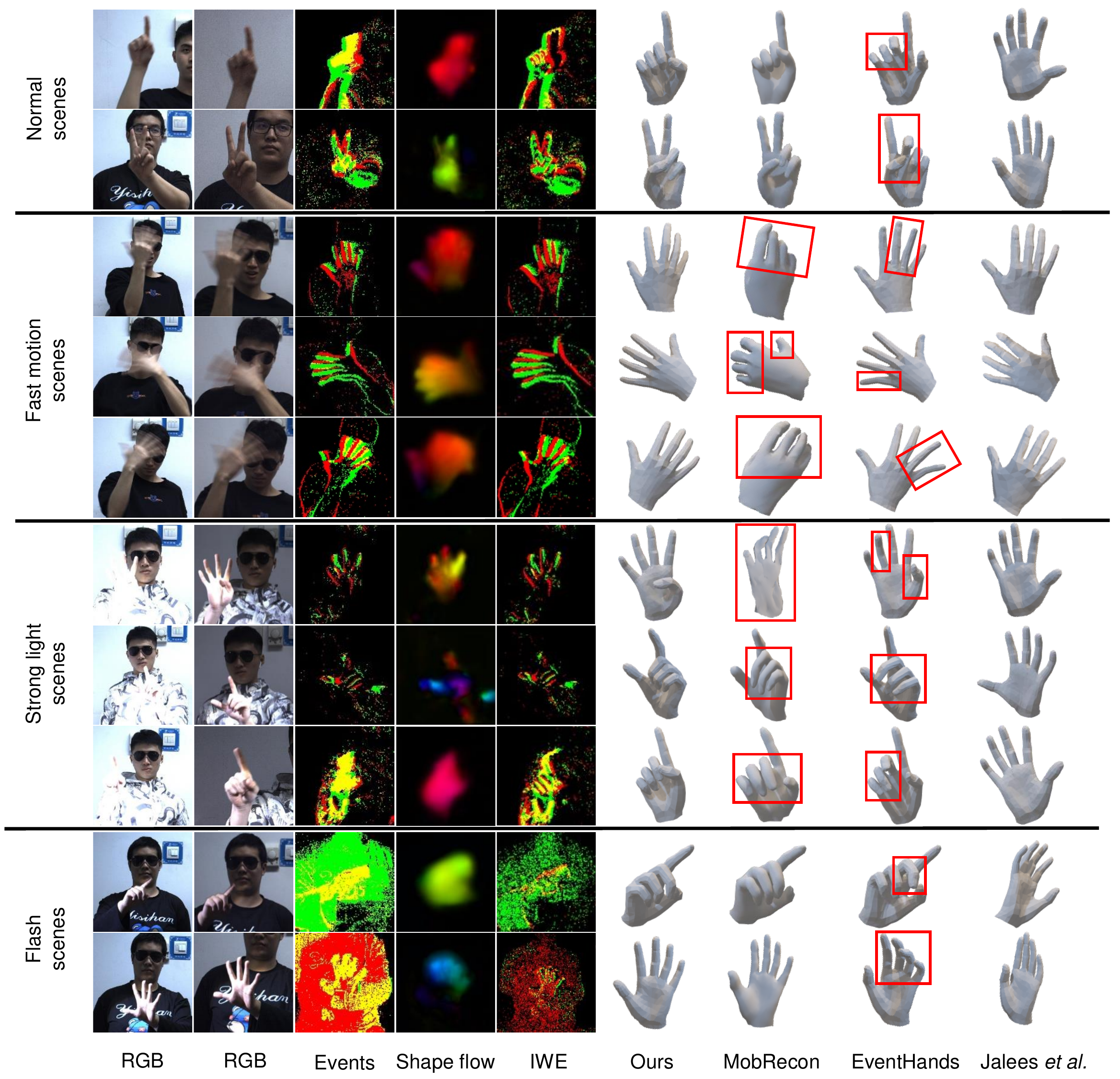}
  \caption{Qualitative analysis. 
  Columns from left to right: Two views of RGB images, events, predicted shape flows, IWEs, poses from EvHandPose (ours), poses from MobRecon \cite{chen2021mobrecon}, poses from EventHands \cite{rudnev2021eventhands}, and poses from the method of Jalees \etal \cite{nehvi2021differentiable}.
  EvHandPose achieves accurate and robust hand pose estimation compared with EventHands \cite{rudnev2021eventhands}, Jalees \etal \cite{nehvi2021differentiable}, and MobRecon \cite{chen2021mobrecon} as highlighted in red boxes.
  }
  \label{qualitative results}
\end{figure*}

\subsubsection{Results}\label{sec: exper results}

In this section, we compare the state-of-the-art color-based and event-based hand pose methods, \ie, MobRecon \cite{chen2021mobrecon}, EventHands \cite{rudnev2021eventhands}, and Jalees \etal \cite{nehvi2021differentiable}, under different typical hand scenarios including normal scenes, strong light scenes, and fast motion scenes.
We show the results of different methods qualitatively and quantitatively in \Fref{qualitative results} and \Tref{quantitative results}, respectively.
Quantitative results show that EvHandPose predicts more accurate hand pose than EventHands~\cite{rudnev2021eventhands} and Jalees \etal \cite{nehvi2021differentiable} by MPJPE over 20 mm lower. This is because our method leverages the temporal and spatial features of events effectively, and addresses the issues of motion ambiguity and sparse annotations.
Compared with MobRecon~\cite{chen2021mobrecon}, our method achieves better performance in challenging scenarios such as strong light and fast motions.

\begin{table*}[t]
    \caption{Quantitative experiments results of different methods and ablation study results.}
    \centering
    \resizebox{\linewidth}{!}{
    \begin{tabular}{c|ccc|ccccccc}
    \toprule
    \multirow{2}{*}{Purposes} & \multirow{2}{*}{Methods} & \multirow{2}{*}{\thead{Training \\ data}}& \multirow{2}{*}{Metrics} &  \multicolumn{2}{c}{Normal} & Fast motion & \multicolumn{2}{c}{Strong light} & \multicolumn{2}{c}{Flash} \\
    & & & & Fixed & Random & Random & Fixed & Random & Fixed & Random \\
    \cmidrule(r){1-11}

    \multirow{10}{*}{\thead{Main \\ results}} & \multirow{2}{*}{EvHandPose} & \multirow{2}{*}{\thead{EvRealHands \\ (events)}} & 3D/2D-MPJPE & 19.82/4.39 & 29.19/6.70 & -/\textbf{8.18} & \textbf{28.98}/\textbf{6.80} & \textbf{33.59}/\textbf{7.64} & 32.80/7.55 & 50.41/10.97\\
    & & & PA-MPJPE & \textbf{10.19} & \textbf{12.72} & - & 11.67 & \textbf{14.59} & 12.85& 16.92 \\
    \cmidrule(r){2-11}

    & \multirow{2}{*}{EventHands~\cite{rudnev2021eventhands}}& \multirow{2}{*}{\thead{EvRealHands \\ (events)}} & 3D/2D-MPJPE & 38.51/10.57 & 55.37/14.86 & -/15.04 & 47.97/13.49 & 57.43/15.32 & 54.30/15.07 & 70.86/18.63\\
    & & & PA-MPJPE & 14.18 & 17.42 & - & 17.41 & 17.78 & 17.18 & 21.48 \\
    \cmidrule(r){2-11}

    & \multirow{2}{*}{Jalees \etal~\cite{nehvi2021differentiable}}& \multirow{2}{*}{\thead{EvRealHands \\ (events)}} & 3D/2D-MPJPE & 77.50/17.42 & 76.15/35.70 & -/33.85 & 73.34/13.13 & 80.18/21.46 & 75.88/32.09 & 89.58/31.92\\
    & & & PA-MPJPE & 27.56 & 17.14 & - & 28.47 & 19.19 & 24.21 & 19.76 \\
    \cmidrule(r){2-11}

    & \multirow{2}{*}{\thead{MobRecon~\cite{chen2021mobrecon}}}& \multirow{2}{*}{\thead{EvRealHands \\ (RGB)}} & 3D/2D-MPJPE & \textbf{19.28}/\textbf{3.82} & \textbf{24.53}/\textbf{5.07} & -/13.75 & 40.54/11.94 & 59.84/16.32 & \textbf{30.30}/\textbf{7.00} & \textbf{44.24}/\textbf{10.26}\\
    & & & PA-MPJPE & 10.90 & 12.85 & - & 19.51 & 25.38 & 13.22 & 17.66 \\
    % \cmidrule(r){2-11}
    % & \multirow{2}{*}{\thead{MobRecon~\cite{chen2021mobrecon}\\ (original)}}& \multirow{2}{*}{\thead{EvRealHands \\ (RGB)}} & MPJPE & 17.39 & 17.34 & 15.60 & 33.14 & 47.19 & 24.94 & 29.66\\
    % & & & PA-MPJPE & 9.99 & 9.83 & - & 17.42 & 21.69 & 11.67 & 13.51 \\
    \cmidrule(r){2-11}
    & \multirow{2}{*}{ \thead{SingleHDR~\cite{liu2020single} \\ + MobRecon~\cite{chen2021mobrecon}}}& \multirow{2}{*}{\thead{EvRealHands \\ (RGB)}} & 3D/2D-MPJPE & 24.80/5.04 & 30.66/6.24 & -/19.24 & 63.96/22.26 & 73.35/22.91 & 35.10/8.98 & 49.55/12.42\\
    & & & PA-MPJPE & 12.54 & 14.02 & - & 25.38 & 27.61 & 14.90 & 18.52 \\
    \cmidrule(r){1-11}

    \multirow{14}{*}{\thead{Ablation: \\ Event \\ representation}} & \multirow{2}{*}{\thead{EvHandPose \\ (w/o flow)}} & \multirow{2}{*}{\thead{EvRealHands \\ (events)}} & 3D-MPJPE & 25.19 & 37.36 & 11.03 & 40.30 & 42.81 & 39.90 & 60.17\\
    & & & PA-MPJPE & 10.70 & 14.77 & - & 13.15 & 17.21 & 12.90 & 19.78\\
    \cmidrule(r){2-11}

    & \multirow{2}{*}{ECI~\cite{maqueda2018count}}& \multirow{2}{*}{\thead{EvRealHands \\ (events)}} & 3D-MPJPE  & 23.13 & 33.76 & 9.81 & 32.16 & 39.80 & 36.29 & 49.54\\
    & & & PA-MPJPE & 10.56 & 14.03 & - & 12.39 & 16.27 & 13.01 & 17.81 \\
    \cmidrule(r){2-11}

    & \multirow{2}{*}{Voxel~\cite{zhu2019unsupervised}}& \multirow{2}{*}{\thead{EvRealHands \\ (events)}} & 3D-MPJPE  & 26.41 & 35.48 & 9.67 & 35.96 & 40.14 & 33.69 & 47.04\\
    & & & PA-MPJPE & 11.31 & 13.99 & - & 13.47 & 16.47 & \textbf{12.32} & 18.34 \\
    \cmidrule(r){2-11}

        & \multirow{2}{*}{TORE~\cite{Baldwin23TORE}}& \multirow{2}{*}{\thead{EvRealHands \\ (events)}} & 3D-MPJPE  & 23.78 & 33.57 & 9.39 & 32.55 & 38.54 & 35.90 & 51.47\\
    & & & PA-MPJPE & 10.83 & 13.46 & - & \textbf{11.43} & 15.47 & 12.45 & 18.18 \\
    \cmidrule(r){2-11}

    & \multirow{2}{*}{\thead{Time \\surface~\cite{lagorce2016hots}}}& \multirow{2}{*}{\thead{EvRealHands \\ (events)}} & 3D-MPJPE  & 31.01 & 40.32 & 9.94 & 38.28 & 44.64 & 40.65 & 58.89\\
    & & & PA-MPJPE & 12.25 & 15.13 & - & 13.14 & 16.87 & 14.66 & 20.31 \\
    \cmidrule(r){2-11}

    & \multirow{2}{*}{EST~\cite{gehrig2019spiketensor}}& \multirow{2}{*}{\thead{EvRealHands \\ (events)}} & 3D-MPJPE  & 24.01 & 36.13 & 11.83 & 30.44 & 39.37 & 35.11 & 55.61\\
    & & & PA-MPJPE & 11.14 & 14.84 & - & 12.71 & 16.68 & 12.63 & 19.47 \\
    \cmidrule(r){2-11}

    & \multirow{2}{*}{\thead{Matrix-\\LSTM~\cite{cannici2020matrixlstm}}}& \multirow{2}{*}{\thead{EvRealHands \\ (events)}} & 3D-MPJPE  & 58.36 & 67.01 & 34.88 & 59.71 & 67.61 & 65.00 & 78.22\\
    & & & PA-MPJPE & 27.12 & 26.73 & - & 28.52 & 26.30 & 25.06 & 30.37 \\
    \cmidrule(r){1-11}

    \multirow{8}{*}{\thead{Ablation: \\ Weakly- \\ supervision \\ framework}} & \multirow{2}{*}{\thead{EvHandPose \\ (w/o \edge{})}} & \multirow{2}{*}{\thead{EvRealHands \\ (events)}} & 3D-MPJPE & 20.44 & 29.34 & 9.00 & 29.13 & 34.58 & 33.52 & 51.60\\
    & & & PA-MPJPE & 10.42 & 12.83 & - & 11.52 & 15.07 & 12.49 & 17.16\\
    \cmidrule(r){2-11}

    & \multirow{2}{*}{\thead{EvHandPose \\ (w/o CM)}}& \multirow{2}{*}{\thead{EvRealHands \\ (events)}} & 3D-MPJPE  & 20.85 & 29.62 & 8.77 & 29.90 & 34.89 & 36.42 & 54.40\\
    & & & PA-MPJPE & 10.43 & 12.73 & - & 12.08 & 14.88 & 13.21 & \textbf{16.69} \\
    \cmidrule(r){2-11}

    & \multirow{2}{*}{\thead{EvHandPose \\ (w/o smooth)}}& \multirow{2}{*}{\thead{EvRealHands \\ (events)}} & 3D-MPJPE  & 20.48 & 29.59 & 8.76 & 29.16 & 33.77 & 34.77 & 52.57\\
    & & & PA-MPJPE & 10.24 & 12.87 & - & 12.23 & 15.17 & 12.75 & 17.28 \\
    \cmidrule(r){2-11}

    & \multirow{2}{*}{\thead{EvHandPose \\ (super)}}& \multirow{2}{*}{\thead{EvRealHands \\ (events)}} & 3D-MPJPE  & 22.76 & 32.91 & 9.05 & 30.16 & 36.26 & 37.26 & 54.91\\
    & & & PA-MPJPE & 10.33 & 13.54 & - & 11.68 & 15.17 & 12.68 & 17.64 \\
    \cmidrule(r){1-11}

    \multirow{8}{*}{\thead{Ablation: \\ Real-synthetic \\ domain \\ gap}} & \multirow{2}{*}{\thead{EvHandPose }} & \multirow{2}{*}{\thead{EventHands~\cite{rudnev2021eventhands} \\ (events)}} & 3D-MPJPE & 70.13 & 77.19 & 32.46 & 79.07 & 82.35 & 92.20 & 99.47\\
    & & & PA-MPJPE & 22.10 & 24.02 & - & 23.15 & 23.11 & 21.75 & 29.31\\
    \cmidrule(r){2-11}

     & \multirow{2}{*}{\thead{EvHandPose}} & \multirow{2}{*}{\thead{InterHand2.6M~\cite{moon2020interhand2} \\ (synth. events)}} & 3D-MPJPE & 51.01 & 65.96 & 21.13 & 56.88 & 68.57 & 61.29 & 80.23\\
    & & & PA-MPJPE & 24.02 & 18.18 & - & 26.00 & 18.35 & 21.50 & 20.61\\
    \cmidrule(r){2-11}

    & \multirow{2}{*}{\thead{NGA~\cite{hu20nga}}} & \multirow{2}{*}{\thead{EvRealHands\\(events \& RGB)}} & 3D-MPJPE & 39.34 & 58.63 & 20.60 & 37.56 & 47.65 & 56.33 & 62.34\\
    & & & PA-MPJPE & 12.68 & 14.93 & -- & 15.19 & 18.15 & 16.24 & 20.18\\
    \cmidrule(r){2-11}

    & \multirow{2}{*}{\thead{EvTransfer~\cite{messikommer22evtransfer}}} & \multirow{2}{*}{\thead{EvRealHands\\(events \& RGB)}} & 3D-MPJPE & 32.65 & 37.64 & 22.42 & 41.65 & 43.44 & 46.33 & 58.90\\
    & & & PA-MPJPE & 12.90 & 15.46 & -- & 14.78 & 17.86 & 15.64 & 19.62\\

    % & & AUC & 1 & 1 & 1 & 1& 1 & 1\\
    % \multirow{2}{*}{Strong light} & \multirow{2}{*}{Random} & MPJPE & 1 & 1 & 1 & 1& 1 \\
    % & AUC & 1 & 1 & 1 & 1& 1 \\
    \bottomrule
    \end{tabular}
    }
    \label{quantitative results}
\end{table*}

\begin{figure*}[t]
  \centering
  \includegraphics[width=\textwidth]{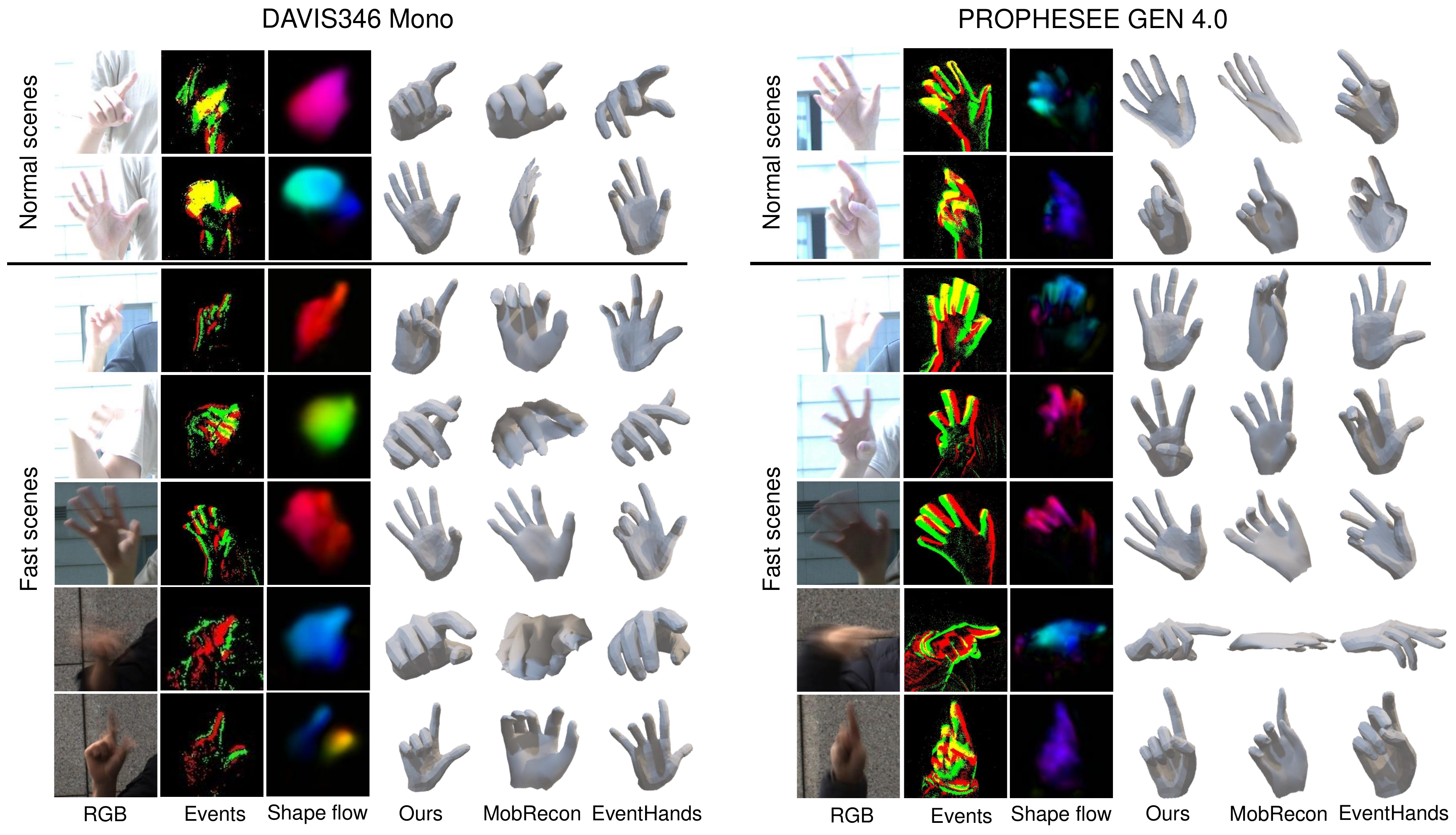}
  \caption{Qualitative results for evaluating the generalization ability to outdoor scenes and another type of event camera.
  Left part is for the results on the event camera DAVIS346 Mono, and the right for PROPHESEE GEN 4.0.
  Columns from left to right: RGB images, events, predicted shape flows, poses from EvHandPose (ours), poses from MobRecon \cite{chen2021mobrecon}, poses from EventHands \cite{rudnev2021eventhands}.
    When facing strong illumination or fast hand motion issues in outdoor scenes, EvHandPose outperforms RGB-based method MobRecon~\cite{chen2021mobrecon} and previous event-based method EventHands~\cite{rudnev2021eventhands} by a large margin.
  }
  \label{outdoor results}
\end{figure*}

\vspace{1mm}
\noindent\textbf{Normal Scenes.}
In normal scenes, our method outperforms EventHands \cite{rudnev2021eventhands} on MPJPE by 19 $\sim$ 26 mm. 
As shown in \Tref{quantitative results}, MPJPE of our method in the fixed pose sequences reaches 19.82 mm.
As stated in Supancic \etal \cite{supancic2015depth}, human annotators could achieve an accuracy about 20 mm for hand, and can easily distinguish two poses if the error exceeds 50 mm.
Therefore, our method can already achieve sufficient hand pose performance for practical gesture interaction applications, as shown in Appendix A in the supplemental material.
The performance of our method is slightly worse (but comparable) than MobRecon \cite{chen2021mobrecon}. This is reasonable due to the richer texture features of RGB images under normal scenes.

% \begin{figure*}[t]
%   \centering
%   \subfloat{\includegraphics[width=\linewidth]{images/main_exper/120 fps/120fps.pdf}}
%   \caption{
%     Illustration of 120 fps event-based hand pose estimation.
%     The two columns on the left and right end are neighboring RGB frames at 15 fps and their predicted hand poses by MobRecon \cite{chen2021mobrecon}.
%     The nine middle columns are the corresponding event frames and predicted poses by EvHandPose at 120 fps.
%     EvHandPose achieves robust pose estimation compared with MobRecon \cite{chen2021mobrecon} at much higher fps.
%   }
%   \label{120 fps}
% \end{figure*}

\vspace{1mm}
\noindent\textbf{Fast Motion Scenes.}
In fast motion sequences, EvHandPose outperforms EventHands with lower 2D-MPJPE.
This is because in training procedure EvHandPose learns the optical flow generated by hand movement on the one hand, and the dynamic time window make EvHandPose more adaptable to sub-segments of different intervals.
Compared to MobRecon \cite{chen2021mobrecon}, EvHandPose has a 2D-MPJPE reduction of 40.5\%. 
The main reasons for worse performance of MobRecon can be summarized as: 
Fast motion will lead to severe hand motion blur effect of RGB camera, and 
it is often infeasible for MobRecon \cite{chen2021mobrecon} or human annotators to extract accurate hand region and keypoints from blurred images.
However, the asynchronous imaging nature of the event camera allows it to capture microsecond motion information and EvHandPose adopts motion representations to record the hand movement precisely as shown in \Fref{qualitative results} and \Fref{outdoor results}. 
The edge representation of the hand is still clear in event frames by setting the event time window to very short intervals (\eg, 5 ms).
The stable edge and motion representations enable EvHandPose to achieve accurate high fps hand pose estimation, although there are no fast motion data in training.
% \Fref{120 fps} shows the estimation of hand pose at 120 fps. 
% Thus, our method can inspire future applications containing fast hand motions.

\vspace{1mm}
\noindent\textbf{Strong Light Scenes.}
% \begin{figure}[t]
%   \centering
%   \subfloat{\includegraphics[width=\linewidth]{images/main_exper/highlight_sequences/26.pdf}}\\
%   \vspace{-3mm}
%   \subfloat{\includegraphics[width=\linewidth]{images/main_exper/highlight_sequences/55.pdf}}\\
%   \caption{Errors on two strong light sequences. Under strong light, EvHandPose has lower prediction error than MobRecon \cite{chen2021mobrecon}.}
%   \label{strong light analysis}
% \end{figure}
As shown in \Tref{quantitative results}, EvHandPose predicts more stable and accurate hand poses than MobRecon \cite{chen2021mobrecon} on strong light sequences with MPJPE 11.6 $\sim$ 26.2 mm lower.
As shown in \Fref{qualitative results}, under strong light, hand overexposure leads to loss of texture information, and it is challenging to estimate hand pose from overexposed images in RGB-based methods.
However, EvHandPose can still extract edge and hand flow feature from the event stream under strong light, thus achieving robust 3D hand pose estimation.
As shown in \Fref{qualitative results}, EvHandPose involves hand motion information in Event-to-Pose module, which makes it more robust and accurate than EventHands \cite{rudnev2021eventhands}.
EvHandPose has a 4.4 $\sim$ 9.2 mm increase in MPJPE in strong light scenes compared to normal scenes. 
One reason is that the distribution of the events generated under strong light is different from that under normal scenes, even though they both record the edge motion. The other reason is that background and shadow of the hand under strong light will also produce a large number of events, making it hard to predict accurate hand poses.
The HDR-preprocessing method does not perform better than MobRecon~\cite{chen2021mobrecon} under strong light scenes. While analyzing the HDR-processed images, we find that the images with overexposure in our dataset are difficult to recover using HDR techniques. Furthermore, HDR techniques tend to distort the edges and texture distribution of the hands, which interferes in hand pose estimation.

\vspace{1mm}
\noindent\textbf{Flash Scenes.}
As shown in \Tref{quantitative results}, EvHandPose outperforms EventHands~\cite{rudnev2021eventhands} and Jalees \etal's method \cite{nehvi2021differentiable} with MPJPE over 20 mm lower.
Because EvHandPose can capture multiple neighboring hand motions from event streams and utilize the temporal information to alleviate the interference of noisy events from the background illumination change.
However, the MPJPE of EvHandPose is slightly lower than MobRecon~\cite{chen2021mobrecon} by 2.5 $\sim$ 6.2 mm.
This is due to some inherent limitations from the DAVIS346 camera we use, which shows a large amount of events erupting in a short time for the scene with a broad dynamic range. This phenomena causes too few effective event observations in a short period of time, which has little useful information for pose estimation. This might be improved by proposing effective temporal filter for hand pose or using more advanced event cameras.

\noindent\textbf{Generalization to Outdoor Scenes.}
In order to compare the generalization ability to outdoor scenes, we train EvHandPose and the baselines in indoor scenes and evaluate them in outdoor scenes.
As shown in \Fref{outdoor results}, EvHandPose predicts more stable hand poses from event streams in outdoor scenes compared with the event-based method~\cite{rudnev2021eventhands}, and MobRecon~\cite{chen2021mobrecon} which faces overexposure and motion blur issues.
The strong generalization performance in outdoor scenes benefits from the effectiveness of our motion representation to retain hand motions and the weakly-supervision framework to learn diverse hand poses from event streams.

\begin{table}[t]
    \caption{Comparison of computational cost, latency, and robustness of different methods.}
    \begin{center}
    \resizebox{\linewidth}{!}{
    \begin{tabular}{c|cccccc}
    \toprule
    \multirow{2}{*}{Method} &  \multicolumn{2}{c}{Computation $\downarrow$} & \multicolumn{2}{c}{Latency $\downarrow$} & \multicolumn{2}{c}{MPJPE$\downarrow$}\\
    & \thead{Params \\ (M)} & \thead{FLOPs \\ (G)} & \thead{Imaging \\ (ms)} & \thead{Inference \\ (ms)} & \thead{Strong light \\(mm)} & \thead{Fast motion \\ (pixel)}\\
    \cmidrule(r){1-7}
    \thead{MobRecon~\cite{chen2021mobrecon}} & \textbf{10.73}  & \textbf{0.42}  & 15 & 5.15 & 53.01 & 13.75 \\
    % \thead{Jalees \etal \cite{nehvi2021differentiable} \\ (500 iter)} & -  & -  & \textless 1 & 5029.60 & XX & 33.86 \\
    % \thead{SingleHDR~\cite{liu2020single} \\+ MobRecon~\cite{chen2021mobrecon}} & 39.76  &  63.88  & 15 & xx & 70.02 & 19.24 \\
    % \cmidrule(r){1-7}
    \thead{Jalees \etal \cite{nehvi2021differentiable}} & -  & -  & \textless 1 & 212.26 & 77.67 & 33.85 \\
    EventHands~\cite{rudnev2021eventhands} & 20.33  & 6.47  & \textless 1 & 3.34 & 53.96 & 15.04 \\
    EvHandPose & 47.36  & 3.44  & \textbf{\textless 1 }& \textbf{1.55} & \textbf{31.91} & \textbf{8.18} \\
    \bottomrule
    \end{tabular}
    }
    \end{center}
    \label{computational cost comparison}
    \vspace{-2em}
\end{table}

\noindent\textbf{Generalization to Another Type of Event Camera.}
In order to demonstrate the generalization ability of EvHandPose to the another type of event camera, we also qualitatively evaluate those methods on outdoor PROPHESEE sequences, which are trained on indoor DAVIS346 sequences.
As shown in \Fref{outdoor results} on the sequences from the outdoor PROPHESEE camera, EvHandPose significantly outperforms MobRecon~\cite{chen2021mobrecon} than those in DAVIS346 outdoor scenes, mainly due to the higher pixel resolution (1280$\times$720 pixels) thus richer events information of PROPHESEE. 
Besides, our motion representation also enables the capability of cross-camera generalization.

\noindent\textbf{Computational Cost, Latency, and Accuracy.}
We conduct a further comparison about computational cost, latency, and accuracy among these methods in \Tref{computational cost comparison}.
Computational cost is measured by model parameters (Params) and floating point operations per inference (FLOPs).
Latency is mainly composed of data imaging process time (Imaging) and model inference time (Inference).
Since Jalees \etal \cite{nehvi2021differentiable}'s method is an optimization-based methods, the number of iteration steps strongly influences the computational cost and latency. We try our best to find the results with best accuracy and least iteration steps (1000 iterations in their official codes).
As shown in \Tref{computational cost comparison}, EvHandPose can achieve 40\% lower MPJPE in strong light and fast motion scenes with the lowest latency, thus achieving robust estimation with high temporal resolution.
This derives from low latency and robustness of the asynchronous imaging mechanism of event cameras compared with frame-based cameras. Also, the simple yet effective motion representations of EvHandPose contribute to the superior performance, too.
Additionally, our backbone is based on the vanilla ResNet~\cite{he2016deep}, and we anticipate future modification on the model structure similar to MobRecon~\cite{chen2021mobrecon} to reduce computational cost and latency.

% \begin{figure*}[t]
%   \centering
%   \captionsetup[subfigure]{labelformat=empty}
%   \subfloat[Motion ambiguity issue]{
%      \centering
%      \includegraphics[height=3.1cm]{images/main_exper/failure case/ambiguity/ambiguity.pdf}
%   }
%     \subfloat[New pose]{
%      \centering
%      \includegraphics[height=3.1cm]{images/main_exper/failure case/new pose/new_pose.pdf}
%   }
%     \subfloat[Flash scenes]{
%      \centering
%      \includegraphics[height=3.1cm]{images/main_exper/failure case/flash/flash.pdf}
%   }
%   \caption{ Three kinds of failure cases: motion ambiguity issue, new pose, and flash scenes. From left to right for each picture combination: RGB view, hand pose from MobRecon \cite{chen2021mobrecon}, event frame, hand pose from EvHandPose.
%   }
%   \label{failure cases}
% \end{figure*}

\subsection{Ablation Study}\label{sec: exper ablation}
We conduct ablation studies on each contributed module of our method.
Specially, we evaluate the effectiveness of the predicted hand flow as event representations for hand pose estimation, the effect of weakly-supervision framework for sparse annotation issue, and the effect of EvRealHands to bridge the real-synthetic domain gap in event-based 3D hand pose estimation.

\vspace{1mm}
\noindent\textbf{Effect of Hand Flow as Event Representation.}
To demonstrate the effectiveness of our hand motion representation, we compare our representation with other widely adopted event representations, including ECI~\cite{maqueda2018count}, time surface~\cite{lagorce2016hots}, voxel~\cite{zhu2019unsupervised}, TORE~\cite{Baldwin23TORE}, EST~\cite{gehrig2019spiketensor}, and Matrix-LSTM~\cite{cannici2020matrixlstm}.
For fair comparison, these methods share the same ResNet34~\cite{he2016deep} backbone with our method.
Quantitative results in \Tref{quantitative results} show that our hand flow representation outperforms them with MPJPE 4 $\sim$ 28 mm lower and PA-MPJPE 1 $\sim$ 17 mm lower.
Removing hand flow representation from EvHandPose (denoted as ``w/o flow" in \Tref{quantitative results}) results in a 4 $\sim$ 11 mm increase in MPJPE.
As qualitative results shown in \Fref{qualitative results}, EvHandPose predicts hand flow from events in its Event-to-Pose module.
On the one hand, shape flow can alleviate motion ambiguity issue because it contains the temporal information of hand movements as shown in the first row of the results under strong light in \Fref{qualitative results}. 
On the other hand, hand shape flow can separate the hand from the background, thus improving the accuracy of estimation.
% Under supervised training, we compare the experimental results with and without hand flow, denote as EvHandPose (super) and EvHandPose (super, w/o flow).
% Under normal and strong light scenes, shape flow provides a 2 $\sim$ 3 mm MPJPE decrease and a 0.02 $\sim$ 0.03 AUC increase.
% On fast motion sequences, shape flow makes EvHandPose outperform MobRecon \cite{chen2021mobrecon}, which preliminarily verify the potential of the event camera on robust hand pose estimation.

\vspace{1mm}
\noindent\textbf{Effect of Weakly-supervision Framework for Spare Annotation.}
We evaluate the influence of the weakly-supervision framework and each loss item on the performance of EvHandPose.
In \Tref{quantitative results}, EvHandPose, EvHandPose (w/o \edge{}), EvHandPose (w/o CM), EvHandPose (w/o smooth), and EvHandPose (super) respectively mean the methods with complete loss, without hand-edge loss, without CM loss, without smooth loss, and supervised method without any weakly-supervision loss.
As shown in \Tref{quantitative results}, compared with the EvHandPose under supervision, our weakly-supervision framework lead to a 1.2 $\sim$ 4.5 mm decrease in 3D-MPJPE in each scene.
According to the experiments, all losses contribute to the performance on the normal, fast motion, and flash scenes.
This improvement derives from the effective usage of dense event streams under sparse 3D annotation.

\vspace{1mm}
\noindent\textbf{Effect of EvRealHands for Real-Synthetic Domain Gap.}
To demonstrate how we solve the real-synthetic domain gap in event-based hand pose estimation, we train EvHandPose on EventHands~\cite{rudnev2021eventhands} or synthetic data from the existing RGB-based dataset Interhand2.6M~\cite{moon2020interhand2} and evaluate the models on our real sequences.
The synthetic data from Interhand2.6M~\cite{moon2020interhand2} are generated by the event simulator v2e~\cite{gehrig20v2e} with single hand RGB sequences of 7 subjects.
The results in \Tref{quantitative results} show that the issues in previous synthetic dataset~\cite{rudnev2021eventhands} (unnatural hand poses and limitations of event simulators) result in a 20 $\sim$ 50 mm increase in MPJPE for EvHandPose.
Although synthetic Interhand2.6M are composed of natural hand poses, there is still a 25 $\sim$ 30 mm MPJPE gap.
We select NGA~\cite{hu20nga} and EvTransfer~\cite{messikommer22evtransfer} as representative domain adaptation methods and use the same ResNet34~\cite{he2016deep} backbone for comparison.
Even though we train them using our EvRealHands dataset, NGA~\cite{hu20nga} and EvTransfer~\cite{messikommer22evtransfer} still underperform our method by 12 $\sim$ 23 mm MPJPE lower.
This performance gap shows that collecting a large-scale real-world event-based hand dataset is beneficial for event-based 3D hand pose estimation.

\subsection{Limitations}
\vspace{1mm}
% \noindent\textbf{Failure Cases.}
%%%DXM:合并一下
Although EvHandPose shows promising hand pose results even in several challenging scenarios, we find that there are still several cases in which our method fails to predict correct hand poses.
Typical cases showing large hand pose errors can be summarized as follows:
\begin{itemize}[\IEEEsetlabelwidth{Z}]
\item Motion ambiguity: Although motion information and recurrent model are adopted to alleviate the motion ambiguity issue, it is not completely removed. It is still challenging to predict hand pose from  event streams if a hand is kept stationary for a long time span;
\item New pose: We observe that MPJPE of our method is 5 $\sim$ 15 mm smaller in the fixed poses than in the random poses. 
%%%DXM: I comment the following sentence
% One reason is that directly regressing the MANO parameters from input data is a high abstract approach. It is challenging for EvHandPose to predict the hand poses that do not appear in the training data. The other reason is 
One possible reason is that the distribution of random poses has greater diversity, leading to a slightly performance drop in random pose sequences. This effect can be reduced by including more data  with diverse hand poses;
% \jianping{better expression?}
% \item Flash scenes: Due to some inherent limitations from the DAVIS346 camera we use, a large amount of events will erupt in a short time for the scene with broad dynamic range. This phenomena causes too few effective event observations in a short period of time, which has little useful information for pose estimation. This might be improved by proposing effective temporal filter for hand pose in the future.
\end{itemize}

\section{Conclusion}
We propose a novel neural network for
event-based monocular 3D hand pose estimation on sparse-labeled event streams.
In order to deal with the asynchronous data format and motion ambiguity issue, we adopt edge and hand flow representations to utilize the spatial-temporal information of event streams effectively.
To tackle the sparse annotation challenge, we design contrast maximization and edge constraints in a weakly-supervision framework.
And we construct the first large-scale real-world dataset for event-based hand pose estimation to address the domain gap between the synthetic and real data.
Experiments on our dataset demonstrate that our method outperforms the recent event-based hand pose method \cite{rudnev2021eventhands} in all cases of EvRealHands testing dataset and RGB-based method \cite{chen2021mobrecon} in challenging scenarios, and can generalize well to outdoor scenes and a different type of event camera.
Although our model has not been trained on fast motion sequences, it can achieve stable hand pose estimation at 120 fps.
Our dataset and method provides a new benchmark and baseline of robust event-based 3D hand pose estimation, and they might inspire related researches in the future, such as hand gesture recognition directly from event streams.
We also hope to improve EvRealHands by capturing hands using more advanced event cameras (such as \cite{chen2019live} and \cite{thomas2020prophesee}) with higher resolution and less noise.

% Although our dataset and method demostrate inspiring results, our dataset and method can be further improved. 
% Our EvRealHands dataset can be enhanced by capturing hands in outdoor scenes or using more advanced event cameras (such as \cite{chen2019live} and \cite{thomas2020prophesee}) with higher resolution and less noise.

% % use section* for acknowledgment
\ifCLASSOPTIONcompsoc
  \section*{Acknowledgments}
\else
  \section*{Acknowledgment}
\fi
This work is supported by the National Key R\&D Program of China (2021ZD0109800) and National Natural Science Foundation of China under Grand No. 62136001, 62088102, and Beijing Natural Science Foundation No. L232028 and L233024.
The authors would like to thank Yiran Zhang, Qida Hao, and Bingxuan Wang for their help to collect and annotate the EvRealHands dataset, Jian Cheng for his help in the downstream hand gesture recognition task, Xinyu Zhou for his help to collect outdoor datasets, and Yixin Yang for her help in the HDR-preprocessing method.

%\IEEEpubidadjcol

\ifCLASSOPTIONcaptionsoff
  \newpage
\fi

{
\bibliographystyle{IEEEtran}
\bibliography{eg}
}

\begin{IEEEbiography}[{\includegraphics[width=1in,height=1.25in,clip,keepaspectratio]{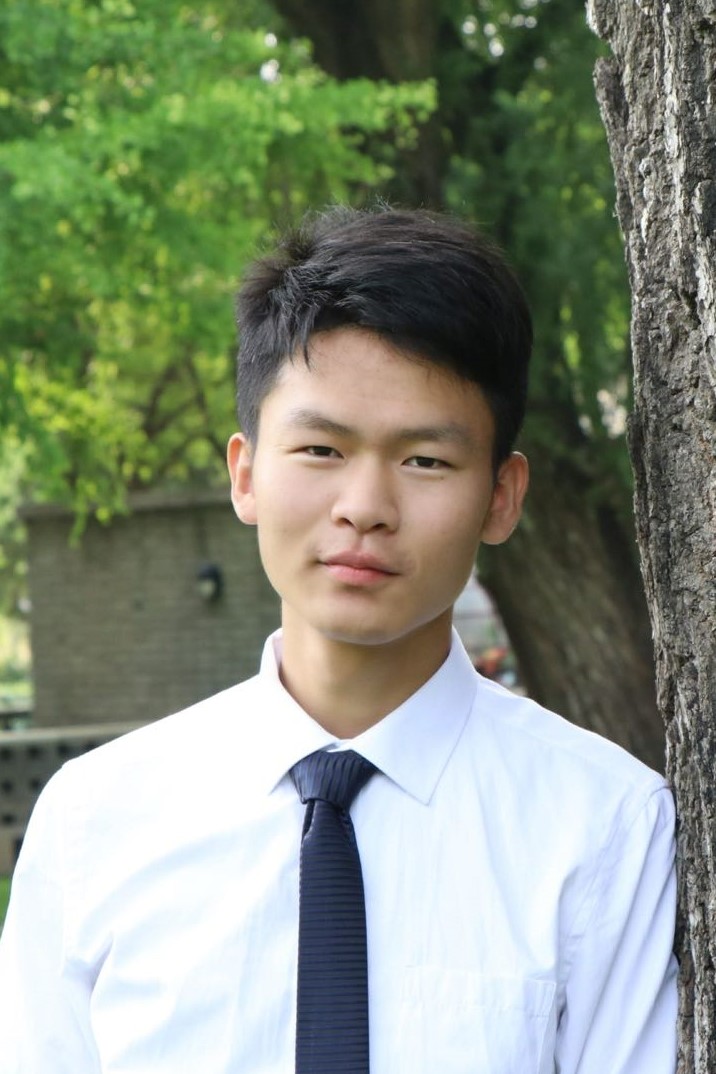}}]{Jianping Jiang}
received his B.Eng. degree from Tsinghua University in 2020 and M.Sci. degree from the School of Computer Science, Peking University. His research interests span mixed reality, 3D avatar.
\end{IEEEbiography}

\vspace{-3.5em}

\begin{IEEEbiography}[{\includegraphics[width=1in,height=1.25in,clip,keepaspectratio]{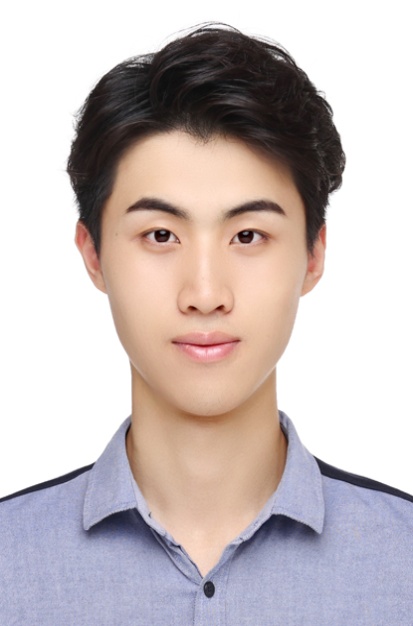}}]{Jiahe Li}
received his B.S. degree in School of Artificial Intelligence, Beijing Normal University. He is currently a master student at the Institute of Software, Chinese Academy.  His research interests include computer vision, 3D reconstruction, human motion tracking and synthesis.
\end{IEEEbiography}

\vspace{-3.5em}

\begin{IEEEbiography}[{\includegraphics[width=1in,height=1.25in,clip,keepaspectratio]{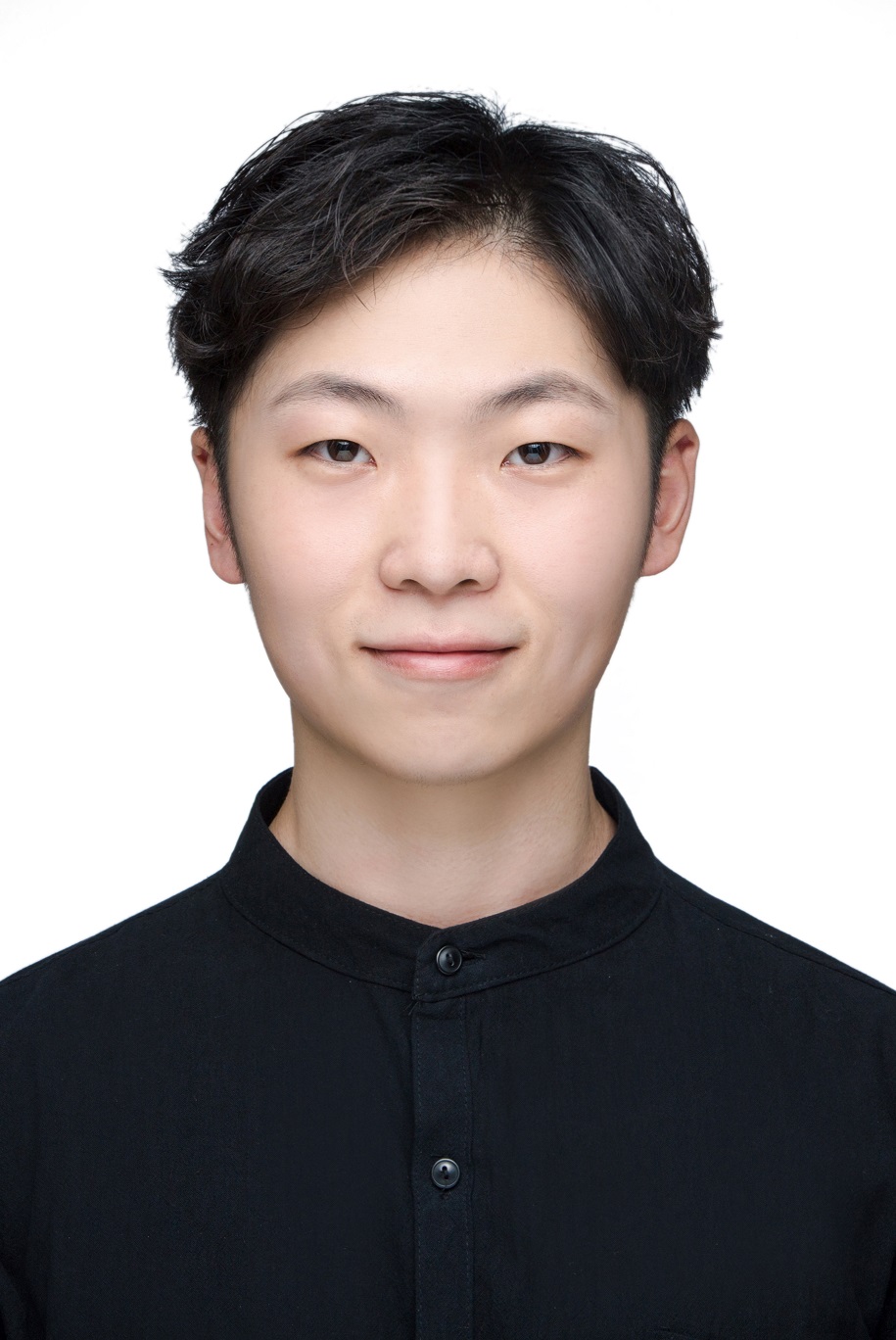}}]{Baowen Zhang}
received his B.Eng. degree from Southeast University, in 2020. He is currently a master graduate student at the Institute of Software, Chinese Academy of Sciences. His main research interest is computer vision.
\end{IEEEbiography}

\vspace{-3.5em}

\begin{IEEEbiography}[{\includegraphics[width=1in,height=1.25in,clip,keepaspectratio]{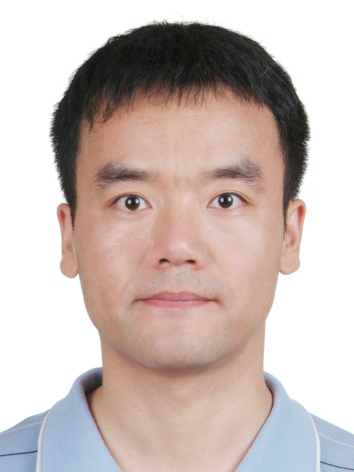}}]{Xiaoming Deng} is currently a  Professor with the Institute of Software, Chinese Academy of Sciences (CAS). He received the BS and MS degrees from Wuhan University, and the PhD degree from the Institute of Automation, CAS. He has been a Research Fellow at the National University of Singapore, and a Postdoctoral Fellow at the Institute of Computing Technology, CAS, respectively. His main research topics are in computer vision, and specifically related to 3D reconstruction, human motion tracking and synthesis, and natural user interfaces.
\end{IEEEbiography}

\vspace{-3.5em}

\begin{IEEEbiography}[{\includegraphics[width=1in,height=1.25in,clip,keepaspectratio]{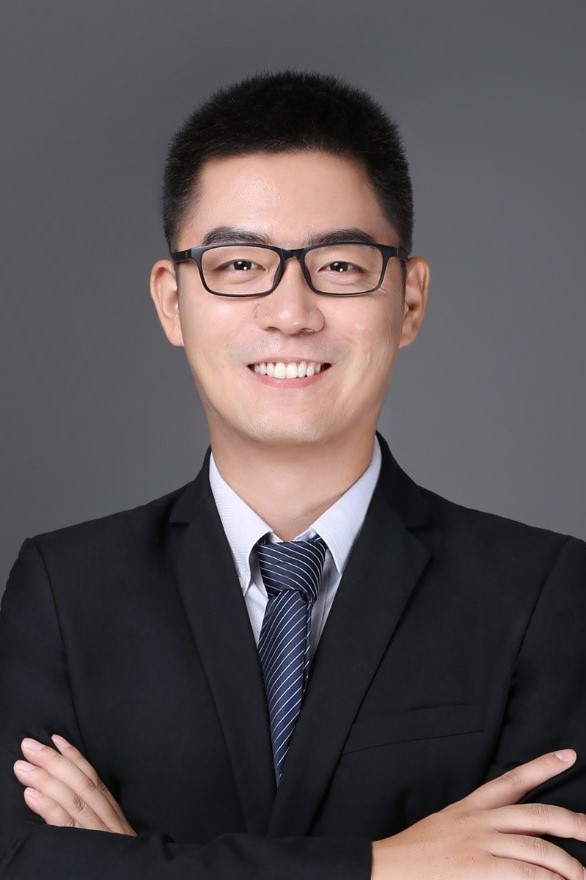}}]{Boxin Shi}
received the BE degree from the Beijing University of Posts and Telecommunications, the ME degree from Peking University, and the PhD degree from the University of Tokyo, in 2007, 2010, and 2013. He is currently a Boya Young Fellow Assistant Professor and Research Professor at Peking University, where he leads the Camera Intelligence Lab. Before joining PKU, he did postdoctoral research with MIT Media Lab, Singapore University of Technology and Design, Nanyang Technological University from 2013 to 2016, and worked as a researcher in the National Institute of Advanced Industrial Science and Technology from 2016 to 2017. His papers were awarded as Best Paper Runner-Up at International Conference on Computational Photography 2015 and selected as Best Papers from ICCV 2015 for IJCV Special Issue. He has served as an associate editor of TPAMI/IJCV and an area chair of CVPR/ICCV/ECCV. He is a senior member of IEEE.
\end{IEEEbiography}

\vfill

\clearpage
\newpage

\input{revision2/supple}

\end{document}

%% file: revision2/supple.tex
\appendices

\section{Additional Experimental Results}

\subsection{Downstream Application: Gesture Recognition}
\label{hand gesture recognition}
To demonstrate the potential of event-based hand pose estimation on downstream tasks, we perform skeleton-based hand gesture recognition~\cite{Zhang20STAGCN} on fixed hand pose sequences (15 types similar to \cite{de20173d}) in EvRealHands.
We train STA-GCN~\cite{Zhang20STAGCN}, a state-of-the-art hand gesture recognition model built on graph convolution network, on ground truth poses and evaluate it on predicted hand poses from EvHandPose and other baselines.
The training and testing splitting is the same as EvHandPose.
As results shown in \Tref{hand gesture recognition}, results from EvHandPose can achieve 20\% accuracy over those from EventHands~\cite{rudnev2021eventhands} and Jalees \etal \cite{nehvi2021differentiable} in strong light and flash scenes. Using predicted hand poses from EvHandPose can improve the gesture recognition accuracy in strong light scenes by 13\% compared with the RGB-based method MobRecon~\cite{chen2021mobrecon}.

\subsection{Implementation Details}\label{Implementation Details}
Since we use temporal information through recurrent model, we need to divide an event stream into $N$ sub-segments.
In our experiment, $N$ is 12, and we utilize the outputs of the last three sub-segments for supervision.
For supervision, the interval of each sub-segment is 66.66 ms, which is the same with the period of annotations.
For weakly-supervision, we only let second last sub-segment have 3D annotation, so that we can dynamically choose the interval of the sub-segment and the model can learn more hand information with different intervals.
For the sub-segment with annotation, we use its ground truth hand pose instead of predicted hand pose in Pose-to-IWE module.
The interval is randomly selected from 10 ms to 100 ms for weakly-supervision. 
Certainly, all the sub-segments have the same duration in a sequence.
For evaluation, we set the interval of each sub-segment to 5 ms for fast motion sequences and 66.66 ms for other sequences.

In order to make the network effectively extract the feature of the hand, we calculate a bounding box for each sub-segment and then crop out the events inside this bounding box.
Obtaining a bounding box for a event sub-segment is different from the RGB image due to the different temporal distributions.
Given 3D joints at some timestamp, we project them onto the image plane to obtain the 2D joints and cover the 2D joints with an exact rectangle. 
We set the square with the center of the rectangle and 1.5 times the length of the longest side of the rectangle as bounding box.
To compute a bounding box for a event sub-segment, we first approximately estimate the 3D joints of the start time and end time of this sub-segment with quadratic interpolation. Then we calculate the bounding boxes of the 3D joints and apply a bounding box that covers the two bounding boxes exactly as the final bounding box.
For fast motion sequences, our machine annotated 3D joints are not very precise. We apply annotated 2D joints interpolation for generating bounding boxes.

We apply data augmentation including scale, rotation, and translation on events and resize the cropped event frames to 128$\times$128.
We use ResNet34 \cite{he2016deep} as the backbone of feature encoders, and apply Adam \cite{kingma2014adam} as the optimizer for all our training procedures.
The learning rate of FlowNet and EvHandPose in supervision is 0.0005 and for EvHandPose in weakly-supervision is 0.0001.
We train EvHandPose on two TITAN RTX 4090 GPUs with batch size 32 for 44 K iterations.

\begin{table}[t]
    % \captionsetup{name=\textcolor{magenta}{TABLE}, labelformat=magentatext, labelsep=magentacolon}
    % \color{magenta}
    \caption{Accuracy (\%) of hand gesture recognition.}
    \begin{center}
    % \resizebox{\linewidth}{!}{
    \begin{tabular}{c|ccc}
    \toprule
    \thead{Input \\ results} &  Normal & Strong light & Flash  \\ 
    \cmidrule(r){1-4}
    
    EventHands~\cite{rudnev2021eventhands} & 68.3 & 60.0 &  33.3\\
    Jalees \etal \cite{nehvi2021differentiable} & 75.0 & 70.0 &  53.3\\
    MobRecon~\cite{chen2021mobrecon}  & \textbf{88.3} & 76.7 &  \textbf{76.7}\\
    EvHandPose  & 83.3 & \textbf{90.0} &  73.3\\
    \bottomrule
    \end{tabular}
    % }
    \end{center}
    \label{hand gesture recognition}
    \vspace{-2em}
\end{table}

\begin{figure*}[t]
  \centering
  % \captionsetup{name=\textcolor{magenta}{Fig.}, labelformat=magentatext, labelsep=magentacolon}
  \subfloat{\includegraphics[width=\linewidth]{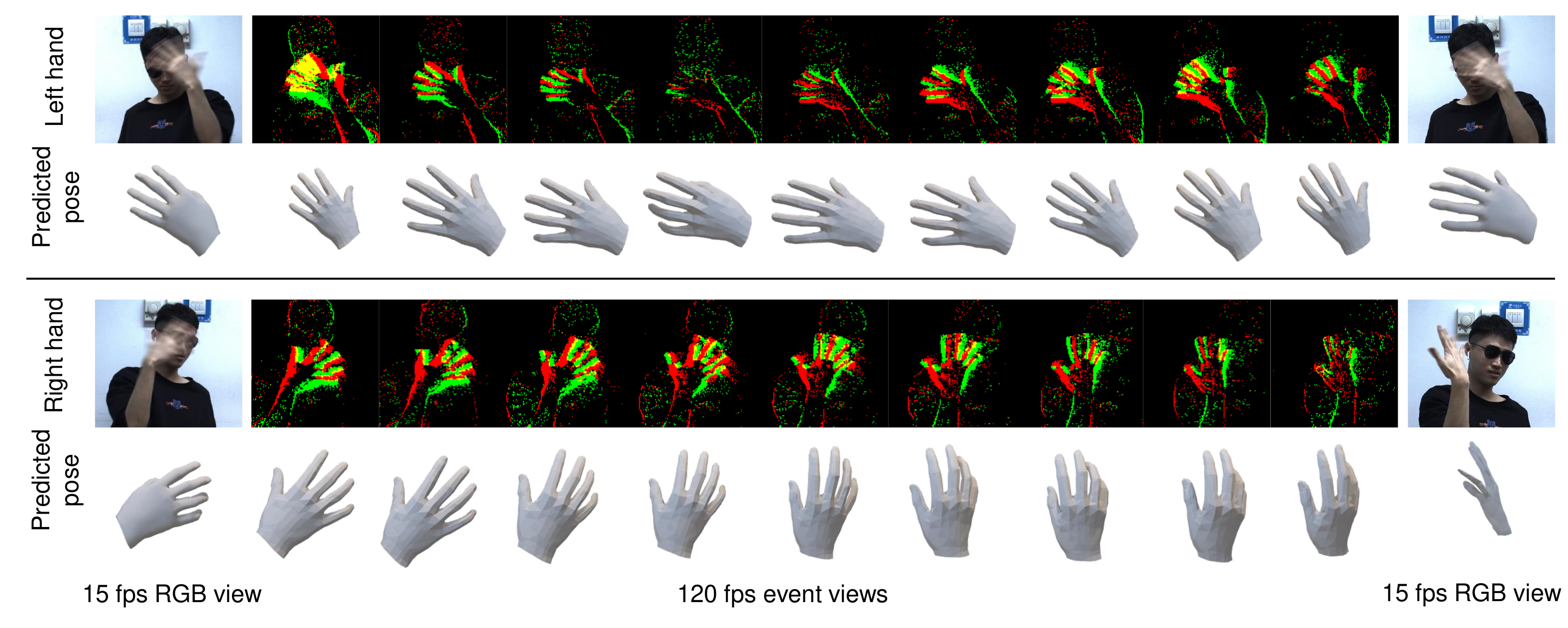}}
  \caption{Illustration of 120 fps event-based hand pose estimation.
    The two columns on the left and right end are neighboring RGB frames at 15 fps and their predicted hand poses by MobRecon \cite{chen2021mobrecon}.
    The nine middle columns are the corresponding event frames and predicted poses by EvHandPose at 120 fps.
    EvHandPose achieves robust pose estimation compared with MobRecon \cite{chen2021mobrecon} at much higher fps.}
  \label{120 fps}
\end{figure*}

\subsection{Training and Evaluation Data}
We collect sequences from 7 out of 10 subjects as training data, 1 as validation data, and 2 as evaluation data.
The training data included about 4 minutes of strong light sequences, 2 minutes of flash sequences, and no fast motion data.
We evaluate the method in indoor (8 sequences under normal scenes, 4 sequences of strong light, 4 sequences of flash, and 2 sequences of fast motion), DAVIS346 outdoor scenes (3 sequences under challenging outdoor illumination, 3 fast motion sequences), and PROPHESEE outdoor scenes.
In the sequences mentioned above, the left and right hands are equally divided.
In our training process, we flip the left hand data into the right hand for increasing the amount of data.

\subsection{Additional Main Results}
\begin{figure*}[t]
    \centering
    \subfloat{\includegraphics[width=0.33\textwidth]{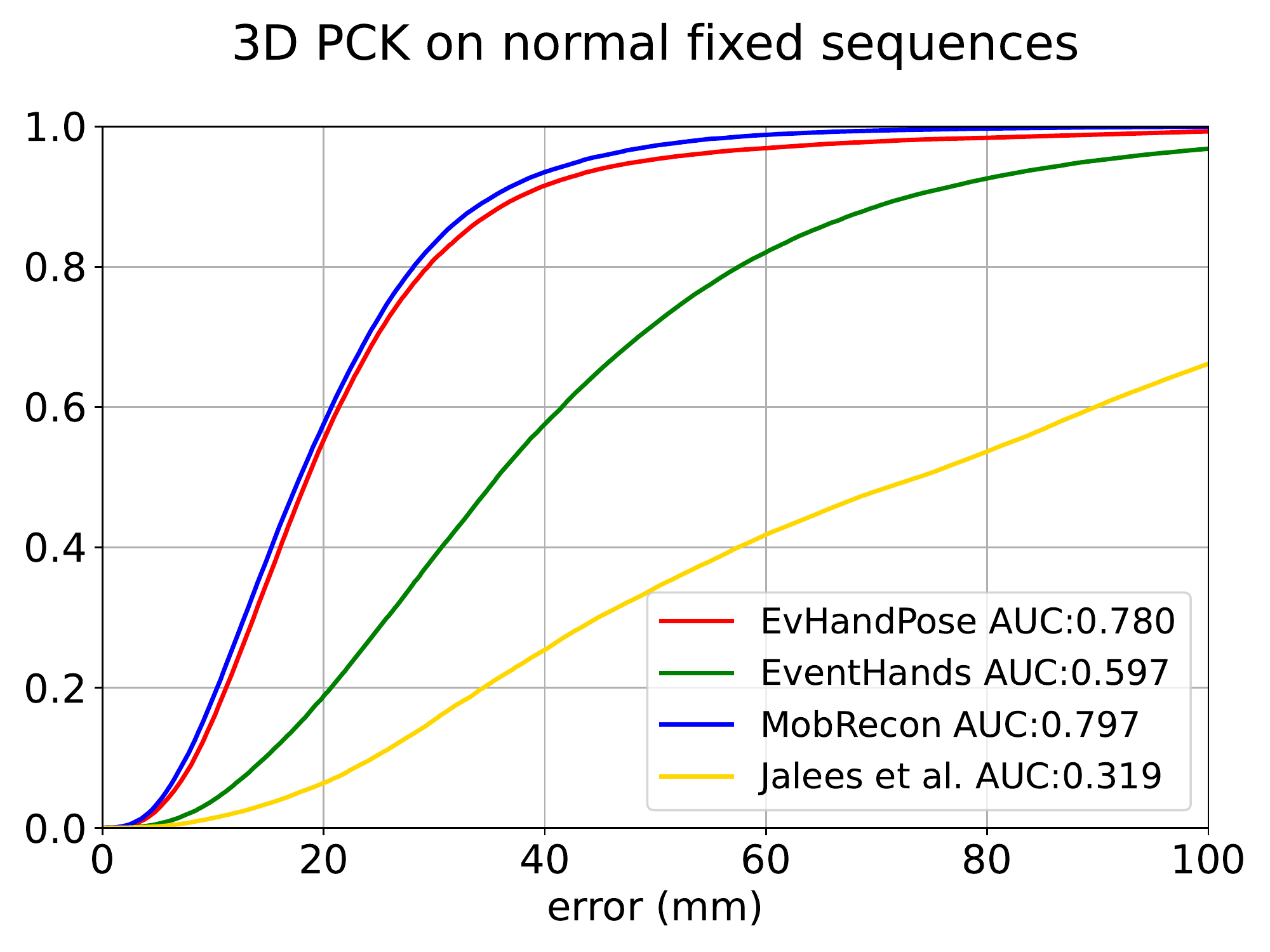}}
    \subfloat{\includegraphics[width=0.33\textwidth]{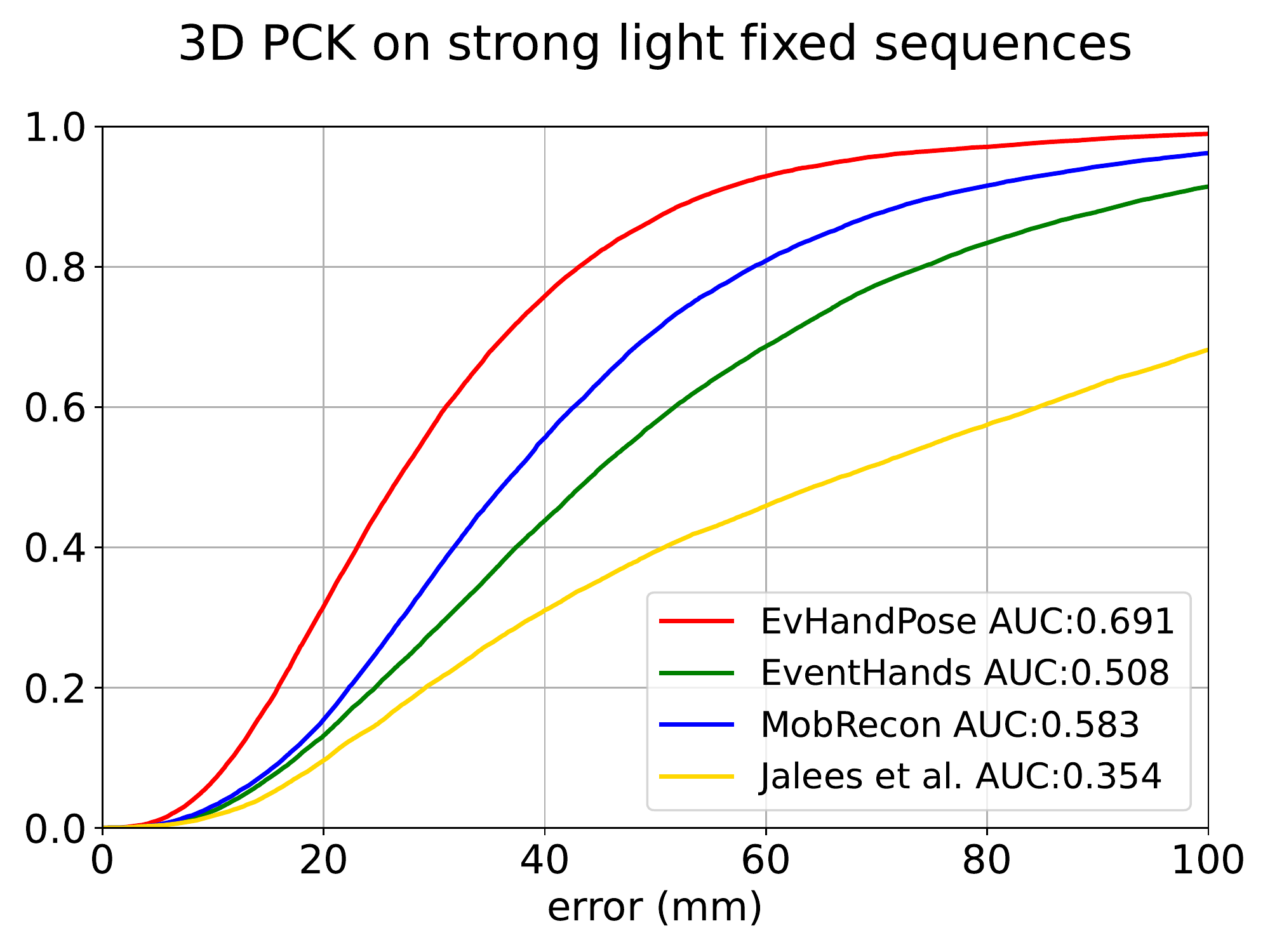}}    \subfloat{\includegraphics[width=0.33\textwidth]{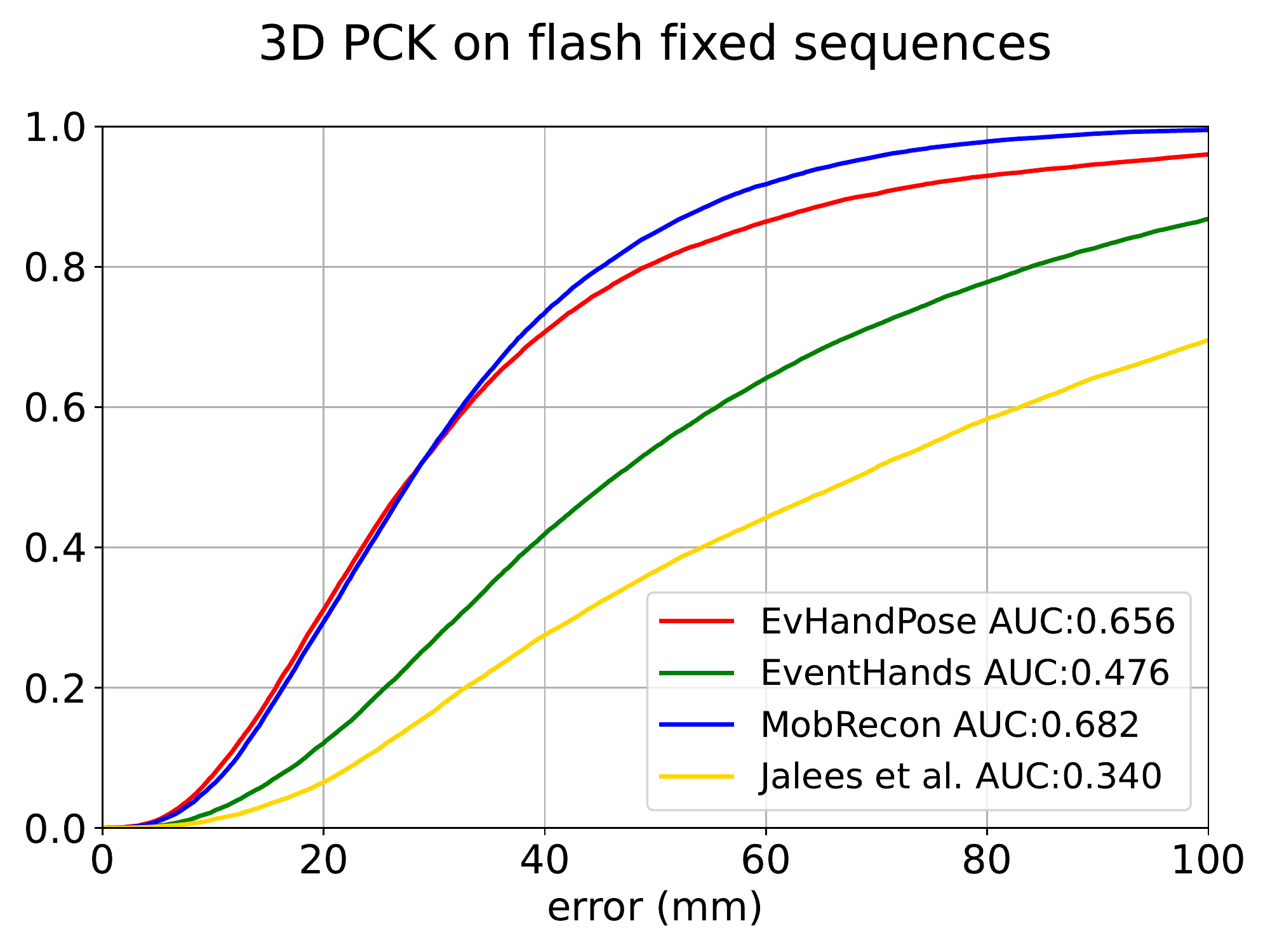}}\\
    \subfloat{\includegraphics[width=0.33\textwidth]{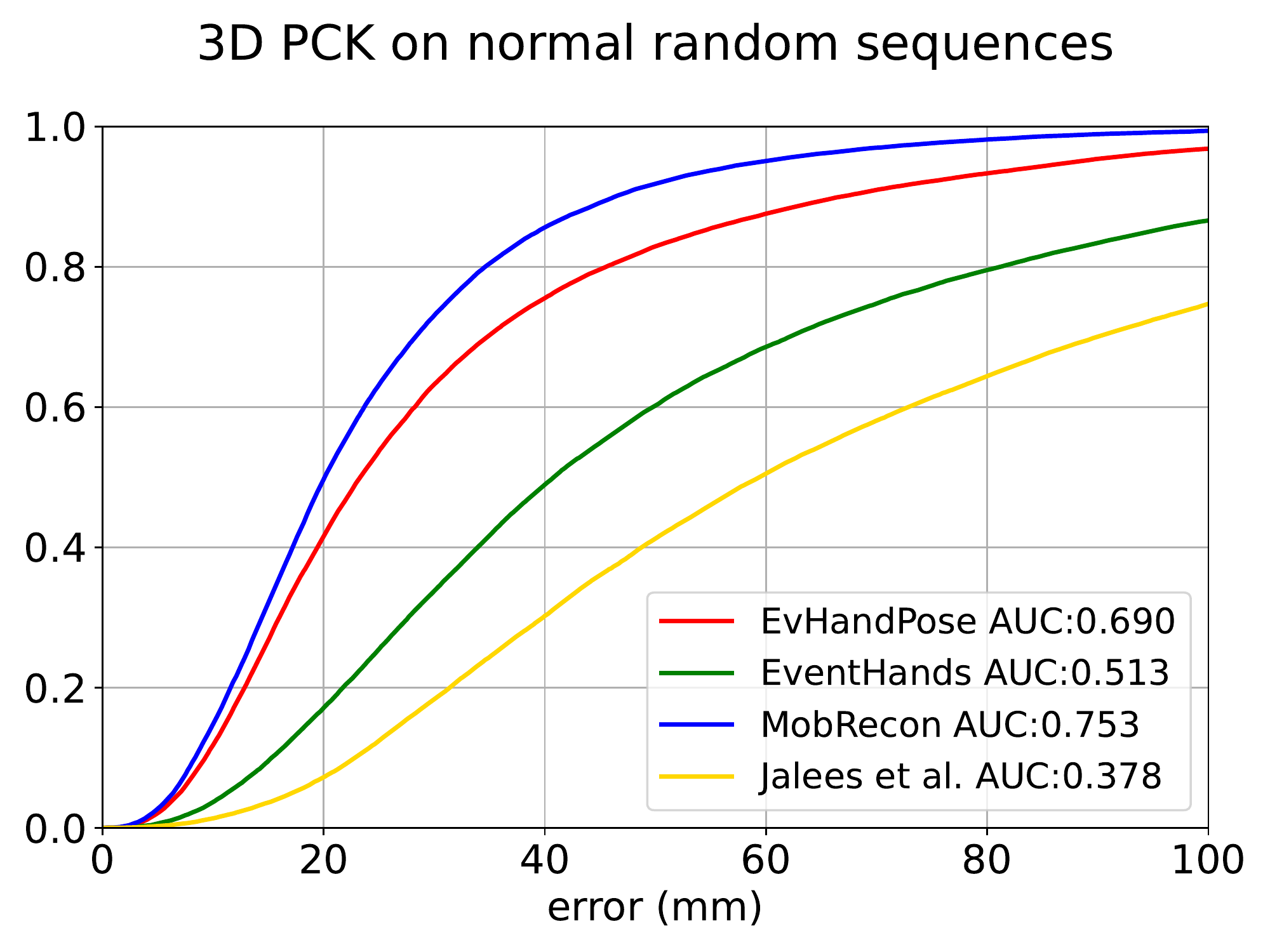}}
    \subfloat{\includegraphics[width=0.33\textwidth]{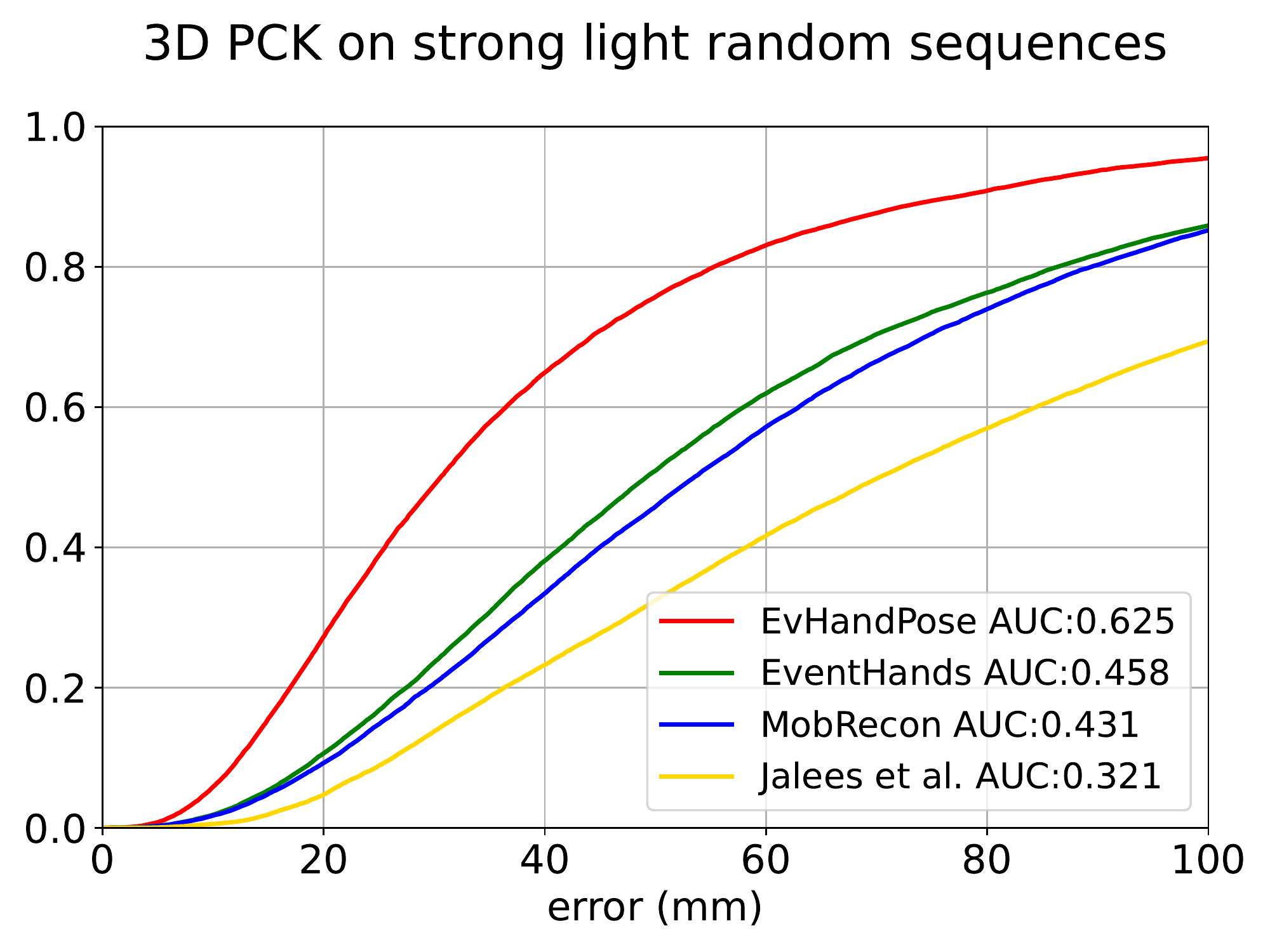}}    \subfloat{\includegraphics[width=0.33\textwidth]{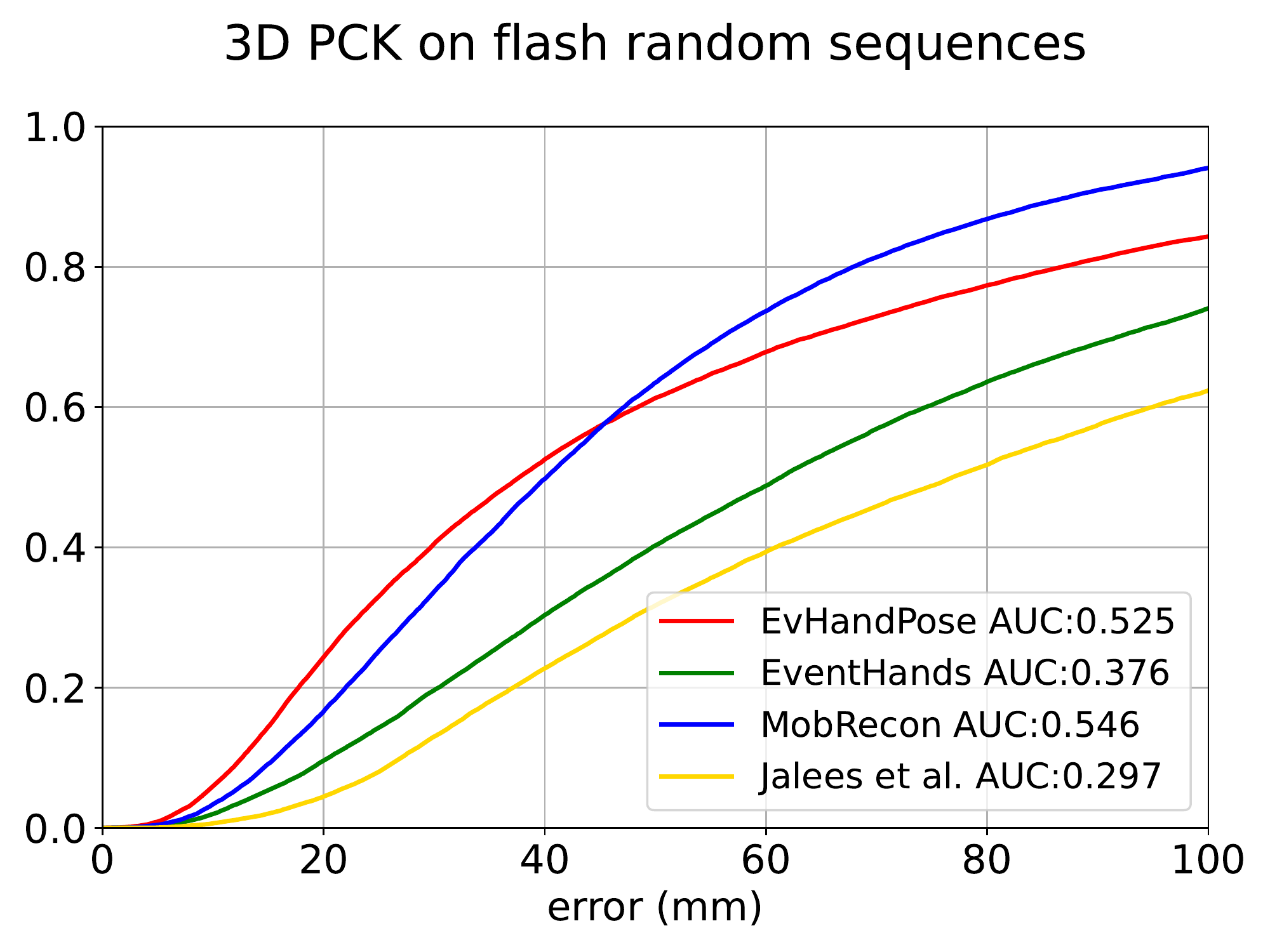}}\\
    \caption{3D PCK curves of different methods on different scenes.}
    \label{AUC curves}
\end{figure*}

\begin{figure}[t]
  \centering
  \subfloat{\includegraphics[width=\linewidth]{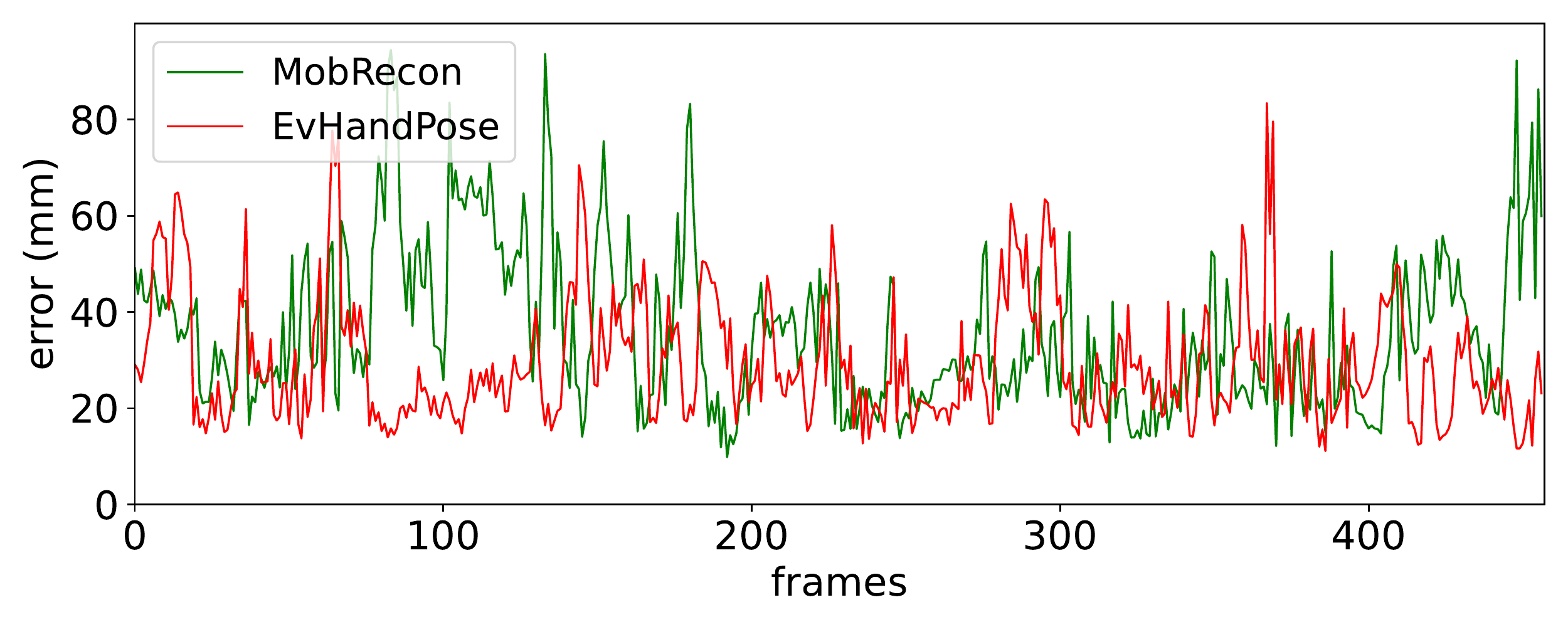}}\\
  \vspace{-3mm}
  \subfloat{\includegraphics[width=\linewidth]{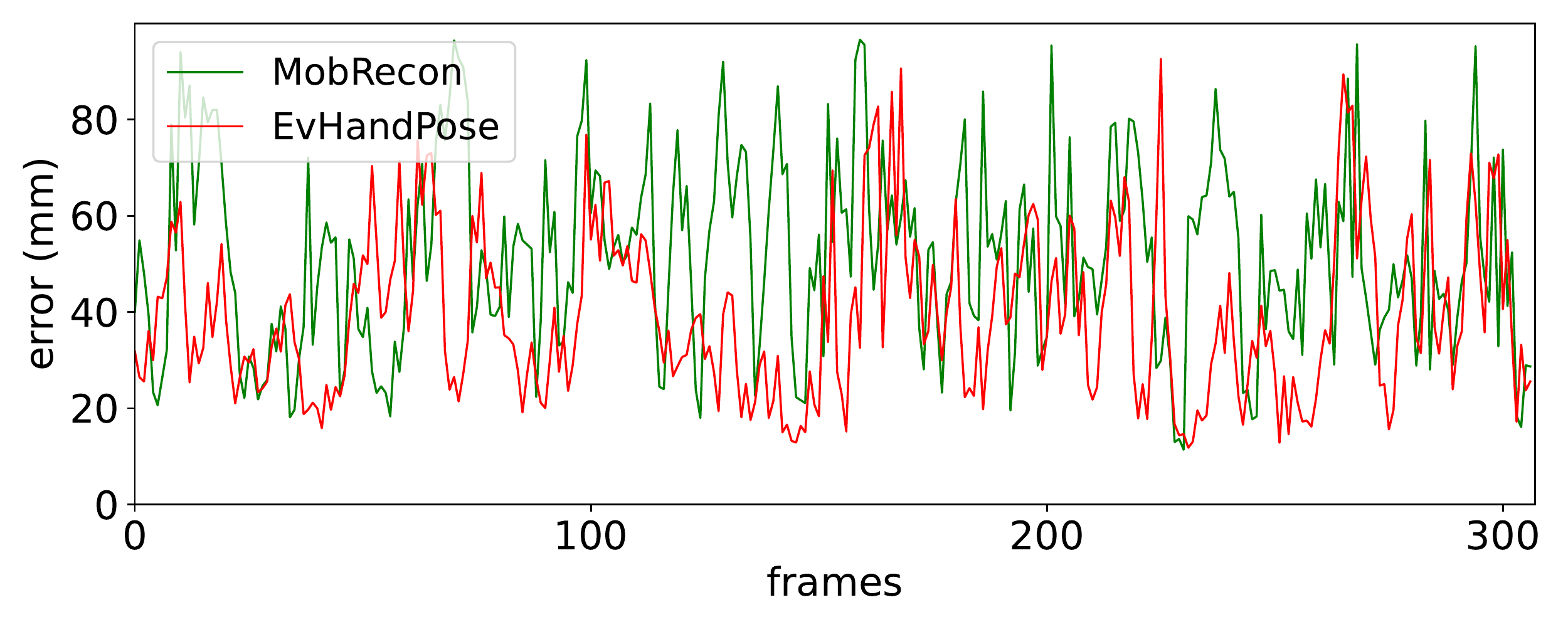}}\\
  \caption{Errors on two strong light sequences. Under strong light, EvHandPose has lower prediction error than MobRecon \cite{chen2021mobrecon}.}
  \label{strong light analysis}
\end{figure}

\Fref{AUC curves} shows the 3D PCK curves and AUC of EvHandPose and baselines under various scenes.
EvHandPose outperforms EventHands~\cite{rudnev2021eventhands} and the method of Jalees \etal \cite{nehvi2021differentiable} in all the scenes with AUC 0.16 higher.
Besides, EvHandPose outperforms RGB-based method MobRecon~\cite{chen2021mobrecon} by AUC 0.11 $\sim$ 0.22 higher under strong light scenes, and achieves comparable results under normal and flash scenes.
\Fref{strong light analysis} shows additional quantitative results in strong light scenes between EvHandPose and MobRecon~\cite{chen2021mobrecon}.
It can be concluded that EvHandPose can predict more stable hand poses under strong light scenes than MobRecon~\cite{chen2021mobrecon}, which derives from the effective hand motion representations to retain the hand movements under challenging illumination.

\subsection{Comparison of Flow Representations}
Several existing works encode the meshes into the flow representations.
EventHPE \cite{zou2021eventhpe} projects body flow with vertex movements to the image plane with
bilinear interpolation, denoted as linear flow, to encode the body shape.
Hasson \etal{} \cite{hasson2020leveraging} perform linear interpolation on vertex motion and differentiably render
the hand motion onto the image plane, denoted as vertex flow.
We perform qualitative and quantitative comparison among linear flow, vertex flow, and our mesh flow to show the superiority of mesh flow for event-based hand pose estimation.

\begin{figure*}[ht]
    \centering
    \subfloat{\includegraphics[width=\linewidth]{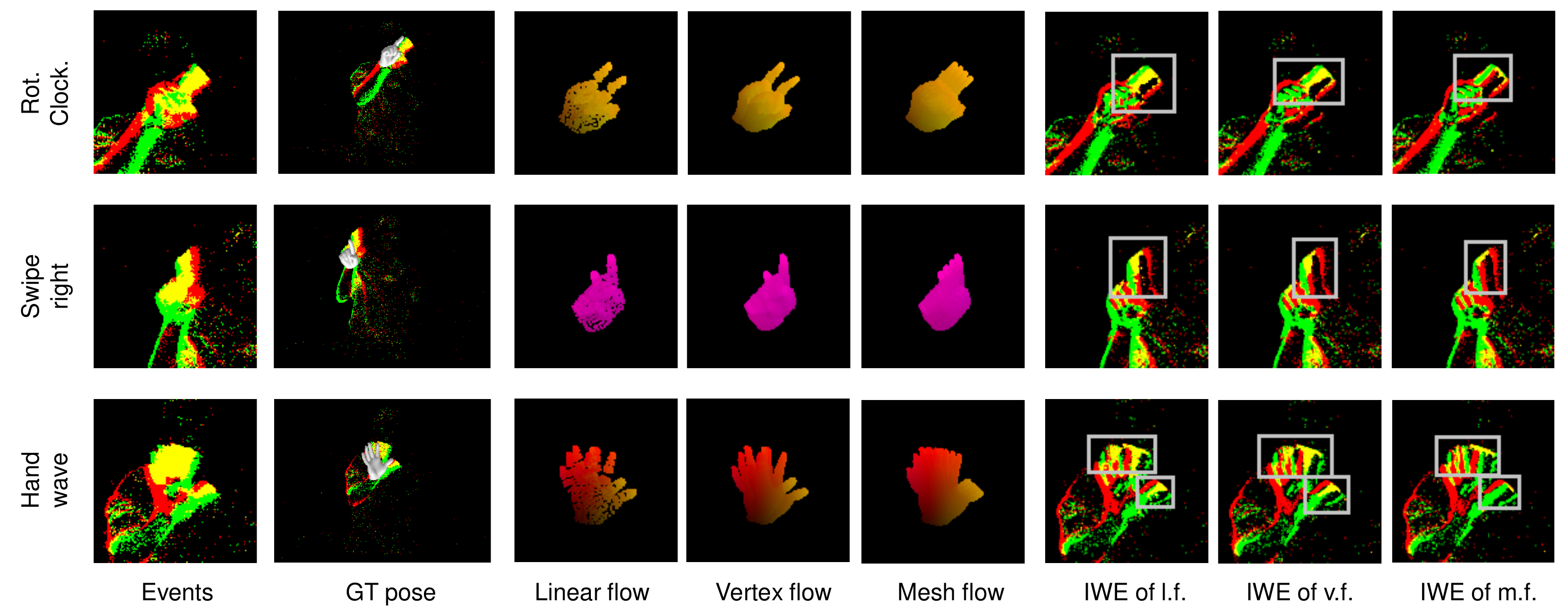}}
    \caption{Comparison among linear flow (l.f.), vertex flow (v.f.), and mesh flow (m.f.). In IWEs, red pixels represent positive events, green pixels represent negative events, and yellow pixels
represent both types of events.
%%%DXM: it is ok to claim the visualization of event frames here.
An IWE with higher contrast will have aggregations of red and green pixels and fewer yellow pixels. As shown in the gray boxes, IWEs by our mesh flow have sharper edges than the others. 
    }
    \label{qualitative flow comparison}
\end{figure*}

\noindent\textbf{Qualitative Comparison of Flow Representations}
Hand flow representation should approximate the hand motion as precisely as possible, \ie{}, the events triggered at the same hand position should gather together to form an IWE with high contrast after being warped by the hand flow. 
In \Fref{qualitative flow comparison}, we show three kinds of hand flow representations and their corresponding IWEs on three sequences.
As shown in the visualization of hand flows, linear flow leads to holes in flow representation, because the vertices of triangles on the mesh are not dense enough to cover the image plane.
Vertex flow can not approximate the rotation motion of fingers, and the events triggered from the fingers are not warped efficiently.
Mesh flow performs interpolation on the hand parameters to approximate the finger motion and renders the hand motion differentiably, thus leading to IWEs with higher contrast.

\noindent\textbf{Quantitative Comparison of Flow Representations}
The quality of hand flow representation is essential for the contrast of IWE, thus influencing the effect of contrast maximization constraint in our self-supervision framework.
We evaluate the performance of EvHandPose under the CM loss by different hand flow representations.
As shown in \Tref{quantitative flow comparison}, our mesh flow leads to 1.5 $\sim$ 3 mm MPJPE decrease compared with linear flow and vertex flow on EvRealHands.
The main reason is that mesh flow can describe the hand movement precisely and impose strong priors for the constraints in self-supervision.

\begin{table}[t]
    \caption{Quantitative experiments results of different hand flow representations.}
    \centering
    \resizebox{\linewidth}{!}{
    \begin{tabular}{c|ccccc}
    \toprule
    Scenes & Poses & Metrics & \thead{EvHandPose\\ mesh flow} & \thead{EvHandPose \\ linear flow} & \thead{EvHandPose\\ vertex flow} \\
    \cmidrule(r){1-6}
    \multirow{4}{*}{Normal} & \multirow{2}{*}{Fixed} & MPJPE & \textbf{20.98} & 24.29 & 23.91\\
    & & AUC & \textbf{0.780} & 0.761 & 0.765 \\
    \cmidrule(r){2-6}
    & \multirow{2}{*}{Random} & MPJPE & \textbf{31.16} & 34.25 & 34.42\\
    & & AUC & \textbf{0.690} & 0.669 & 0.667 \\
    \cmidrule(r){1-6}
    
    \multirow{1}{*}{Fast motion} & \multirow{1}{*}{Random} & 2D-MPJPE & \textbf{8.18} & 11.23 & 11.69\\
    \cmidrule(r){1-6}   
    
    \multirow{4}{*}{Strong light} & \multirow{2}{*}{Fixed} & MPJPE & \textbf{29.62 }& 31.38 & 31.04 \\
    & & AUC & \textbf{0.691} & 0.683 & 0.684 \\
    \cmidrule(r){2-6}
    & \multirow{2}{*}{Random} & MPJPE & \textbf{38.45} & 38.77 & 38.47 \\
    & & AUC & \textbf{0.625} & 0.620 & 0.622 \\
    \cmidrule(r){1-6}
    
    \multirow{4}{*}{Flash} & \multirow{2}{*}{Fixed} & MPJPE & \textbf{33.31} & 37.88 &38.51 \\
    & & AUC & \textbf{0.656} & 0.639 & 0.636 \\
    \cmidrule(r){2-6}
    & \multirow{2}{*}{Random} & MPJPE & \textbf{49.77} & 51.99 & 52.11\\
    & & AUC & \textbf{0.525} & 0.518 & 0.520 \\

    \bottomrule
    \end{tabular}
    }
    \label{quantitative flow comparison}
\end{table}

\subsection{Comparison of Hand-edge Loss}
Hand-edge loss is designed to enforce the alignment of the projection of predicted hand mesh and the edge of IWE, while it is challenging to find reliable correspondences between events and hand mesh vertices.
EventCap \cite{xu2020eventcap} obtains the closest event for each vertex by the criterion based on the spatial and temporal distance in its event-based pose refinement step (denoted as vertex-to-event).
However, it suffers from two issues: 1) Events are not aligned at the same time and 2) fingers will gather together in its optimization loop.

%%%DXM: evaluate->compare
We compare the effect of the hand-edge losses using Pose-to-IWE module (denoted as IWE-to-vertex) and vertex-to-event strategies by applying them as constraints in hand model fitting problem.
We select 65 annotated event timestamps in normal random sequences as testing samples.
Given an annotated event sub-segment, we add Gaussian noise with variance $\sigma$ on its ground truth hand parameters as its initial hand parameters:
\begin{equation}
    {\boldsymbol{\varphi}}_{\text{init}} = {\boldsymbol{\varphi}}_{\text{gt}} + N(0, \sigma),
\end{equation}
where ${\boldsymbol{\varphi}}_{\text{init}}$ is initial hand parameters, ${\boldsymbol{\varphi}}_{\text{gt}}$ is ground truth hand parameters, and $N(\cdot, \cdot)$ is the Gaussian noise.
%%%DXM: the symbol \boldsymbol{\varphi} is not consistent with the main submission. 
%%% in the main submssion,  "we denote $\boldsymbol{\varphi}=(\boldsymbol{\beta}, \boldsymbol{\theta})$ to be \textit{hand parameters} for easy reference."

Then the hand model fitting problem can be formulated as solving a non-linear optimization problem:
\begin{equation}
     {\boldsymbol{\varphi}}_{\text{opt}}=\underset{{\boldsymbol{\varphi}}}{\text{arg min}}\quad (\lambda_{\text{edge}}\loss_{\text{edge}} + \lambda_{\text{smooth}}\loss_{\text{smooth}}),
\end{equation}
where ${\boldsymbol{\varphi}}_{\text{opt}}$ is the optimal hand parameters, $\loss_{\text{edge}}$, $\loss_{\text{smooth}}$ and their loss weights $\lambda_{\text{edge}}$, $\lambda_{\text{smooth}}$ are the same with those in our self-supervision framework. 
We use Adam \cite{kingma2014adam} optimizer with learning rate 0.005 for the iterative optimization loop.
For quantitative analysis, we use aligned 3D MPJPE as evaluation metric.

\begin{figure}[t]
    \centering
    \subfloat{\includegraphics[width=\linewidth]{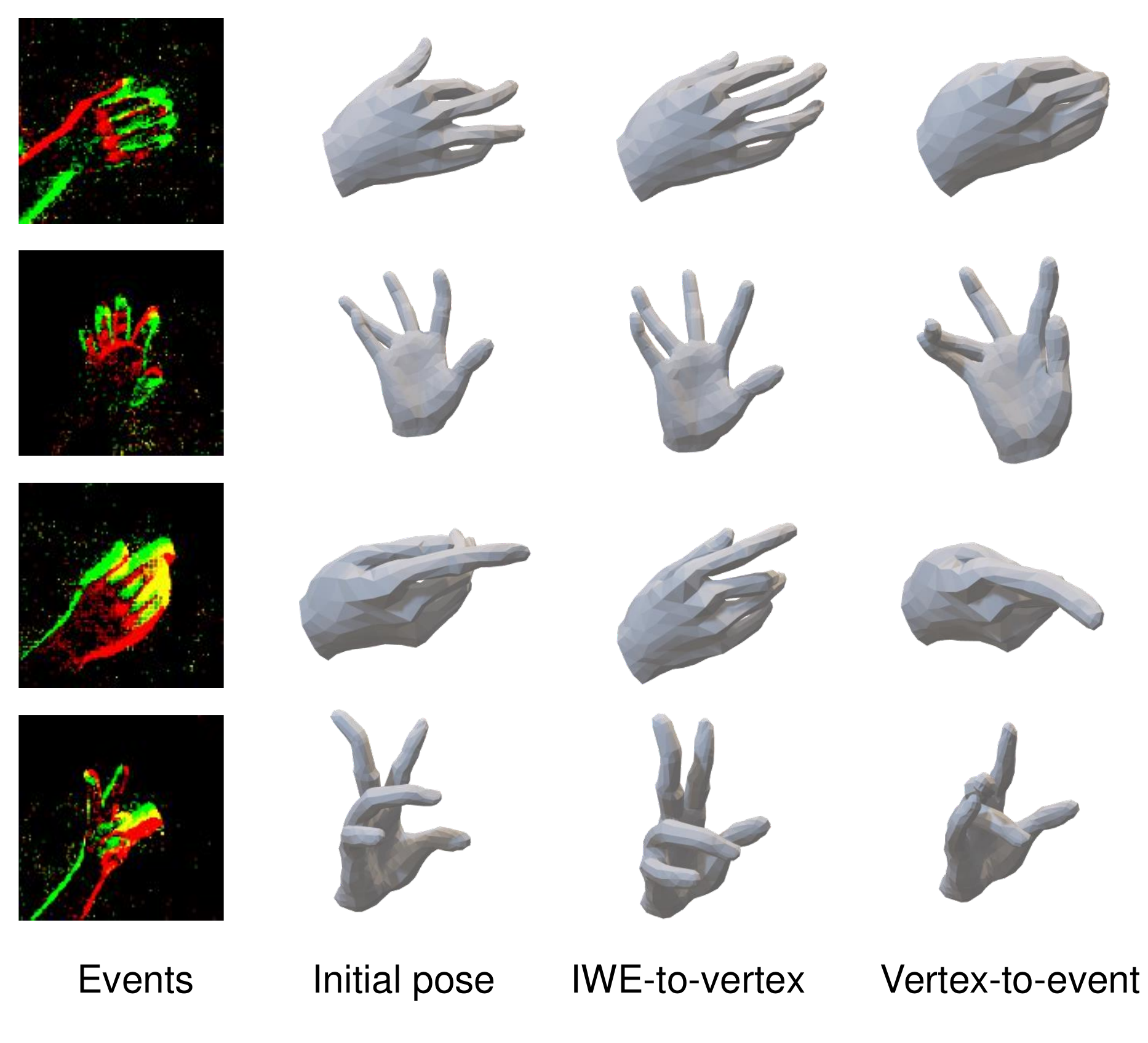}}
    \caption{Optimized hand poses with hand-edge losses using IWE-to-vertex and vertex-to-event strategies.
    Our hand-edge loss with IWE-to-vertex can prevent the fingers from gathering together and produce stable hand poses.
    }
    \label{qualitative edge comparason}
\end{figure}

\begin{figure*}[ht]
    \centering
    \subfloat{\includegraphics[width=\linewidth]{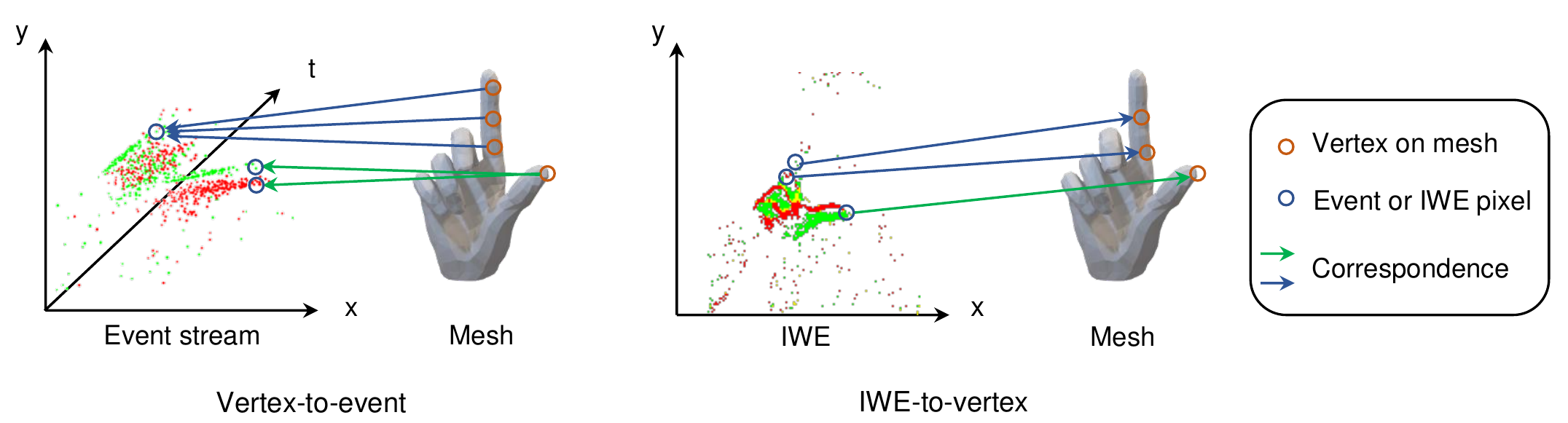}}
    \caption{Comparison of vertex-to-event and IWE-to-vertex method for hand-edge losses. In vertex-to-event method, an event may correspond to several vertices (shown in blue arrows), which will make fingers gather together in optimization. However, in IWE-to-vertex method, one IWE pixel can only match one vertex, which can avoid the issue of vertex-to-event method.
    In vertex-to-event method, vertex matches its corresponding event using the spatial and temporal distance, which will lead to misalignment (shown in green arrows) because temporal distance is not a good measure of the spatial distance at the target timestamp. In IWE-to-vertex method, we solve this issue by warping the events to the target timestamp.
    }
    \label{edge method comparison}
\end{figure*}

As shown in \Tref{quantitative edge comparison}, our hand-edge loss can effectively reduce the MPJPE by 10$\%$ in the optimization process, while vertex-to-event hand-edge loss is less effective.
Qualitative analysis in \Fref{qualitative edge comparason} shows that IWE-to-vertex hand-edge loss can prevent fingers from gathering together in the optimization and obtain stable results.
%%%DXM: the reasons below are not strong 
As shown in \Fref{edge method comparison}, one reason is that we warp the events into the target timestamp to align the events temporally.
The other reason is that finding vertex correspondence for each IWE pixel leads to reliable alignments than finding event correspondence for each vertex.
\iffalse
%%%DXMv2
The other reason is that finding vertex correspondence for each IWE pixel leads to reliable alignments than finding event correspondence for each vertex.
\fi 

\begin{table}[t]
    \centering
    \caption{MPJPEs of the two hand-edge losses using IWE-to-vertex and Vertex-to-event strategies. $\Delta$MPJPE is the change of MPJPE before and after optimization.}
    \resizebox{0.8\linewidth}{!}{
    \begin{tabular}{c|ccc}
    \toprule
    \thead{$\sigma$} & \thead{Initial \\MPJPE} & \thead{IWE-to-vertex\\$\Delta$MPJPE} & \thead{Vertex-to-event\\$\Delta$MPJPE} \\
    \cmidrule(r){1-4}
    0.16 & 23.25 & -1.95 & +3.40 \\
    0.2 & 28.90 & -2.98 & +0.81 \\
    0.24 & 34.44 & -3.40 & -0.81\\
    0.28 & 39.87 & -3.84 & -0.53\\
    0.32 & 45.16 & -5.35 & -1.55\\
    \bottomrule
    \end{tabular}
    }
    \label{quantitative edge comparison}
\end{table}

\section{Addtional Details of Dataset EvRealHands}

\begin{table}[t]
    % \captionsetup{name=\textcolor{magenta}{TABLE}, labelformat=magentatext, labelsep=magentacolon}
    % \color{magenta}
    \caption{Detailed attributes of EvRealHands. The duration of event streams, the quantity of RGB images, and their annotations in each scene are shown below. Annotations [mc] means the 3D annotations are checked or annotated  manually, 
    while Annotations [all] refer to all the machine and manual annotations. For fast motion sequences, the annotations are 2D keypoints.}
    \begin{center}
    \resizebox{\linewidth}{!}{
    \begin{tabular}{c|ccccccc }
    \toprule
    \multirow{2}{*}{Data} & \multicolumn{2}{c}{Normal} & \multicolumn{2}{c}{Strong light} & \multicolumn{2}{c}{Flash} & Fast motion \\
    & Fixed & Random & Fixed & Random & Fixed & Random & \\
    \cmidrule(r){1-8}
    Event streams (s) & 850.7 & 2874.5 & 210.3 & 348.2 & 158.8 & 157.9 & 242.4 \\
    RGB images  & 79.7 K & 267.6 K & 14.8 K & 28.3 K & 14.9 K & 13.8 K & 7.8 K\\
    Annotations [mc] & 11.4 K & 9.7 K & 2.0 K & 1.9 K & 2.1 K & 1.9 K & 0.8 K  \\
    Annotations [all] & 11.4 K & 37.2 K & 2.0 K & 3.9 K & 2.1 K & 1.9 K & 0.8 K  \\
    \bottomrule
    \end{tabular}
    }
    \end{center}
    \label{dataset composition}
\end{table}

\subsection{Adequacy of Data}
\begin{figure}[t]
    \centering
    \subfloat{\includegraphics[width=\linewidth]{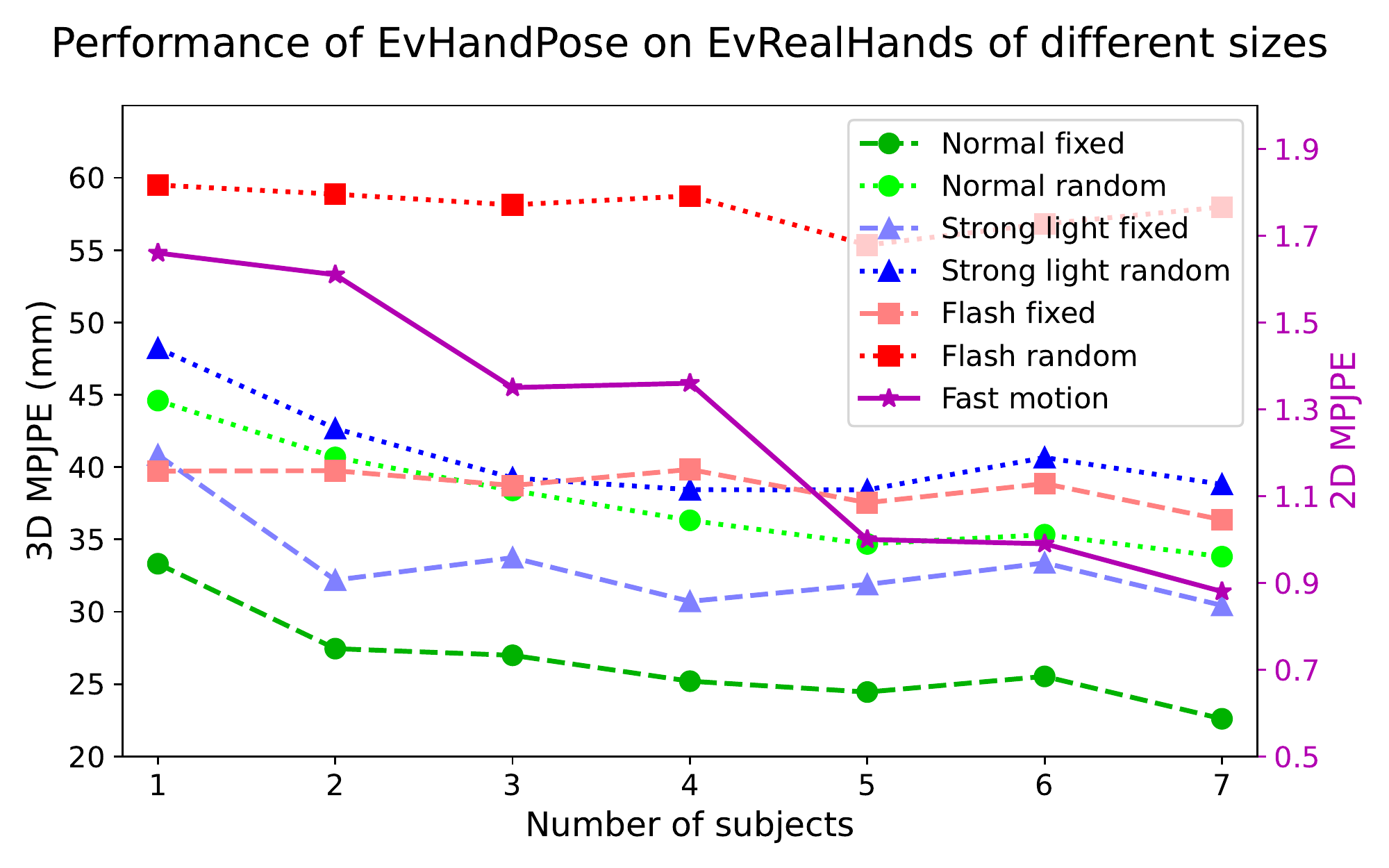}}
    \caption{Performance of EvHandPose on EvRealHands with different sizes.
    EvHandPose nearly reaches the best performance when there are 5 subjects in the training set. 
    }
    \label{data adequacy}
\end{figure}

The amount of sequences and annotations of our dataset under various scenes is shown in \Tref{dataset composition}.
Verifying the adequacy of our dataset is important in dataset creation \cite{varol2017learning}.
To achieve this, we evaluate the performance of EvHandPose on training sets with different numbers of subjects ranging from 1 to 7.
As shown in \Fref{data adequacy}, EvHandPose achieves better performance as the number of subjects in training data increases.
Since there is only one subject captured under strong light and flash in our training set, the evaluations are conducted mostly on the collected data in normal scenes.
When more than 5 subjects are used for model training, the increase in subjects will no longer significantly improve the performance.
%%%DXMv2:我注释了下面这句话
% of EvHandPose in normal scenes.
%%%DXMv2:我注释了下面这句话
% It is worth noting that with the increase of subjects of training data in normal scenes, 
EvHandPose achieves better results in strong light and fast motion scenes but very little boost in flash scenes.
One explanation is that the edge and hand flow features learned by EvHandPose in normal scenes can be effectively generalized to the strong light and fast motion scenes.
In flash scenes, the events generated by the background often overwhelm the events generated from the hand movement, so it
makes it challenging for EvHandPose to extract effective motion clues.
In future research, more data under challenging scenes (strong light, fast motion, \etc{},) can be collected to bridge the performance gap between normal scenes and challenging scenes.

\subsection{Scanned Hand Meshes}
\begin{figure}[h]
  \centering
  \captionsetup[subfigure]{labelformat=empty}
  \subfloat[Left hands]{\includegraphics[width=0.1\linewidth]{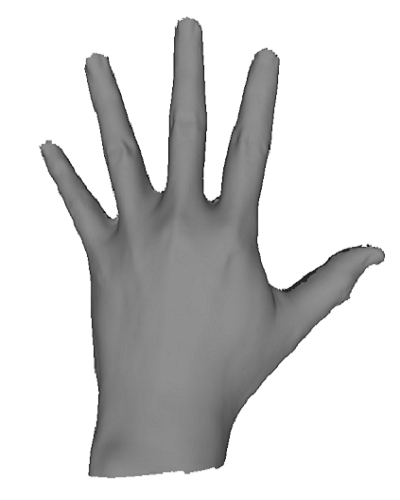}
  \includegraphics[width=0.1\linewidth]{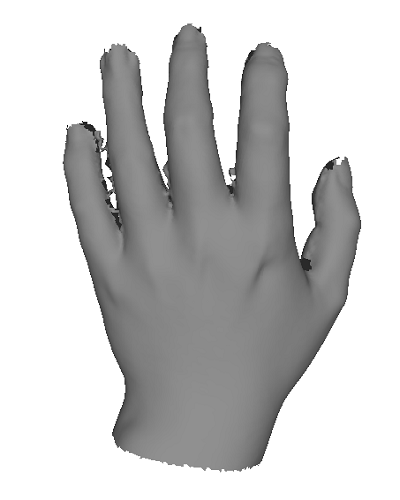}
  \includegraphics[width=0.1\linewidth]{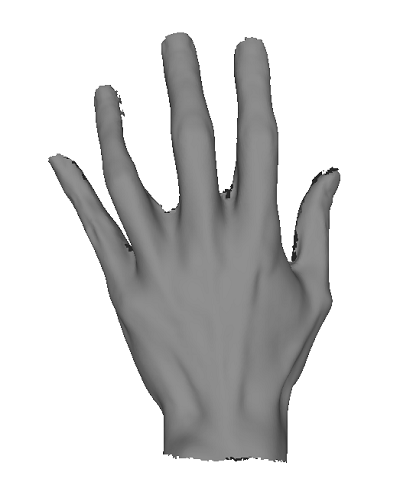}
  \includegraphics[width=0.1\linewidth]{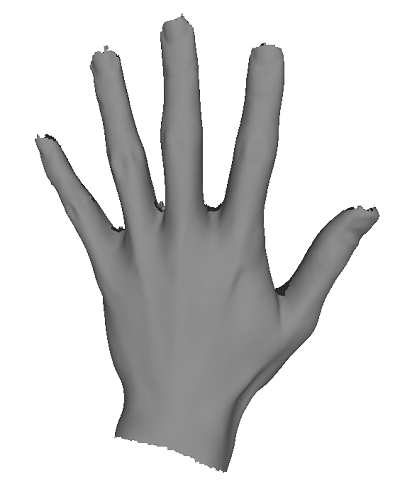}
  \includegraphics[width=0.1\linewidth]{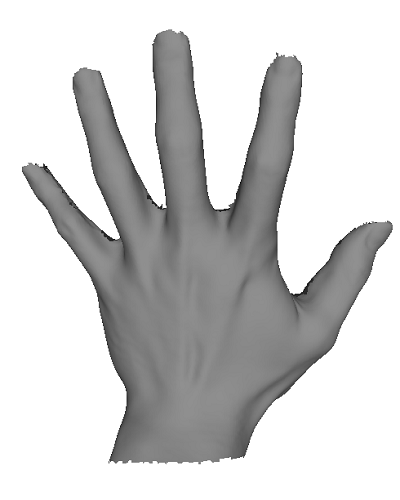}
  \includegraphics[width=0.1\linewidth]{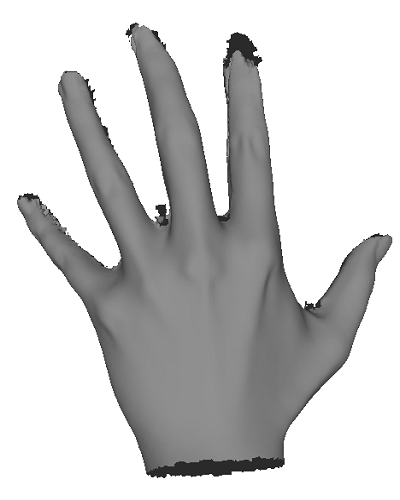}
  \includegraphics[width=0.1\linewidth]{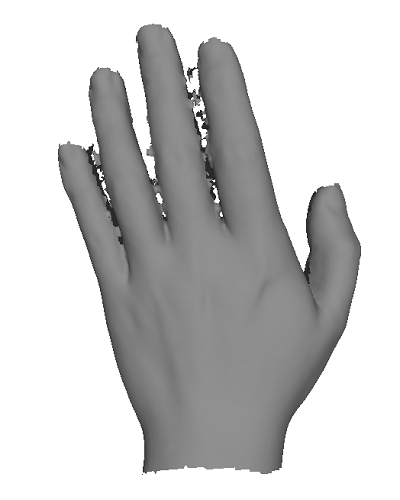}
  \includegraphics[width=0.1\linewidth]{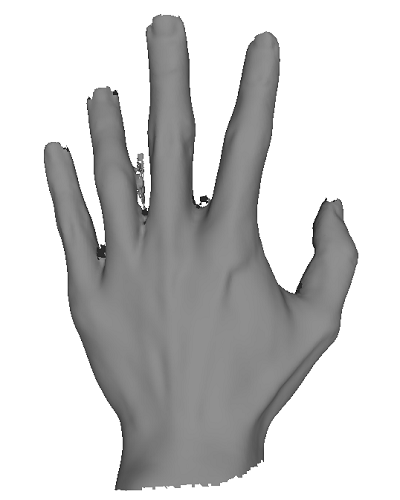}
  \includegraphics[width=0.1\linewidth]{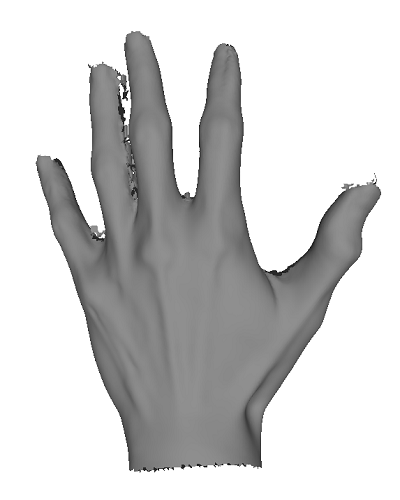}
  }\\
  \subfloat[Right hands]{\includegraphics[width=0.1\linewidth]{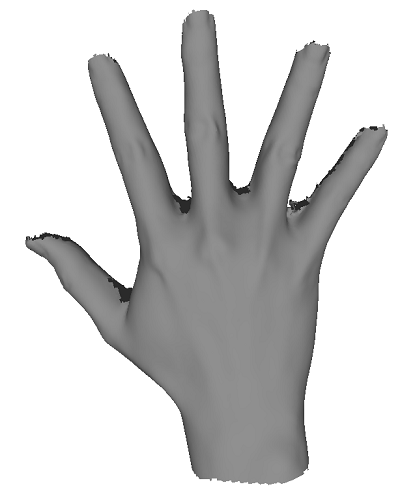}
  \includegraphics[width=0.1\linewidth]{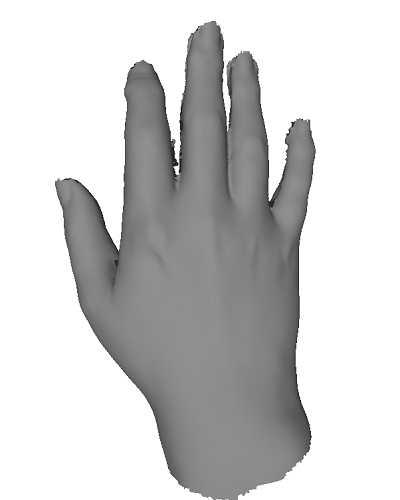}
  \includegraphics[width=0.1\linewidth]{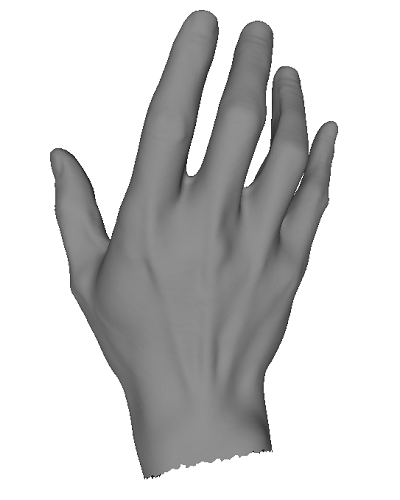}
  \includegraphics[width=0.1\linewidth]{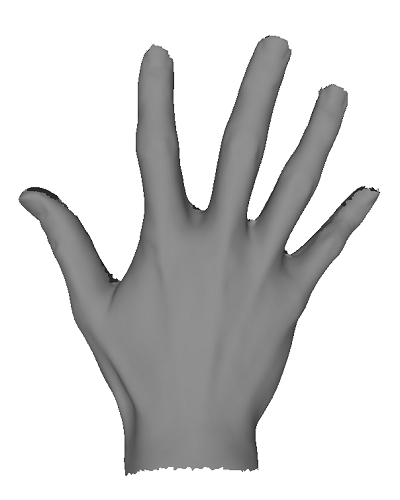}
  \includegraphics[width=0.1\linewidth]{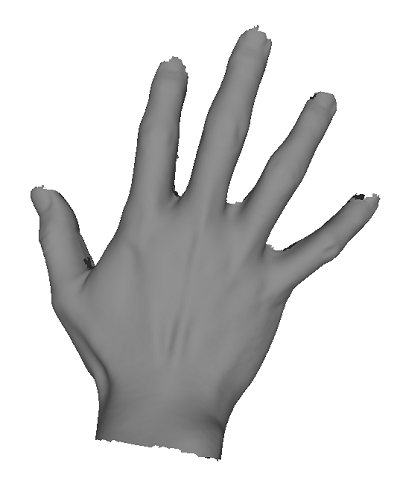}
  \includegraphics[width=0.1\linewidth]{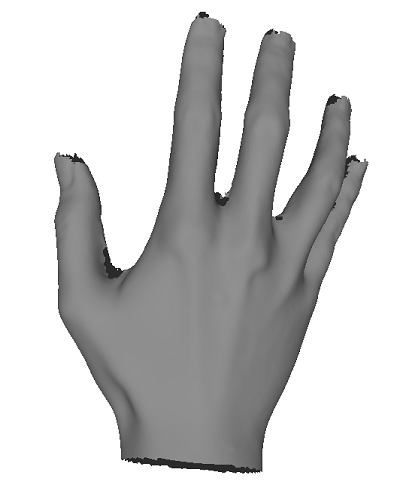}
  \includegraphics[width=0.1\linewidth]{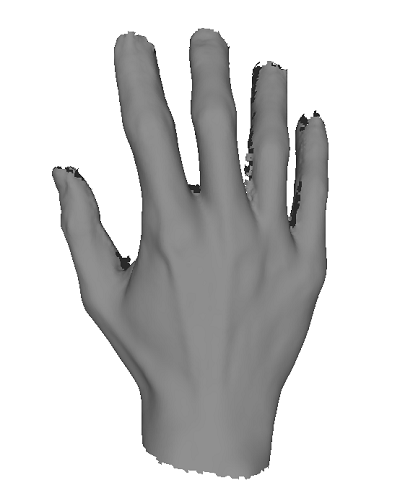}
  \includegraphics[width=0.1\linewidth]{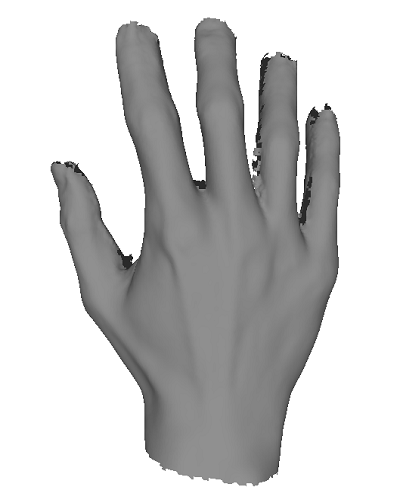}
  \includegraphics[width=0.1\linewidth]{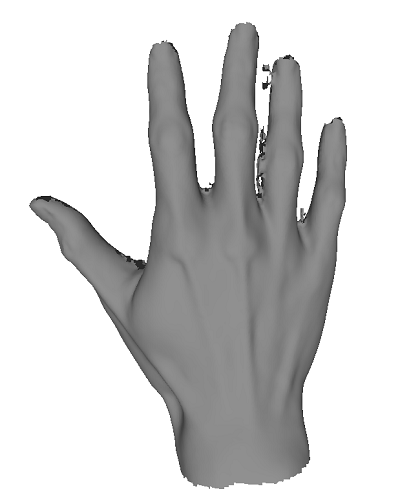}
  }
  \caption{Scanned hand meshes. We use 3D scanner to get precise hand mesh for shape annotation.}
  \label{handmeshes}
\end{figure}

For shape annotation, we use an Artec Eva 3D Scanner to scan the hands of each subject, and the shapes are shown in \Fref{handmeshes}.

\begin{figure}[t]
    \captionsetup[subfigure]{labelformat=empty}
\centering
    \subfloat[Grab]{\includegraphics[width=0.3\linewidth]{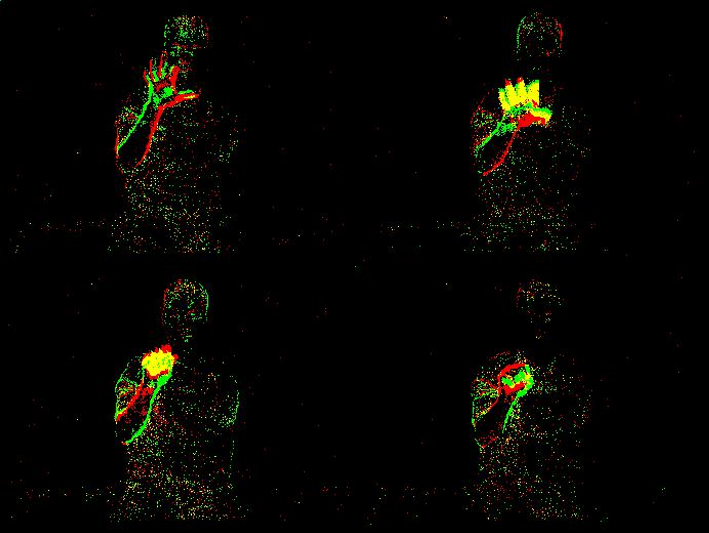}}
    \hfill
    \subfloat[Tap]{\includegraphics[width=0.3\linewidth]{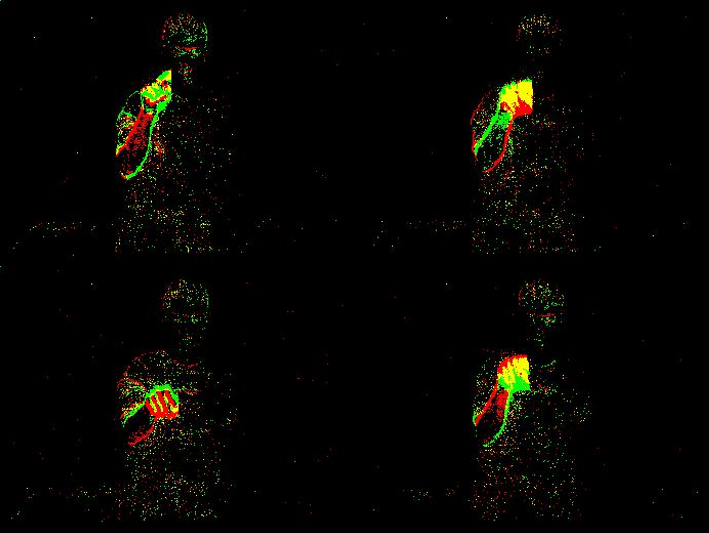}}
    \hfill
    \subfloat[Expand]{\includegraphics[width=0.3\linewidth]{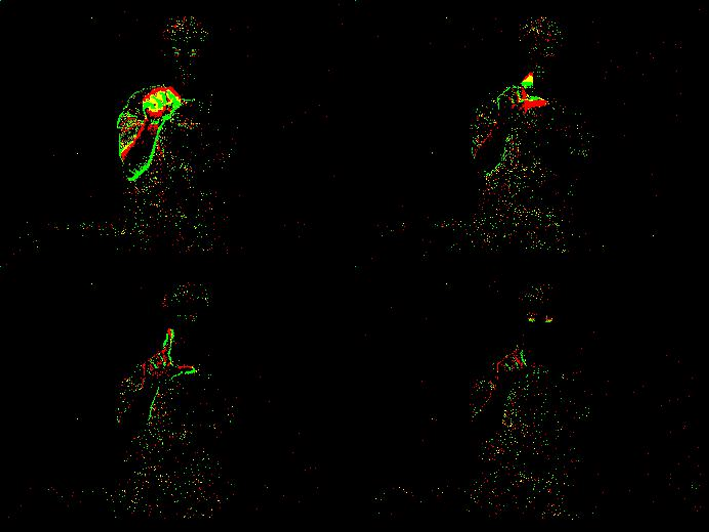}}\\
    \subfloat[Pinch]{\includegraphics[width=0.3\linewidth]{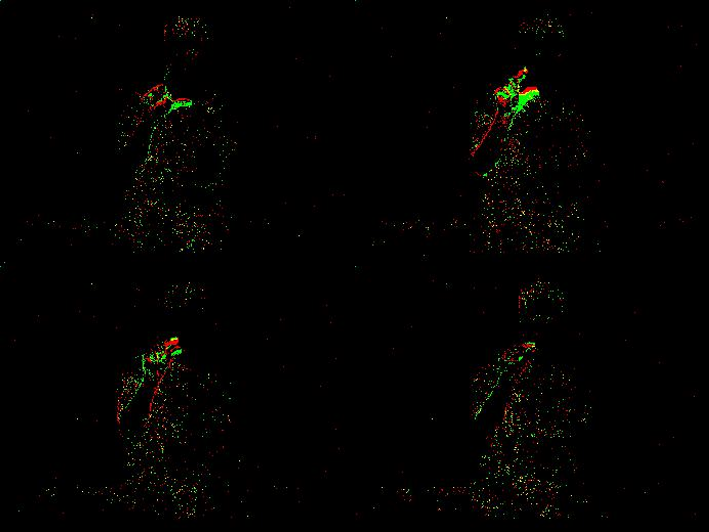}}
    \hfill
    \subfloat[Rot. clock.]{\includegraphics[width=0.3\linewidth]{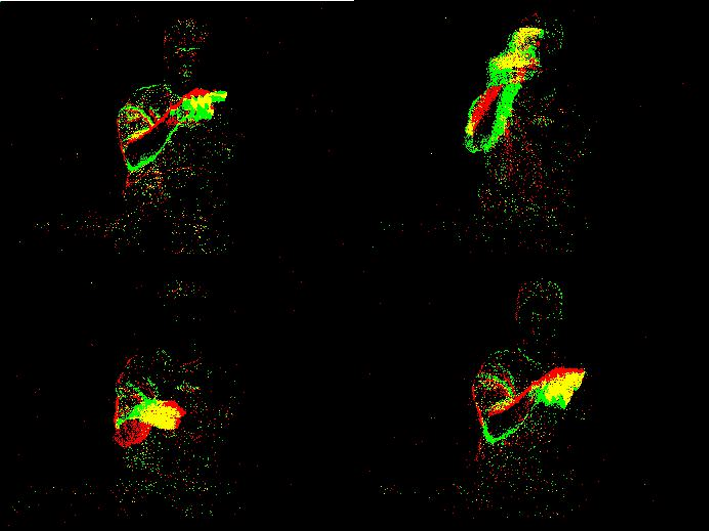}}
    \hfill
    \subfloat[Rot. counter clock.]{\includegraphics[width=0.3\linewidth]{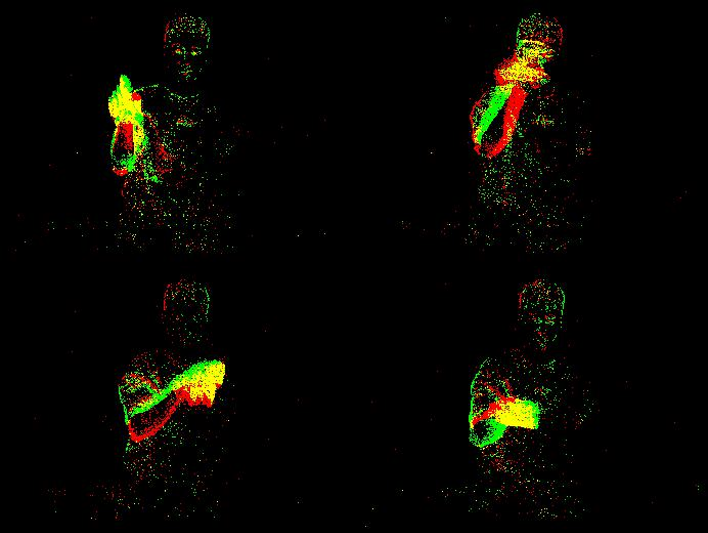}}\\
    \subfloat[Swipe right]{\includegraphics[width=0.3\linewidth]{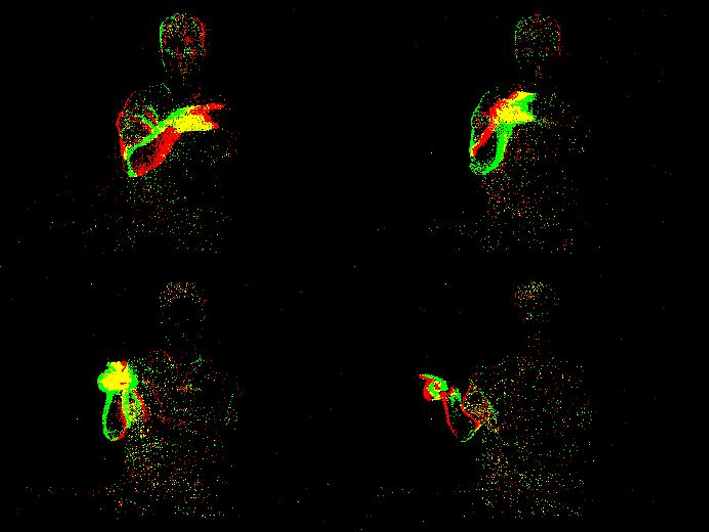}}
    \hfill
    \subfloat[Swipe left]{\includegraphics[width=0.3\linewidth]{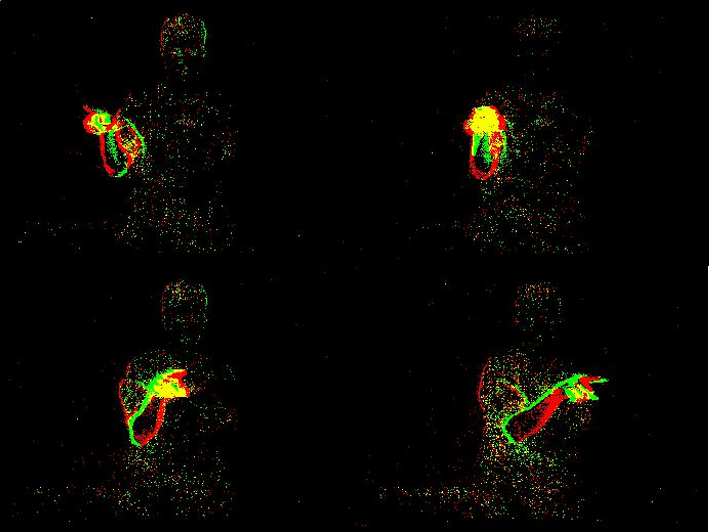}}
    \hfill
    \subfloat[Swipe up]{\includegraphics[width=0.3\linewidth]{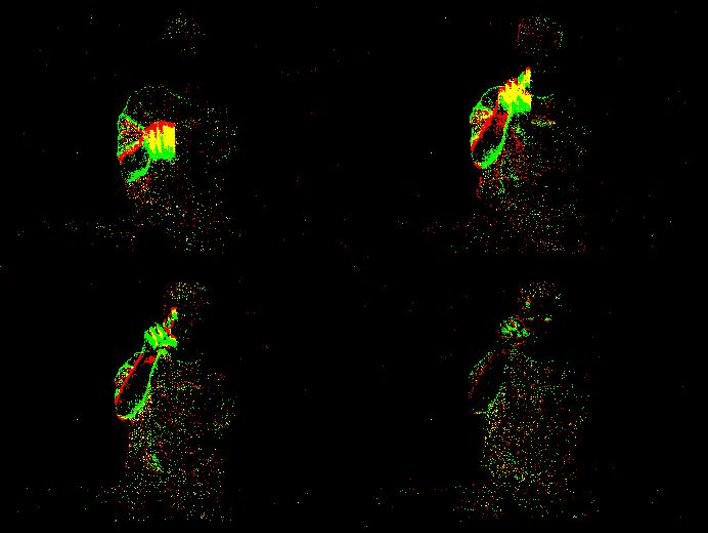}}\\
    \subfloat[Swipe down]{\includegraphics[width=0.3\linewidth]{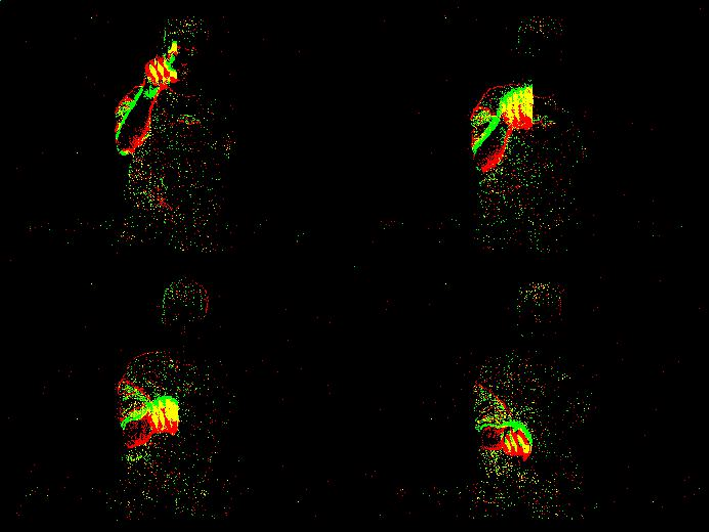}}
    \hfill
    \subfloat[Swipe X]{\includegraphics[width=0.3\linewidth]{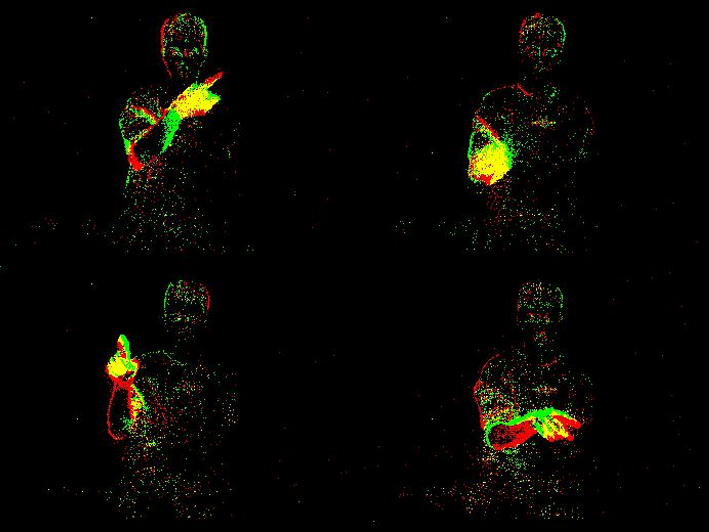}}
    \hfill
    \subfloat[Swipe $+$]{\includegraphics[width=0.3\linewidth]{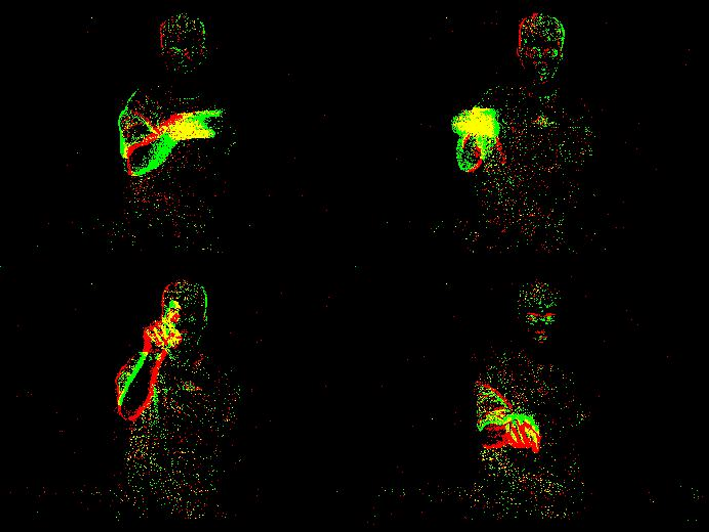}}\\
    \subfloat[Swipe V]{\includegraphics[width=0.3\linewidth]{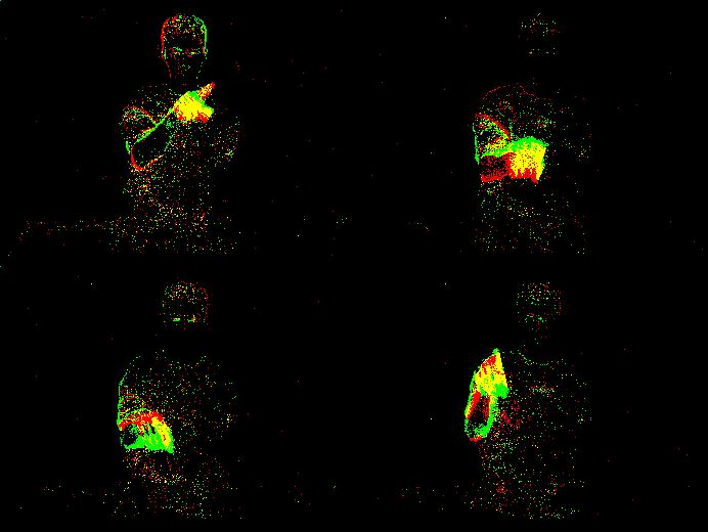}}
    \hfill
    \subfloat[Shake]{\includegraphics[width=0.3\linewidth]{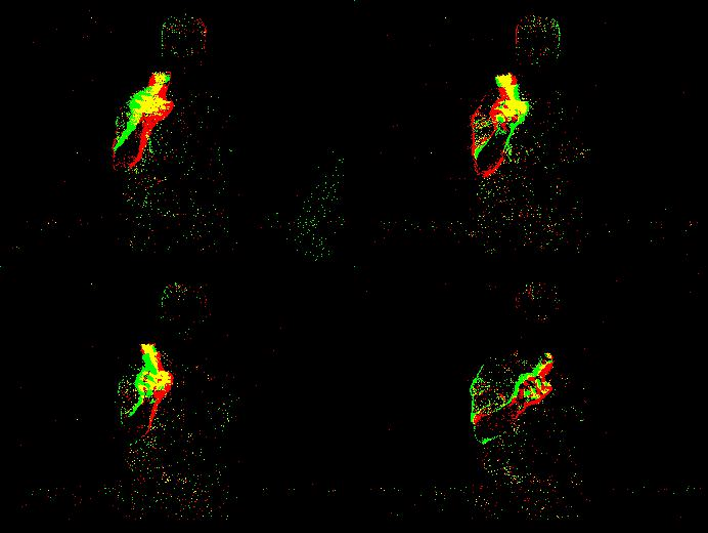}}
    \hfill
    \subfloat[Hand wave]{\includegraphics[width=0.3\linewidth]{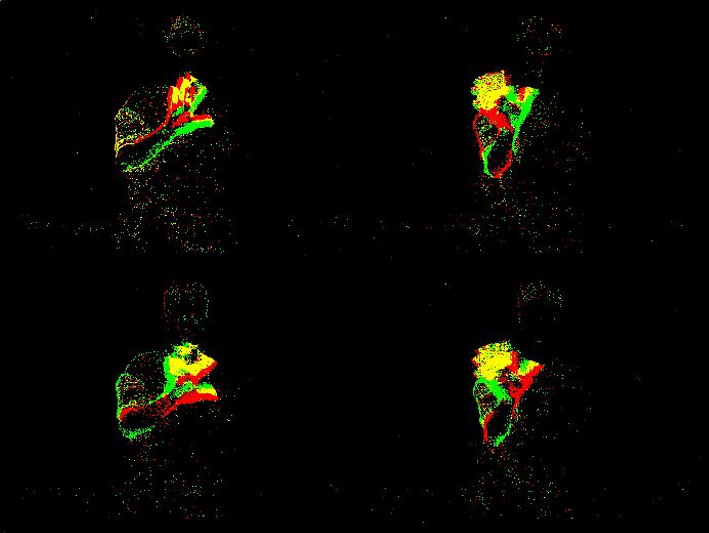}}
\caption{Illustration of the 15 fixed pose gestures in event frames (``Rot." and ``clock." are short for rotation and clockwise).}
\label{fixed poses}
\end{figure}

\subsection{Hand Gestures}
We visualize the event streams of 15 hand gestures similar to \cite{de20173d} in \Fref{fixed poses}.
These gestures are \textit{Grab}, \textit{Tap}, \textit{Expand}, \textit{Pinch}, \textit{Rotation}, \textit{Rotation counter clockwise}, \textit{Swipe right}, \textit{Swipe left}, \textit{Swipe up}, \textit{Swipe down}, \textit{Swipe X}, \textit{Swipe +}, \textit{Swipe V}, \textit{Shake}, \textit{Handwave}.
The visualization shows that event cameras can effectively capture the hand movements from the still background.
By annotating the sequences, we obtain 15 gesture classes for hand gesture recognition.

% or


% \appendices

% % you can choose not to have a title for an appendix
% % if you want by leaving the argument blank
% \section{}

% \ifCLASSOPTIONcaptionsoff
%   \newpage
% \fi
% {
% \bibliographystyle{IEEEtran}
% % \bibliography{eg}
% }

% \vfill

% % Can be used to pull up biographies so that the bottom of the last one
% % is flush with the other column.
% %\enlargethispage{-5in}

% \end{document}